%% file: main.tex
\documentclass{article}

\usepackage{iclr2026_conference,times}
\iclrfinalcopy

\usepackage[utf8]{inputenc}
\usepackage[T1]{fontenc}
\usepackage{microtype}

\usepackage{amsmath}
\usepackage{alphalph}
\usepackage{amssymb}
\usepackage{amsfonts}
\usepackage{amsthm}
\usepackage{bm}
\usepackage{nicefrac}

\theoremstyle{definition}
\newtheorem{proposition}{Observation}

\usepackage{booktabs}
\usepackage{multirow}
\usepackage{longtable}
\usepackage{graphicx}
\usepackage{wrapfig}
\usepackage{subcaption}
\usepackage{xcolor}
\graphicspath{{figures/}}

\usepackage{enumitem}
\usepackage{algorithm}
\usepackage{algorithmic}

\usepackage[section]{placeins}

\usepackage{url}
\usepackage{hyperref}
\hypersetup{colorlinks=true,citecolor=blue,linkcolor=blue,urlcolor=blue}
\usepackage[capitalize,noabbrev]{cleveref}
\crefname{proposition}{Observation}{Observations}
\Crefname{proposition}{Observation}{Observations}

\input{macros}

\title{\tesseravtwo: Scaling Pixel-wise Earth \\ Foundation Models}

\author{%
  Zhengpeng Feng$^{1}$ \quad Sadiq Jaffer$^{1}$ \quad Ira Shokar$^{2}$ \quad Jovana Knezevic$^{1}$ \quad James Ball$^{1}$ \\
  \textbf{Pedro Sousa$^{1}$ \quad Mark Elvers$^{1}$ \quad Madeline Lisaius$^{1}$ \quad Clement Atzberger$^{3}$} \\
  \textbf{Robin Young$^{1}$ \quad Aneesh Naik$^{1}$ \quad Niall Robinson$^{2}$ \quad David Coomes$^{1}$} \\
  \textbf{Anil Madhavapeddy$^{1}$ \quad Srinivasan Keshav$^{1}$}\thanks{Corresponding author.} \\
  $^{1}$University of Cambridge \qquad $^{2}$NVIDIA \qquad $^{3}$dClimate Labs 
}

\begin{document}
\maketitle
\raggedbottom

\lhead{Preprint.}

\begin{abstract}
\input{sections/00_abstract}
\end{abstract}

\input{sections/01_introduction}
\input{sections/02_related_work}
\input{sections/03_scaling_laws}
\input{sections/04_train_large_teacher}
\input{sections/05_pixel_distillation}
\input{sections/06_discussion}

\subsubsection*{Reproducibility Statement}
We will release training code, 
distilled pixel-wise student and teacher checkpoints and a frozen \geotessera-style embedding-as-data
product.

\subsubsection*{Ethics Statement}
\tesseravtwo{} is a generic representation model for publicly available
Sentinel-1/2 imagery. We discuss dual-use considerations and release
practices in \Cref{sec:disc:impact}.

\subsubsection*{Acknowledgment}
The authors acknowledge the use of resources provided by the Isambard-AI National AI Research Resource (AIRR). Isambard-AI is operated by the University of Bristol and is funded by the UK Government’s Department for Science, Innovation and Technology (DSIT) via UK Research and Innovation; and the Science and Technology Facilities Council [ST/AIRR/I-A-I/1023]. We also acknowledge funding from Jane Street to support compute and storage, and hardware resources from Vultr and AMD that we have used for inference. We further thank Google and Dr.\ Robert Sansom for their financial support of this research.

\bibliographystyle{iclr2026_conference}
\bibliography{references}

\clearpage
\appendix
\renewcommand{\thesection}{\AlphAlph{\value{section}}}
\crefalias{section}{appendix}
\crefalias{subsection}{appendix}
\input{sections/appendix}

\end{document}

%% file: macros.tex
\newcommand{\tessera}{\texorpdfstring{\textsc{TESSERA}}{TESSERA}}
\newcommand{\tesseravone}{\texorpdfstring{\textsc{TESSERA}\,v1}{TESSERA v1}}
\newcommand{\tesseravtwo}{\texorpdfstring{\textsc{TESSERA}\,v2}{TESSERA v2}}
\newcommand{\teacher}{\ensuremath{\mathcal{T}}}      
\newcommand{\geotessera}{\texorpdfstring{\textsc{GeoTessera}}{GeoTessera}}

\newcommand{\alphaearth}{\texorpdfstring{\textsc{AlphaEarth}}{AlphaEarth}}
\newcommand{\presto}{\texorpdfstring{\textsc{Presto}}{Presto}}

\newcommand{\barlowtwins}{\texorpdfstring{\textsc{Barlow Twins}}{Barlow Twins}}
\newcommand{\matryoshka}{\texorpdfstring{\textsc{Matryoshka}}{Matryoshka}}

\newcommand{\sone}{Sentinel-1}
\newcommand{\stwo}{Sentinel-2}
\newcommand{\sonetwo}{Sentinel-1/2}
\newcommand{\dpixel}{\texorpdfstring{\textit{d}-pixel}{d-pixel}}
\newcommand{\dpixels}{\texorpdfstring{\textit{d}-pixels}{d-pixels}}

\newcommand{\nruns}{395}
\newcommand{\ngpus}{1{,}024}
\newcommand{\gputype}{H100}
\newcommand{\nscalingtasks}{15}
\newcommand{\encexp}{0.36}
\newcommand{\projexp}{0.011}
\newcommand{\dataexp}{0.63}
\newcommand{\encexploss}{0.02}   
\newcommand{\wastemult}{2\text{--}5\times}
\newcommand{\wasteexcess}{254\%}
\newcommand{\nbuckets}{nine}     
\newcommand{\teacherparams}{2\,B}
\newcommand{\seqlen}{\{8,16\}}   
\newcommand{\studentL}{44\,M}
\newcommand{\studentM}{21\,M}
\newcommand{\studentS}{7\,M}
\newcommand{\studentN}{1\,M}

\newcommand{\dimset}{\{16,32,64,128\}}
\newcommand{\dimfull}{128}            
\newcommand{\dimteacher}{d_{T}}        

\newcommand{\aebench}{\textsc{AlphaEarth} suite}
\newcommand{\npixeldata}{15}

\newcommand{\nheldout}{14}            
\newcommand{\nfullsuite}{29}          

\newcommand{\R}{\mathbb{R}}

\newcommand{\bz}{\bm{z}}
\newcommand{\bt}{\bm{t}}
\newcommand{\bs}{\bm{s}}
\newcommand{\bo}{\bm{o}}
\newcommand{\loss}{\mathcal{L}}
\newcommand{\Lbt}{\loss_{\mathrm{BT}}}
\newcommand{\Lmix}{\loss_{\mathrm{MIX}}}

\newcommand{\Ldist}{\loss_{\mathrm{DIST}}}


%% file: sections/00_abstract.tex
Pixel-wise Earth-observation (EO) foundation models are 
now achieving state-of-the-art performance via generated spatial embeddings.
However, how these models scale and how best to spend a pretraining budget remain poorly understood. We present the largest controlled scaling
study for EO to date: 395 training runs within a
fixed pixel-wise \barlowtwins{} family, each evaluated on 15 diverse downstream tasks. We find that pretraining loss barely predicts
downstream performance ($\left|\mathrm{Pearson}\ r\right| < 0.2$), so selecting models by
loss wastes a large share of the compute. We also find that, as the
training budget grows, the encoder and the data should grow together
while the projector stays fixed, which gives a simple rule for
allocating compute. Using this rule, we train a family of pixel-wise teachers (0.5B, 1B, and 2B) and distil the largest into compact students for embeddings-as-data deployment. In aggregate, our 44-million-parameter distilled student outperforms every open and proprietary embedding product we test, several of them an order of magnitude larger. These students produce Matryoshka representations that are inexpensive to serve: a 16-dimensional prefix keeps 92\% of the full 128-dimensional performance at 1/8 of the storage. Together, these results give a concrete, empirically grounded recipe for scaling pixel-wise EO foundation models: train large encoders, select by downstream performance, and distil into flexible student models. 
We plan to release global 10 m annual embeddings covering 2017--2025 as version 2 of the TESSERA foundation-model embeddings product.
All code is available at: \url{https://github.com/ucam-eo/tessera}

%% file: sections/01_introduction.tex
\section{Introduction}
\label{sec:intro}

\begin{figure}[t]
  \centering
  \includegraphics[width=\linewidth]{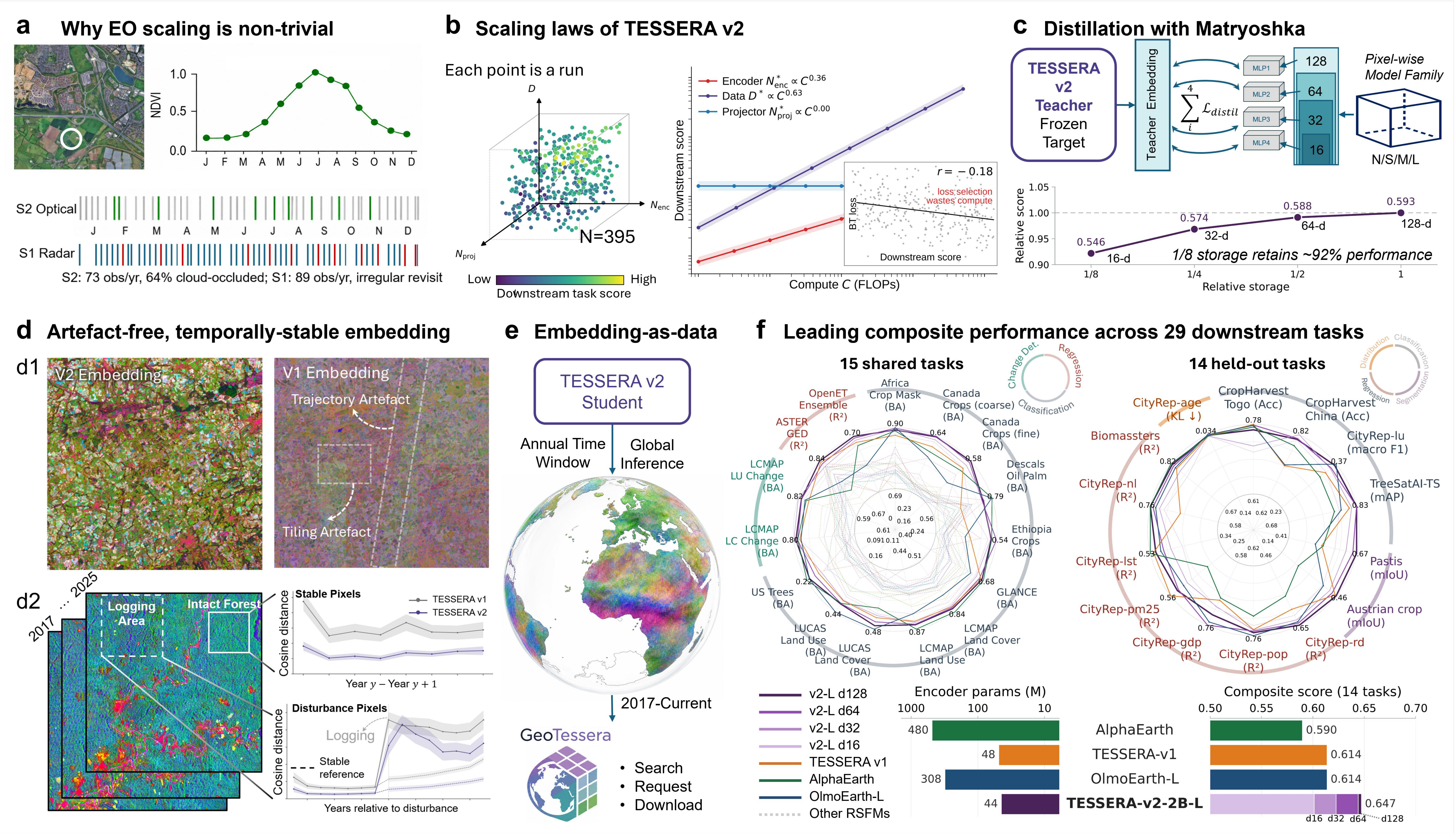}
  \caption{Overview of \tesseravtwo{}.
  \textbf{(a)}~Sparse, irregular \stwo{}/\sone{} sampling at one
  location.
  \textbf{(b)}~$\nruns$-run downstream-driven scaling sweep and
  compute-optimal fits; inset: pretraining loss vs.\ downstream
  score.
  \textbf{(c)}~\matryoshka{} distillation of the $\teacherparams$
  teacher into N/S/M/L students, $d\!\in\!\dimset$.
  \textbf{(d)}~Artefact removal and inter-annual stability vs.\ v1.
  \textbf{(e)}~Deployment via \geotessera{}.
  \textbf{(f)}~Leading composite performance: radars give per-task
  scores on the shared \aebench{} (left) and the held-out datasets
  (right), bars the held-out composite (\tesseravtwo-2B-L, shades
  $d\!\in\!\dimset$) and encoder size.}
  \label{fig:teaser}
\end{figure}

A key bottleneck in using Earth observation (EO) for downstream
tasks is preparing the data: radiometric calibration, cloud and
atmospheric correction, cross-sensor harmonisation, and expensive computation
over raw imagery. Beyond this preparation, the ground-truth labels
these tasks need are often scarce and tied to specific regions and
seasons~\citep{metcalfe2025gaps,hou2026alphaearth}.
The data themselves are awkward in ways that natural images are not: \stwo{} and \sone{}
observe a given pixel at different, irregular cadences, cloud removes
much of the optical record (\Cref{fig:teaser}a), and the cloud-free
composites most models train on remove temporal dynamics that might be highly informative for certain downstream tasks~\citep{zeng2020review,xiao2025foundation}.
A reusable representation layer for
global surface state, delivered as spatially mapped data, can lower these barriers.

Earth embeddings are vector representations of specific places and
times that compress and fuse multi-source
observations~\citep{klemmer2025earth,fang2026earth}. Provided at
global scale, in the `embeddings-as-data' approach, they free analysts
from acquiring heterogeneous raw data, from complex pre-processing,
and from user-side GPU compute; with lightweight task-specific
\textit{heads} they have been shown to match, and often outperform,
\textit{ad hoc} models built on traditional remote sensing data.

An ideal embedding product should be \emph{analysis-ready}, a
geospatial data layer usable without processing raw imagery or running
GPU inference; \emph{transferrable} across tasks, workflows, and
regions rather than tuned to one benchmark~\citep{klemmer2025earth,%
lyu2026structure}; \emph{reproducible}, with inspectable and
extensible algorithms and weights; \emph{economical} in the resources
that limit EO adoption, namely labels, compute and its carbon
footprint, storage, I/O, and engineering time; and \emph{adaptive},
exposing an explicit accuracy-versus-cost trade-off instead of locking
users into a single model size and embedding dimensionality.
Existing systems each satisfy a subset. Remote sensing foundation
models~\citep{cong2022satmae,reed2023scalemae,fuller2023croma,%
guo2024skysense,tseng2024presto,tseng2025galileo,astruc2025anysat}
learn strong representations but release \textit{models} rather than
data products, so users continue to face preprocessing and
inference~\citep{fang2026earth}. Specialised archives are useful
within a limited scope---\textsc{ESD}~\citep{chen2026esd} compresses 25
years of Landsat/MODIS reflectance into quantised $30$\,m
embeddings---but stop short of the specification above in resolution,
sensor scope, or evaluation breadth.
\alphaearth~\citep{brown2025alphaearth} is analysis-ready at global
scale but only open-output: the embeddings are public while the
training procedure and weights are not~\citep{hou2026alphaearth}.
\tesseravone~\citep{feng2026tessera} comes closest, being
analysis-ready, label-efficient, open, and reproducible, but produces a
single fixed $128$-dimensional embedding, so every user inherits the
same storage and I/O cost regardless of their deployment constraints.
No existing system is an open, analysis-ready \emph{family} in which
users pick the model size and embedding dimension that fit their
budget; \Cref{app:desiderata} compares them criterion by criterion.
This is the research gap that we address in our work.

Closing this gap starts upstream, at training time. In the language and
vision domains, \textit{scaling laws} decide how a pretraining budget
is allocated across the network, and are fit empirically to the
pretraining loss~\citep{kaplan2020scaling,hoffmann2022chinchilla,%
zhai2022scaling}. In EO that proxy is suspect, because cloud cover and
orbital sampling dominate input variance and a redundancy-reduction
objective can be minimised through invariances that could carry no
downstream value; the few existing EO scaling studies either entangle
architecture with capacity or fit power laws to the loss rather than
linking performance to downstream
comparison~\citep{dionelis2025scaling,wickrema2025scaling}.
We therefore ran a downstream-driven scaling study, with $\nruns$
controlled pretraining runs of pixel-wise
\barlowtwins~\citep{zbontar2021barlow, lisaius2024barlow} encoders (the
backbone family of \tesseravone),\footnote{The sweep ran on $\ngpus$
NVIDIA GH200 superchips~\citep{mcintosh2024isambard}, each pairing a
Grace CPU with one H100 GPU ($96$\,GB HBM3).} each evaluated on
$\nscalingtasks$ diverse downstream tasks. The loss proved a weak proxy
for downstream utility, and selecting models by it alone needs roughly
$\wastemult$ the compute to match downstream-driven selection (Finding
\textbf{F1}); the downstream-optimal allocation instead increases
encoder capacity and training data together while the projector stays
essentially fixed (Finding \textbf{F2}).

Scaling the encoder this way inflates the inference cost that
dominates an embeddings-as-data product. We recover deployment
efficiency through
\textit{distillation}~\citep{hinton2015distillation}, yielding a
simple production rule: train one large teacher by the fitted rule
rather than by repeated loss-driven search, then distil it into
compact students. This buys adaptivity along two axes: a one-off sweep
over student sizes (N/S/M/L) lets users who generate their own
embeddings trade compute for quality, and \matryoshka{} prefixes
expose $16$-, $32$-, $64$-, and $128$-dimensional views of one
embedding for users who consume embeddings as data.
\Cref{sec:pixel:why-not-bt} explains why this nesting must be learned
through distillation rather than by prefix losses during
self-supervised pretraining.

The resulting large encoder allows \tesseravtwo{} to achieve SOTA
performance across a diverse set of tasks. Across a $\nfullsuite$-task
suite (the $\npixeldata$-task \aebench{} plus $\nheldout$ further
held-out datasets), the distilled students outperform every open and
proprietary embedding product we compare, even the smallest prefixes
keep most of the full-dimensional score, and the embeddings largely
suppress the Sentinel acquisition artefacts visible in v1
(\Cref{fig:teaser}d).

In summary, our contributions are:
\begin{enumerate}[leftmargin=*,itemsep=2pt,topsep=2pt]
  \item \textbf{Downstream-driven scaling for product-grade EO
        embeddings.} Across $\nruns$ runs evaluated on
        $\nscalingtasks$ downstream tasks, the pretraining loss is a
        poor selection target (\textbf{F1}), and the fitted
        allocation rule sends additional budget to encoder capacity
        and data, not projector size (\textbf{F2}).
  \item \textbf{A deployable pixel-wise embedding product family:
        \tesseravtwo{}.} Guided by this rule, we train large teachers
        and distil them into compact N/S/M/L students, served as
        analysis-ready data through \geotessera{}, with the best
        composite score and mean rank on the $\npixeldata$-task
        \aebench{} at a deployment cost two orders of magnitude below
        the teacher's.
  \item \textbf{Storage-adaptive \matryoshka{} embeddings.} Each
        student exposes prefixes at $d\in\dimset$ from one embedding,
        an accuracy/storage knob needing no retraining: $d{=}16$
        keeps ${\sim}92\%$ of the $d{=}128$ score at
        $\nicefrac{1}{8}$ of the storage.
\end{enumerate}

%% file: sections/02_related_work.tex
\section{Related work}
\label{sec:related}
The present work builds on \tesseravone~\citep{feng2026tessera},
which adapts \barlowtwins{} redundancy reduction to cloud-corrupted
EO time series following \citet{lisaius2024barlow}. We reuse its
\dpixel{} formulation and pretraining recipe as the fixed model
family for the scaling study.

\paragraph{EO foundation models.} Most remote-sensing foundation
models pretrain \emph{spatial} backbones on single-time, often
cloud-filtered patches with contrastive or masked-image
objectives~\citep{manas2021seco,cong2022satmae,reed2023scalemae,%
guo2024skysense}, with multi-sensor and multi-resolution variants
fusing optical and SAR or ingesting many sensors at
once~\citep{fuller2023croma,xiong2024dofa,astruc2025anysat,%
tseng2025galileo}. \presto~\citep{tseng2024presto} instead processes
per-pixel time series, and \textsc{MOSAIKS}~\citep{rolf2021mosaiks}
explored lightweight universal features; the full catalogue is in
\Cref{app:extended-related}. Two assumptions recur across these
designs: natural-image scaling intuition is inherited by analogy, and
the model is deployed through per-task fine-tuning of the backbone,
which is compute- and label-intensive for EO users. We are not aware
of a controlled study of where additional compute should go given
irregular revisits, cloud occlusion, and label scarcity.
\Cref{sec:scaling} provides one for the pixel-wise \barlowtwins{}
family.

\paragraph{Embedding products.} A second line publishes precomputed,
analysis-ready embeddings instead of a backbone: \tesseravone{}
releases global annual $10$\,m pixel-wise int8 embeddings with the
\geotessera{} retrieval
library~\citep{feng2026tessera,ucam-eo-geotessera}, \alphaearth{}
provides global annual $10$\,m embedding fields from many
instruments~\citep{brown2025alphaearth},
\textsc{ESD}~\citep{chen2026esd} compresses $25$ years of
Landsat/MODIS reflectance into quantized $30$\,m embeddings, and
\textsc{OlmoEarth}~\citep{herzog2026olmoearth} computes embeddings on
demand, allowing custom time windows at the cost of per-request
inference. Their shared limitation is a fixed embedding specification:
one dimension and one storage budget for every user, with no
coordinate ordering that would let a user truncate without retraining
or loss. \tesseravtwo{} keeps the
paradigm and adds two degrees of freedom: students at four sizes,
produced by scaling-law-guided distillation, and \matryoshka{}
prefixes $d\in\dimset$ from a single embedding without retraining.
Evaluating embeddings across such budgets has begun to attract
dedicated benchmarks~\citep{vinge2025neucobench}.

\paragraph{Scaling laws.} Empirical scaling laws have shaped recent
language and vision
work~\citep{kaplan2020scaling,hoffmann2022chinchilla,zhai2022scaling},
and are typically fit to a self-supervised pretraining loss on the
assumption that the loss proxies downstream quality. This assumption
can fail: in language modelling the mapping from pretraining loss to
downstream performance is sometimes noisy or non-monotone, so a lower
loss need not yield a better task model~\citep{hu2025scalingunreliable}.
The two EO studies we are aware of do not test it directly.
\citet{dionelis2025scaling} sweep architecture, size, and data on
PhilEO Bench~\citep{fibaek2024phileo} but entangle architecture with
capacity over a coarse grid, and \citet{wickrema2025scaling} fit
peta-pixel power laws to the validation loss in a data-limited regime
that, as they note, is confounded by under-trained large models, and
defer the loss-vs-downstream comparison to future work. Neither
isolates encoder, projector, and data under matched compute. We
isolate all three and fit the compute allocation against task
performance directly, evaluating every run on \nscalingtasks{}
downstream tasks.

\paragraph{Distillation and nested embeddings.} Knowledge
distillation~\citep{hinton2015distillation} transfers representations
from a high-capacity teacher to a compact student, and \matryoshka{}
representation learning~\citep{kusupati2022matryoshka} produces nested
embeddings whose prefixes work at multiple dimensionalities. The two
are rarely combined in EO, and rarely with attention to whether the
self-supervised objective can support nested coordinates at all: we
show it cannot, because redundancy-reduction objectives identify
subspaces only up to rotation, whereas distillation against a fixed
teacher supplies the ordering signal that self-supervision lacks
(\Cref{sec:pixel:why-not-bt}).

%% file: sections/03_scaling_laws.tex
\section{Downstream-driven scaling laws}
\label{sec:scaling}
This section answers the allocation question: within a fixed
architectural family, how should pretraining compute be split
between encoder size, projector size, and training data? The
experiments below describe the family we sweep, pixel-wise
\sonetwo{} encoders pretrained with
\barlowtwins~\citep{zbontar2021barlow}. Note that we do \textbf{not} advance them as
universal EO scaling laws (please refer to the discussion in
\Cref{sec:scaling:scope}).

\subsection{Study design}
\label{sec:scaling:design}
\paragraph{Architecture choice.}
\label{sec:scaling:arch}
Attributing performance differences to size or data is only
meaningful when the architecture is fixed. We therefore select it
first, sweeping ten structural axes one at a time by aggregated
downstream score (\Cref{app:arch-sweep}), and hold the selected
configuration fixed for the entire scaling sweep.

\paragraph{Controlled sweep.} With architecture fixed, we pretrain
$\nruns$ models on $\ngpus$ \gputype{} GPUs in an iso-FLOP-style
grid over encoder size $N_{\mathrm{enc}}$ ($16$ widths,
$7$--$278$\,M), projector size $N_{\mathrm{proj}}$ (four widths),
and training data $D$ ($0.03$--$3{,}202$\,M \dpixels).
At each compute level, the compute-optimal size is the vertex of a
quadratic fit in $\log N$ (grid details in
\Cref{app:scaling-setup}). The compute axis is the total training FLOPs,
\begin{equation}
C \;=\; 18\, D \,
        \bigl(N_{\mathrm{enc}}\,\bar{L} + N_{\mathrm{fus}}\,M
              + N_{\mathrm{red}} + N_{\mathrm{proj}}\bigr)
        \;+\; 6\,D\,P^{2},
\qquad \bar{L} = 12,
\label{eq:flops-main}
\end{equation}
where $18 = 6\times 3$ is the textbook $6ND$ factor times the
\emph{three} encoder passes per step (the two \barlowtwins{}
augmentation views and the mixup view, the fourth shuffled view being
a permutation that needs no pass~\citep{bandara2023mixup});
$\bar{L}=12$ is the mean per-view token length actually processed
(the augmentation samples $L\!\in\!\seqlen$ each step);
$N_{\mathrm{fus}}$ and $N_{\mathrm{red}}$ are the modality-fusion and
dimensionality-reduction costs, $N_{\mathrm{proj}}=5P^{2}$ the
projector, and $6DP^{2}$ the \barlowtwins{} cross-correlation cost.
The per-token derivation, the omitted $\mathcal{O}(DL^{2})$
self-attention term (${<}1\%$ of $C$), and a
\texttt{FlopCounterMode}~\citep{torch_flop_counter} check are in
\Cref{app:flops}.

\begin{figure}[!htbp]
  \centering
  \includegraphics[width=\linewidth]{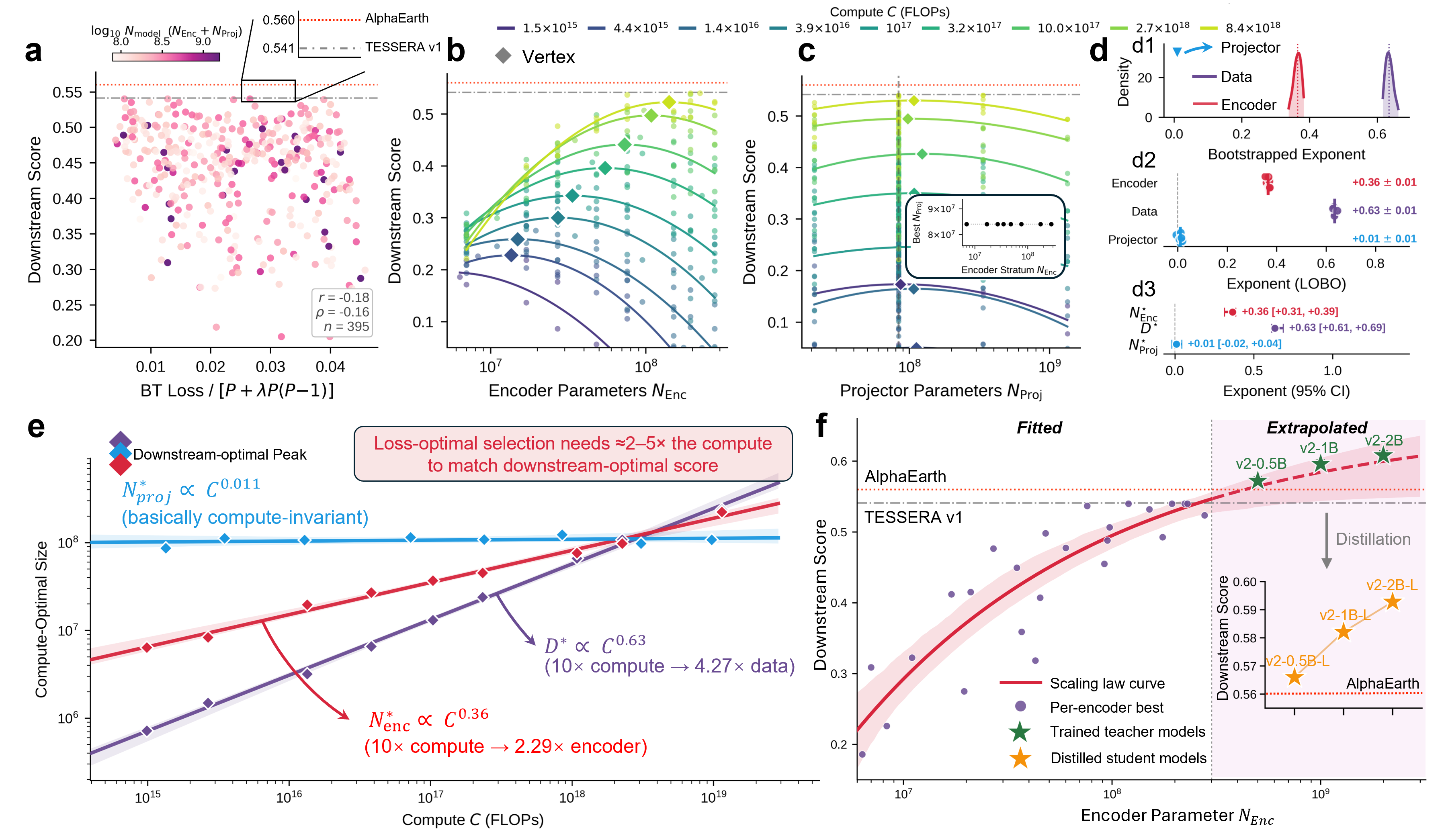}
  \caption{Downstream-driven scaling laws ($\nruns$ runs; composite
  score over $\nscalingtasks$ tasks).
  \textbf{(a)}~Pretraining loss vs.\ downstream score, one point per
  run, coloured by model size ($r\!=\!-0.18$).
  \textbf{(b,\,c)}~Iso-FLOP parabolas for encoder and projector;
  diamonds mark the compute-optimal size (parabola vertex) per compute
  bucket. Inset in~(c): the best $N_{\mathrm{proj}}$ is flat across
  encoder strata.
  \textbf{(d1--d3)}~Bootstrap density, leave-one-bucket-out, and
  $95\%$ CIs of the three exponents.
  \textbf{(e)}~Power-law fits to the vertices,
  $N_{\mathrm{enc}}^{\star}\!\propto\!C^{\encexp}$,
  $D^{\star}\!\propto\!C^{\dataexp}$,
  $N_{\mathrm{proj}}^{\star}\!\propto\!C^{\projexp}$.
  \textbf{(f)}~Out-of-sample validation: the encoder law fit on the
  sweep (solid, ${\le}278$\,M) extrapolates (dashed band) to the
  $0.5$/$1$/$2$\,B teachers (stars), none used in the fit. Inset:
  distilled students.}
  \label{fig:scaling-laws}
\end{figure}

\paragraph{Downstream evaluation.} Every pretrained model is
evaluated on $\nscalingtasks$ \aebench{} tasks drawn from $10$
source datasets (classification, change detection, regression), with
chance-adjusted metrics averaged into one composite downstream score,
the $y$-axis of \Cref{fig:scaling-laws}
(\Cref{app:datasets,app:eval-protocol}). On the same suite
\tesseravone{} scores $0.541$ and \alphaearth{} $0.560$; both appear
as baselines in \Cref{fig:scaling-laws}a--c. Since these are
full-budget production systems while the grid deliberately spans
small, data-limited configurations to trace the compute frontier, it
is the upper envelope of the sweep, not the individual runs, that
approaches them. For every run we also record the converged
\barlowtwins{} loss on held-out \dpixels{}, normalised across
projector widths (\Cref{app:scaling-extra:selection}).

\subsection{Finding 1: the pretraining loss is not a good predictor of downstream task performance}
\label{sec:scaling:loss}
Across the $\nruns$ runs, the converged \barlowtwins{} loss and the
composite downstream score are nearly independent
(\Cref{fig:scaling-laws}a; Pearson $r\!=\!-0.18$, Spearman
$\rho\!=\!-0.16$); \Cref{app:scaling-extra:why-weak} discusses why.
The consequence is quantitative, not just statistical: fitting separate
power laws through loss-selected and score-selected bucket peaks
(\Cref{app:scaling-extra:selection}), loss-based selection needs
roughly $\wastemult$ the compute to reach the same downstream
score. EO scaling laws must therefore be fit against downstream
metrics.

\subsection{Finding 2: encoder size and data requirements scale with compute; the
projector does not}
\label{sec:scaling:asymmetry}
We group runs into \nbuckets{} iso-FLOP buckets and fit, within
each, a quadratic in $\log N$ whose vertex is the compute-optimal
size at that budget. Encoder vertices shift right as compute grows
(\Cref{fig:scaling-laws}b), while projector vertices stack along a
vertical line (\Cref{fig:scaling-laws}c). Power-law fits through
the vertices (\Cref{fig:scaling-laws}e) give
$N_{\mathrm{enc}}^{\star}\!\propto\!C^{\encexp}$
($95\%$ CI $[+0.31,+0.39]$),
$D^{\star}\!\propto\!C^{\dataexp}$ ($[+0.61,+0.69]$), and
$N_{\mathrm{proj}}^{\star}\!\propto\!C^{\projexp}$
($[-0.02,+0.04]$). Full fit statistics are in
\Cref{app:scaling-extra:exponents}. The projector exponent is
statistically indistinguishable from zero: the compute-optimal
projector size does not grow with compute. Encoder capacity, not
projector capacity, is the load-bearing scaling axis.

\paragraph{Deriving a recipe for scaling.}
\label{sec:scaling:from-finding-to-recipe}
Since encoder capacity and data absorb compute while the projector is
compute-invariant, the compute-optimal use of a large budget is one
oversized encoder trained on correspondingly more data, with the
projector held near its optimum as a disposable training scaffold;
deployment flexibility, in model size and embedding dimension, is then
recovered through distillation (\Cref{sec:teacher,sec:pixel}).

\paragraph{The fitted law predicts unseen scales.} A downstream scaling
law is only useful if it holds past the runs it was fit on. We test this
out of sample in \Cref{fig:scaling-laws}f: the encoder law, fit on the
sweep alone (encoders up to $278$\,M), extrapolates to the $0.5$, $1$,
and $2$\,B teachers of \Cref{sec:teacher}, none of which entered the fit.
All three land on the predicted curve and improve monotonically with
scale, from $0.572$ to $0.608$, clearing both \tesseravone{} ($0.541$)
and \alphaearth{} ($0.560$); distilling each into its deployable
student costs at most $0.015$ of composite score, and the students keep
improving with scale too (\Cref{fig:scaling-laws}f, inset). The law
therefore predicts rather than merely describes.

%% file: sections/04_train_large_teacher.tex
\section{A pixel-wise temporal teacher}
\label{sec:teacher}
\begin{figure}[!htbp]
  \centering
  \includegraphics[width=\linewidth]{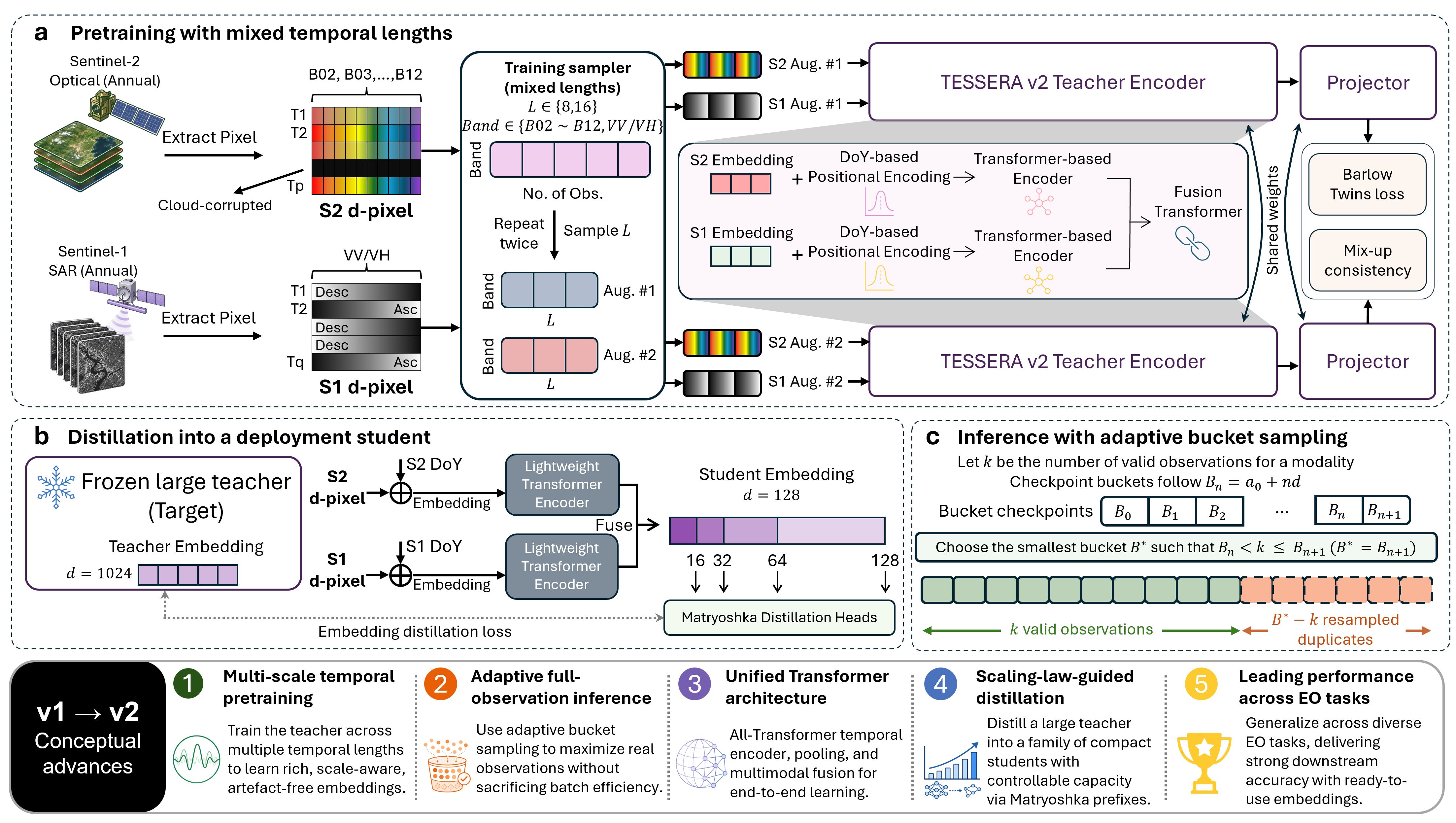}
  \caption{\tesseravtwo{} architecture.
  \textbf{(a)}~Pretraining: two views per \dpixel{} at random
  length $L\!\in\!\{8,16\}$, per-modality Transformers with
  day-of-year encoding, cross-modal fusion, and
  \barlowtwins{}~$+$~mix-up.
  \textbf{(b)}~Distillation: \matryoshka{} prefix heads at
  $d\!\in\!\dimset$ reconstruct the frozen teacher embedding.
  \textbf{(c)}~Inference: each pixel's $k$ valid observations are
  packed into the smallest bucket $B^{\star}\!\geq\!k$, with
  residual slots filled by midpoint resampling.}
  \label{fig:architecture}
\end{figure}

\Cref{fig:architecture} summarises the design of the single large
encoder. Relative to \tesseravone~\citep{feng2026tessera},
\tesseravtwo{} adds multi-scale temporal pretraining, adaptive
full-observation inference, a unified all-Transformer architecture
with cross-modal fusion, and scaling-law-guided distillation into
\matryoshka{} students (\Cref{sec:pixel}).

\paragraph{Inputs.}
\label{sec:teacher:dpixel}
A \dpixel{} at location $(i,j)$ is the time series of all \sone{} and
\stwo{} observations at that $10\,\mathrm{m}$ pixel over one year,
with a binary mask marking valid timesteps~\citep{feng2026tessera},
preserving the full temporal phenology while tolerating cloud
occlusion and irregular revisits.

\paragraph{Architecture.}
\label{sec:teacher:arch}
The teacher is a $\teacherparams$ dual-branch pixel-wise encoder
(\Cref{fig:architecture}a). Each modality branch embeds its valid
observations, adds a sinusoidal day-of-year positional encoding, runs
a per-modality Transformer, and aggregates the variable-length
sequence by learned attention pooling; a fusion Transformer then
combines the two modality tokens into one embedding
$\bt \in \R^{\dimteacher}$, $\dimteacher=1024$.
$\dimteacher$ deliberately exceeds the student's $\dimfull$, leaving
room for compression. The encoder is larger than any point in the \Cref{sec:scaling} sweep
(which tops out at $278$\,M); following the recipe, we extrapolate its
$C^{\encexp}$ encoder law to the $\teacherparams$ teacher, whose
training compute is measured on the same axis (\Cref{eq:flops-main}).
Layer, head, and projector details are in \Cref{app:arch}.

\paragraph{Objective and training.}
\label{sec:teacher:obj}
Similar to \tesseravone, we train using \barlowtwins{} over two temporally
subsampled views, with an additional mix-up consistency
regulariser~\citep{bandara2023mixup} and global shuffling of
\dpixels{} across tiles.
Unlike v1's single fixed sample length, at every step the view
length is a random $L\!\sim\!\mathrm{Uniform}\{8,16\}$, so each view is
a sparse random subsample of the year that forces the encoder to
recover annual phenology from few observations. At inference we instead pass
\emph{all} valid observations (\Cref{sec:teacher:bucket}); because
they are placed at their true day-of-year and pooled
length-agnostically, this is more evidence about the same phenology,
not a longer out-of-distribution input. Whole modalities are also
dropped with small probability, which doubles as the training signal
for a pixel with no valid observations in one modality.
The teacher is pretrained for a single epoch over ${\sim}14$ billion
\dpixels{}. Loss equations, the mix-up derivation, the geographic
distribution of the source patches, and systems details are in
\Cref{app:teacher-objective,app:mixup,app:training:teacher}.

\paragraph{The teacher as a distillation target.}
\label{sec:teacher:dist}
Most users cannot run a $\teacherparams$ pixel-wise encoder over
global \sonetwo{} even once, let alone recurrently. The scaling laws
say \emph{where} to spend pretraining compute, but what users can
afford to \emph{serve} is set by the recurring cost of global
inference. The teacher is therefore the right model to train but the
wrong one to deploy, so we treat the frozen teacher
$\teacher_{\theta}: P_{i,j} \mapsto \bt$ as a fixed
\emph{representation distribution} and obtain every deployed artifact
by distilling against it (\Cref{sec:pixel}).

\paragraph{Adaptive bucket sampling at inference.}
\label{sec:teacher:bucket}
The number of valid observations $k$ varies from a handful in
heavily clouded regions to roughly a hundred in clean ones, yet
batched inference wants fixed-length inputs. Where \tesseravone{}
sampled a fixed $L\!=\!40$ timesteps, discarding or duplicating
observations, \tesseravtwo{} packs each pixel into the smallest bucket
from the ladder $\{16,32,48,\ldots\}$ that fits all $k$ observations,
filling residual slots by midpoint resampling
(\Cref{fig:architecture}c). No observation is discarded, and a batch
still partitions into a few fixed-length groups that run in parallel;
construction and pseudocode are in
\Cref{app:bucket-construction,app:bucket}.

%% file: sections/05_pixel_distillation.tex
\section{Distilling a deployable student family}
\label{sec:pixel}
We expect users to consume our \emph{embeddings as data}: precomputed
annual pixel embeddings with lightweight task heads, without running a
backbone~\citep{feng2026tessera,brown2025alphaearth}. The teacher
cannot support this directly. Inference cost for one global, annual,
$10$\,m \sonetwo{} pass grows near-linearly with encoder parameters,
so the $\teacherparams$ teacher costs roughly $100$ \gputype-years per
global pass against $0.04$ to $2$ for the students, two orders of
magnitude lower (\Cref{app:deploy-cost}). Storage and download cost,
meanwhile, scale with embedding dimension, which is what the
\matryoshka{} prefixes address.

\paragraph{Student family.}
\label{sec:pixel:family}
We distil the teacher into four students of the same architectural
form, differing only in backbone width and depth and each emitting a
$\dimfull$-dimensional embedding: \tesseravtwo-2B-L ($\studentL$, for
provider-side global inference), M ($\studentM$, a balanced default),
S ($\studentS$, low-resource), and N ($\studentN$, edge and
on-device). Distillation uses $1.2$ billion \dpixels{} on $64$
\gputype{} GPUs
(\Cref{app:arch,app:training:distill,app:distill-scaling}).

\paragraph{\matryoshka{} distillation objective.}
\label{sec:pixel:objective}
For each prefix length $k \in \mathcal{K} = \dimset$ we attach an
MLP head $h_{k}: \R^{k}\to\R^{\dimteacher}$, used only during
distillation, and train the student against the frozen teacher
embedding $\bt$ with
\begin{equation}
\Ldist(\bs,\bt)
  \;=\; \sum_{k\in\mathcal{K}}
        \Bigl(1 - \cos\!\bigl(h_{k}(\bs_{1:k}),\; \bt\bigr)\Bigr).
\label{eq:matryoshka-distill}
\end{equation}
Each prefix reconstructs the \emph{full} teacher embedding rather than
a copy of its first $k$ coordinates, making the prefix a
rate--distortion code for $\bt$; at inference the heads are discarded
and the user takes any prefix $\bs_{1:k}$ directly. The frozen
teacher's targets are cached, so compute is spent almost entirely on
the student.

\paragraph{Nesting through distillation, not pretraining.}
The obvious alternative, adding prefix-\barlowtwins{} losses during
self-supervised pretraining, unexpectedly fails: the
redundancy-reduction objective identifies a representation subspace
only up to rotation, so prefix losses break the coordinate symmetry
through gradient imbalance rather than through a signal about axes'
information content. Distillation against a fixed target supplies that
signal. The formal statement, the empirical fingerprints, and a
coordinate-exchangeability ablation are in
\Cref{sec:pixel:why-not-bt,sec:analysis:matryoshka-fails}.

\subsection{Benchmark results}
\label{sec:pixel:bench}
We evaluate in two stages. First, on the $\npixeldata$-task
\aebench{} (classification, regression, and change detection;
\Cref{app:datasets}) we compare against $20$ embedding products,
remote-sensing foundation models, and generic backbones, among them
the two directly comparable embedding-as-data systems
\alphaearth~\citep{brown2025alphaearth} and
\tesseravone~\citep{feng2026tessera}, plus
OlmoEarth~\citep{herzog2026olmoearth} (full list in
\Cref{app:eval-protocol}). Second, we take the four strongest of these
and run a \emph{held-out} evaluation on $\nheldout$ further datasets
that played no part in development, covering classification,
segmentation, regression, and distribution prediction
(\Cref{app:heldout}); the two stages together form a
$\nfullsuite$-task full suite. Scoring follows \alphaearth{}, using
nearest-neighbour classifiers and linear probes, plus a small
($<\!2$M-parameter) CNN on the held-out suite where spatial context
matters (\Cref{app:eval-protocol}). As with all foundation models,
downstream performance is sensitive to the choice of task head, so our
numbers may differ from published results using a different head
(\Cref{app:cityrep-linear}).

\begin{figure}[!htbp]
  \centering
  \includegraphics[width=\linewidth]{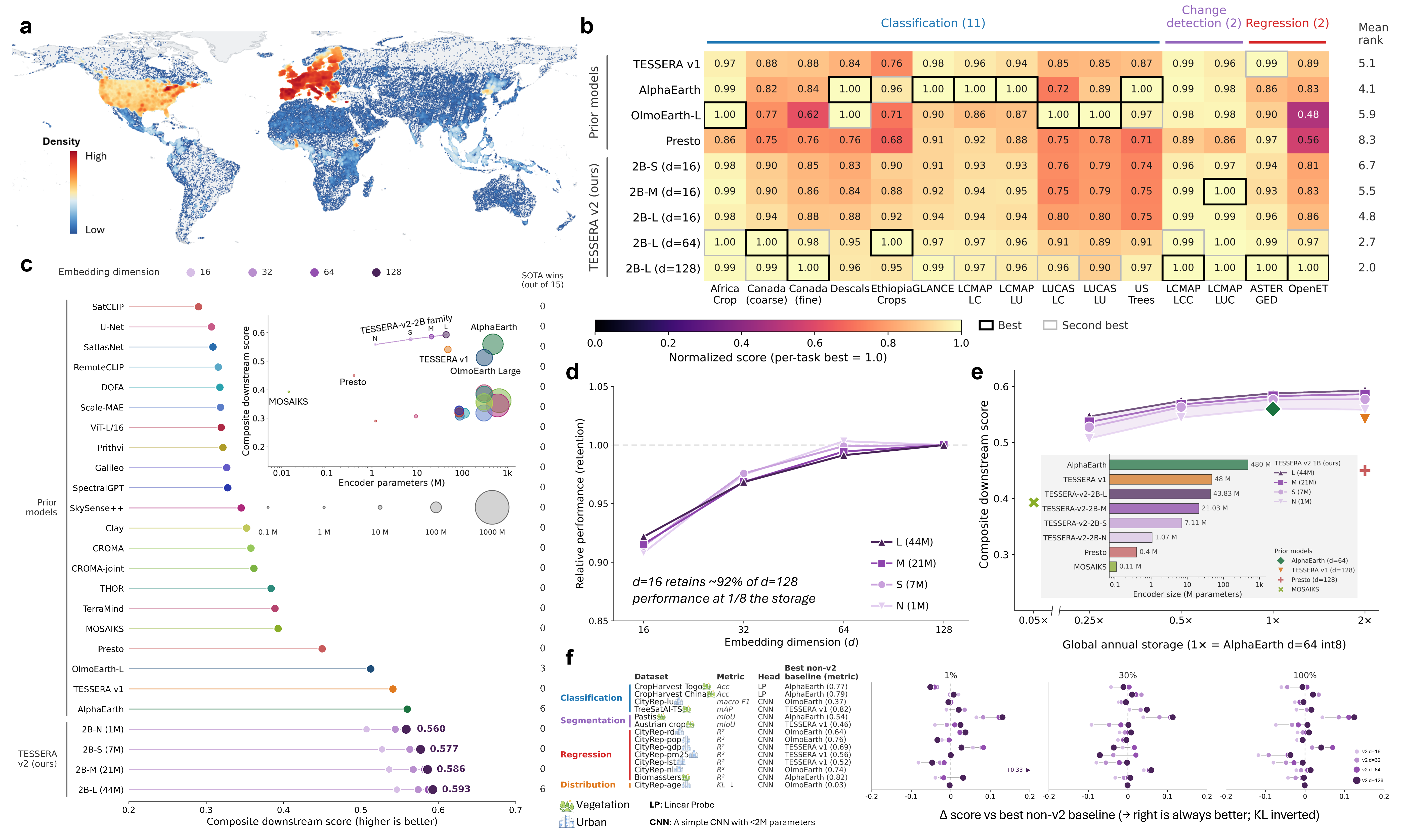}
  \caption{\aebench{} results ($\npixeldata$ tasks) and held-out
  generalisation ($\nheldout$ datasets).
  \textbf{(a)}~Geographic density of all downstream labels.
  \textbf{(b)}~Per-task heatmap with mean rank.
  \textbf{(c)}~Composite score per model; the four markers per
  \tesseravtwo{} row are $d\!\in\!\dimset$.
  \textbf{(d)}~Score vs.\ prefix dimension.
  \textbf{(e)}~Score vs.\ global annual storage, relative to
  \alphaearth{} at $d{=}64$ int8.
  \textbf{(f)}~Label efficiency at $1/30/100\%$ of the labels:
  $\Delta$ score vs.\ the best non-v2 baseline.}
  \label{fig:pixel-bench}
\end{figure}

\paragraph{Headline results.} The $\nheldout$ held-out datasets are
our primary evaluation: they played no part in developing our system,
whereas the \aebench{} labels guided v2 model and hyperparameter
selection and contributed training data to some baselines.
On the held-out suite \tesseravtwo-2B-L has the best composite of any
system, $0.647$, against OlmoEarth-L and \tesseravone{} (both $0.614$)
and \alphaearth{} ($0.590$), using a $\studentL$ encoder an order of
magnitude smaller than \alphaearth{} ($480$\,M) or OlmoEarth-L
($308$\,M) (\Cref{fig:teaser}f).
On the $\npixeldata$ \aebench{} tasks the composite
score is $0.593$ (M, S, N at $0.586$, $0.577$, $0.560$;
\Cref{fig:pixel-bench}c), exceeding \alphaearth{} ($0.560$), \tesseravone{}
($0.541$), and OlmoEarth-L ($0.512$), with the best mean rank ($2.0$ at
$d{=}128$ vs.\ $4.1$ and $5.1$; \Cref{fig:pixel-bench}b).
\alphaearth{} came second on the shared tasks but last of the four on
the held-out ones; this gap is why we treat the held-out suite as
primary (\Cref{app:heldout-robust}).

\paragraph{Graceful degradation across students.} Across the four
students, $d{=}16$ performs at ${\sim}92\%$ of the $d{=}128$ composite
at $\nicefrac{1}{8}$ of the storage, with $d{=}32$ and $d{=}64$ at
${\sim}97\%$ and ${\sim}99\%$ (\Cref{fig:pixel-bench}d); performance is
best on land-cover and change-detection tasks and worst on
fine-class-count and regression, for which $d{=}64$ is the default.
Every student is on the Pareto frontier of score versus storage and
encoder parameters (\Cref{fig:pixel-bench}e): at half the storage of
\alphaearth{}, \tesseravtwo-2B-L already exceeds it ($0.574$ vs.\
$0.560$; per-prefix composites in \Cref{sec:analysis:storage}).

\paragraph{Held-out generalisation.} The held-out datasets split into
a vegetation group (tree species, crop and parcel segmentation,
biomass) and an urban group (land use, road density, population, GDP,
nighttime lights, pollution, surface temperature, demographics).
Across $1/30/100\%$ label budgets \tesseravtwo{} beats the best non-v2
baseline on most of them, with the widest margins in the low-label
regime (\Cref{fig:pixel-bench}f; construction and scoring in
\Cref{app:heldout}).

\paragraph{Embedding perceptual quality and temporal stability.}
\label{sec:pixel:artefacts}
Relative to v1, whose embeddings show along-track striping and
tile-seam discontinuities aligned with \stwo{}/\sone{} acquisition
geometry, v2 has far fewer artefacts while preserving geographic
structure, markedly lower consecutive-year cosine distances where land
cover is stable, and a return to its prior baseline after a temporary
disturbance where v1 stays noisy (\Cref{fig:teaser}d; full analysis in
\Cref{app:artefacts}).

\paragraph{Ablations.} Coordinate-exchangeability, reconstruction, and
teacher-size ablations support the distillation design: only distilled
prefixes carry a genuine semantic ordering of the embedding
coordinates, and the student prefix is a better code for the teacher
embedding than the teacher's own coordinate prefix
(\Cref{sec:analysis}).

%% file: sections/06_discussion.tex
\section{Discussion and Conclusion}
\label{sec:discussion}
In language and vision, the pretraining loss tracks downstream quality
closely enough to fit scaling laws directly to it. EO breaks this
assumption: selecting models by the loss needs roughly $\wastemult$
the compute to reach the same downstream score (\textbf{F1}), so
future EO scaling studies should budget for downstream evaluation
rather than trust in the loss. Once selection is downstream-driven,
\textbf{F2} gives the allocation rule we use: spend the budget on
encoder capacity and matched data, hold the projector fixed as a
disposable training scaffold, and recover deployable encoders by
distillation. Treating the teacher as a frozen representation
distribution then lets one pretraining run amortise across four
student sizes and, within each, four embedding dimensions.
Distillation, not pretraining, is also what makes that dimension a
usable control knob: self-supervision fixes only the subspace the
embedding spans, leaving its coordinates unordered, whereas
distillation against the frozen teacher orders them
(\Cref{sec:pixel:why-not-bt}).

\paragraph{Limitations.}
\label{sec:disc:limits}
The scaling laws are empirical and apply only to pixel-wise
\sonetwo{} encoders, one self-supervised objective, and one
$\nscalingtasks$-task evaluation suite (\Cref{sec:scaling:scope}), so
our analysis of budget allocation is specific to that setting. Second,
the use of an expensive teacher model requires substantial
computation. Finally, our benchmarks are drawn from well-studied
regions, so generalisation to under-represented climates and unseen
seasons remains to be evaluated. Broader-impact and release
considerations are in \Cref{sec:disc:impact}.

\paragraph{Conclusion.}
\label{sec:conclusion}
We have presented \tesseravtwo{}: a $\nruns$-run downstream-driven
scaling study, a $\teacherparams$ pixel-wise teacher trained by the
resulting allocation rule, and a distilled \matryoshka{} student
family that leads both the $\npixeldata$-task \aebench{} and a
$\nheldout$-dataset held-out suite at two orders of magnitude lower
inference cost. We hope that the train-large-then-distil recipe, and
the finding that nested embeddings must be learned by distillation
rather than by naive self-supervised prefix losses, carry over to
other EO embedding products.

%% file: sections/appendix.tex

\section{Design Desiderata and Extended Related Work}
\label{app:desiderata}
This appendix expands two parts of the main text that are stated
compactly there: the five product criteria of \Cref{sec:intro} applied
system by system, and the full catalogue of EO foundation models
summarised in \Cref{sec:related}.

\paragraph{The five criteria.} An embedding product is
\emph{analysis-ready} if a user can treat it as a geospatial data
layer, with no raw-imagery processing and no GPU inference on their
side; \emph{transferrable} if one representation serves
classification, regression, change detection, and high-resolution
mapping across regions and workflows, rather than being tuned to a
single benchmark~\citep{klemmer2025earth,lyu2026structure};
\emph{reproducible} if the algorithms, training procedure, and weights
can be inspected, compared, and extended rather than only their
outputs downloaded; \emph{economical} in the resources that actually
limit EO adoption, namely labels, training and inference compute (and
the associated carbon footprint), storage, I/O, and engineering time;
and \emph{adaptive} if it characterises the accuracy-versus-cost
trade-off explicitly, so that users are not locked into one model size
or one embedding dimensionality.

\Cref{tab:desiderata} scores the systems discussed in
\Cref{sec:intro,sec:related} against these criteria. Remote-sensing
foundation models are reproducible and transferrable but not
analysis-ready, since the user must run the backbone.
\alphaearth{}~\citep{brown2025alphaearth} is analysis-ready at global
scale but open-output only: the embeddings are public while the
training procedure and weights are
not~\citep{hou2026alphaearth}. \textsc{ESD}~\citep{chen2026esd} is
analysis-ready and open, but at $30$\,m from Landsat/MODIS rather than
$10$\,m from \sonetwo{}. \textsc{OlmoEarth}~\citep{herzog2026olmoearth}
is flexible in time window but computes embeddings per request, so it
is not a precomputed archive. \tesseravone~\citep{feng2026tessera}
meets every criterion except adaptivity: one model, one dimension, one
storage budget. \tesseravtwo{} adds both missing degrees of freedom.

\begin{table}[!htbp]
  \centering
  \caption{Embedding products and foundation models against the five
  criteria of \Cref{sec:intro}. \checkmark~= satisfied,
  $\sim$~= partially, $\times$~= not.}
  \label{tab:desiderata}
  \small
  \setlength{\tabcolsep}{5pt}
  \begin{tabular}{lccccc}
  \toprule
  System & Analysis- & Transfer- & Repro- & Econo- & Adaptive \\
         & ready     & rable     & ducible & mical  &          \\
  \midrule
  RSFMs (fine-tuned)             & $\times$   & \checkmark & \checkmark & $\times$   & $\times$ \\
  \textsc{MOSAIKS}               & $\sim$     & $\sim$     & \checkmark & \checkmark & $\times$ \\
  \textsc{ESD}                   & \checkmark & $\sim$     & \checkmark & \checkmark & $\times$ \\
  \textsc{OlmoEarth}             & $\sim$     & \checkmark & \checkmark & $\sim$     & $\times$ \\
  \alphaearth                    & \checkmark & \checkmark & $\times$   & \checkmark & $\times$ \\
  \tesseravone                   & \checkmark & \checkmark & \checkmark & \checkmark & $\times$ \\
  \textbf{\tesseravtwo{} (ours)} & \checkmark & \checkmark & \checkmark & \checkmark & \checkmark \\
  \bottomrule
  \end{tabular}
\end{table}

\paragraph{Extended catalogue of EO foundation models.}
\label{app:extended-related}
\Cref{sec:related} cites a representative subset. The full set of
spatial-backbone models pretrained on single-time, often cloud-filtered
patches with contrastive or masked-image objectives
includes~\citep{manas2021seco,guo2024skysense,cong2022satmae,%
reed2023scalemae,tang2023crossscalemae,li2024s2mae,noman2024satmaepp,%
wang2022dinomm,wanyan2024dinomc}, which have grown into broader
families~\citep{sun2023ringmo,wang2023ringmolite,bastani2023satlas,%
schmude2024prithvi,szwarcman2025prithvieo2,mendieta2023gfm,%
han2024msgfm,wu2025semantic,luo2024skysensegpt,zhu2025skysenseo,%
zhang2025skysensev2}. Multi-sensor and multi-resolution models fuse
optical and SAR or ingest many sensors at
once~\citep{fuller2023croma,yao2023ringmosense,xiong2024dofa,%
astruc2025anysat,perron2026universat,houdre2026ramen,tseng2025galileo}.

\paragraph{Dataset frameworks and embedding archives.} Beyond the
models above, \textsc{Major TOM}~\citep{francis2024majortom}
standardises EO datasets through a global grid index and a shared
metadata schema, releasing the near-global MajorTOM-Core Sentinel
corpus; its embedding expansion~\citep{czerkawski2024majortom} adds
dense patch-level embeddings extracted from pretrained backbones over
that grid, across \stwo{} L1C/L2A and \sone{} RTC. As with the
embedding products in \Cref{sec:related}, each ships at a single fixed
embedding specification per source.

\section{Training-FLOPs Derivation}
\label{app:flops}
This appendix derives the training-FLOPs accounting that
underlies the compute axis \Cref{eq:flops-main}, adopting the
standard convention that one fused multiply--add corresponds to two
FLOPs~\citep{kaplan2020scaling,hoffmann2022chinchilla}. The
scaling-study axis of \Cref{eq:flops-main} is exactly this cost,
evaluated at the mean per-view length $\bar{L}=12$; the two
coincide up to the $\mathcal{O}(DL^{2})$ self-attention residual
(${<}1\%$; \Cref{app:flops:attn}), as collected in
\Cref{app:flops:final}.

\subsection{Notation}
\label{app:flops:notation}
\begin{itemize}[leftmargin=*,itemsep=2pt,topsep=2pt]
  \item $D$: total number of \dpixels{} processed during training.
  \item $L$: per-modality temporal sequence length (number of tokens
        consumed by the temporal encoder per \dpixel{}).
  \item $N_{\mathrm{enc}}$: total parameter count of all temporal
        encoder backbones, summed across modalities.
  \item $N_{\mathrm{fus}}$: parameter count of the modality-fusion
        attention that combines the per-modality summary tokens
        into one joint representation.
  \item $M$: number of modality branches fused per \dpixel{}
        ($M\!\in\!\{2,3\}$ across our configs---e.g.\ S2\,+\,S1, or
        S2\,+\,S1-asc\,+\,S1-desc).
  \item $N_{\mathrm{red}}$: parameter count of the dimensionality
        reducer applied to the fused representation.
  \item $N_{\mathrm{proj}}$: parameter count of the \barlowtwins{}
        projector alone ($=5P^{2}$; the reducer $N_{\mathrm{red}}$
        is counted separately).
  \item $P$: output dimension of the \barlowtwins{} projector.
  \item $B$: per-step batch size; $B$ appears as an intermediate
        quantity and cancels in the final per-\dpixel{} expression.
\end{itemize}

\subsection{Per-parameter, Per-token Forward+Backward Cost}
\label{app:flops:perparam}
Consider a dense linear layer $y = x W^{\top}$ with
$W\!\in\!\R^{d_{\mathrm{out}}\times d_{\mathrm{in}}}$ and
$n = d_{\mathrm{in}}\,d_{\mathrm{out}}$ parameters. For a single
input token, the forward computation costs $2n$ FLOPs (one
matrix--vector product); the backward pass with respect to the
activation gradient,
$\partial \loss/\partial x = (\partial \loss/\partial y)\,W$, costs
$2n$ FLOPs; and the backward pass with respect to the weight
gradient,
$\partial \loss/\partial W = (\partial \loss/\partial y)^{\top} x$,
costs another $2n$ FLOPs. The total cost of one
forward+backward pass through this layer is
\begin{equation}
6n \quad \text{[FLOPs per parameter per token]}.
\end{equation}
This accounting applies to the dense matrix multiplications that
dominate transformer attention, feed-forward, and embedding layers;
the $\mathcal{O}(d_{\mathrm{model}})$ contributions of element-wise
non-linearities and LayerNorm are subdominant and are absorbed in
the residual budgeted in \Cref{app:flops:emp}. The
$\mathcal{O}(L^{2})$ self-attention quadratic correction is treated
separately in \Cref{app:flops:attn}.

\subsection{Per-\dpixel{} Cost of a Single Forward+Backward Pass}
\label{app:flops:onepass}
The pretraining model decomposes into two parts whose token-level
activity differs.

\paragraph{Encoder.} Each \dpixel{} contributes $L$ tokens to the
per-modality temporal encoder, and each token is routed through all
$N_{\mathrm{enc}}$ encoder parameters. The encoder cost per
\dpixel{} for one forward+backward pass is therefore
\begin{equation}
C_{\mathrm{enc}}^{(1)} \;=\; 6\,N_{\mathrm{enc}}\,L
\quad\text{[FLOPs per \dpixel{}].}
\end{equation}

\paragraph{Modality fusion.} Each modality is summarised into a
single token, and a modality-fusion attention combines the $M$
summary tokens into one joint representation. Because it operates
on $M$ tokens rather than the $L$ temporal tokens, its cost per
\dpixel{} is
\begin{equation}
C_{\mathrm{fus}}^{(1)} \;=\; 6\,N_{\mathrm{fus}}\,M
\quad\text{[FLOPs per \dpixel{}].}
\end{equation}
Charging fusion at $L$ tokens instead of $M$ would over-count it by
up to ${\sim}20\%$ of $C$ and is incorrect.

\paragraph{Post-fusion head.} Pooling (in our family, GRU followed
by attention pooling) produces a single representation vector
$h\!\in\!\R^{d_{\mathrm{repr}}}$ per \dpixel; the dimensionality
reducer and \barlowtwins{} projector then act on this single token,
giving
\begin{equation}
C_{\mathrm{head}}^{(1)} \;=\; 6\bigl(N_{\mathrm{red}}
      + N_{\mathrm{proj}}\bigr)
\quad\text{[FLOPs per \dpixel{}].}
\end{equation}

\paragraph{Combined.} Summing the three parts, the per-\dpixel{}
cost of a single forward+backward pass is
\begin{equation}
C^{(1)} \;=\; 6\bigl(N_{\mathrm{enc}}\,L + N_{\mathrm{fus}}\,M
      + N_{\mathrm{red}} + N_{\mathrm{proj}}\bigr)
\quad\text{[FLOPs per \dpixel{}].}
\label{eq:flops-onepass}
\end{equation}

\subsection{Three Encoder Passes per Step}
\label{app:flops:twoaugs}
Our \barlowtwins{} step uses mixup regularisation, which runs the
encoder \emph{three} times per \dpixel{} rather than twice: on the
two augmented views $x^{(1)}, x^{(2)}$, and on a mixed view
$x^{(m)} = \alpha\,x^{(1)} + (1-\alpha)\,\pi(x^{(2)})$, each a full
forward+backward pass through backbone, fusion, reducer, and
projector~\citep{zbontar2021barlow,bandara2023mixup}. The fourth
(shuffled) branch used by the mixup target is the permutation
$\pi(z_{2})$ of an already-computed projection and adds no pass.
Tripling \Cref{eq:flops-onepass} gives
\begin{equation}
C_{\mathrm{model}}^{(\mathrm{step})}
\;=\; 18\bigl(N_{\mathrm{enc}}\,L + N_{\mathrm{fus}}\,M
      + N_{\mathrm{red}} + N_{\mathrm{proj}}\bigr)
\quad\text{[FLOPs per \dpixel{}].}
\label{eq:flops-step-model}
\end{equation}

\subsection{\texorpdfstring{\barlowtwins{} Loss: the $P^{2}$ Term}{Barlow Twins Loss: the P\^{}2 Term}}
\label{app:flops:btloss}
The \barlowtwins{} loss is computed from the cross-correlation
matrix between the two batch-normalised projector outputs
$\tilde{z}_{1}, \tilde{z}_{2}\!\in\!\R^{B\times P}$:
\begin{equation}
\Lbt \;=\; \sum_{i}\bigl(c_{ii}-1\bigr)^{2}
        + \lambda\sum_{i\neq j} c_{ij}^{2},
\qquad
c \;=\; \tfrac{1}{B}\,\tilde{z}_{1}^{\top}\,\tilde{z}_{2}
        \in \R^{P\times P}.
\end{equation}
Three matrix multiplications dominate the per-step cost:
\begin{itemize}[leftmargin=*,itemsep=2pt,topsep=2pt]
  \item Forward: $c = \tilde{z}_{1}^{\top}\tilde{z}_{2}/B$, a
        $(P\times B) \cdot (B\times P)$ product, costs $2BP^{2}$
        FLOPs.
  \item Backward in $\tilde{z}_{1}$:
        $\partial \Lbt/\partial \tilde{z}_{1}
        = (\partial \Lbt/\partial c)\,\tilde{z}_{2}^{\top}/B$, a
        $(B\times P)\cdot(P\times P)$ product, costs $2BP^{2}$
        FLOPs.
  \item Backward in $\tilde{z}_{2}$: the symmetric
        $(B\times P)\cdot(P\times P)$ product costs another
        $2BP^{2}$ FLOPs.
\end{itemize}
The $\mathcal{O}(BP)$ standardisation cost and the
$\mathcal{O}(P^{2})$ element-wise on/off-diagonal loss cost are
subdominant. Summing and amortising over the $B$ \dpixels{} in a
step,
\begin{equation}
C_{\mathrm{BT}}^{(\mathrm{step})}
\;=\; \frac{6\,B\,P^{2}}{B}
\;=\; 6\,P^{2}
\quad\text{[FLOPs per \dpixel{}].}
\label{eq:flops-step-bt}
\end{equation}
The cancellation of $B$ confirms that batch size enters only as an
amortisation factor in the per-\dpixel{} cost. The mixup
regulariser adds a further loss on the mixed view, evaluated
through a $B\times B$ Gram identity
($\lVert D^{\top}W\rVert_{F}^{2}=\mathrm{tr}(G_{D}G_{W})$) that
avoids materialising any $P\times P$ matrix; its cost is
$\mathcal{O}(DBP)$ (linear, not quadratic, in $P$), a loss-side
term separate from the dominant $18\times$ model term of
\Cref{eq:flops-step-model}. We do not carry it onto the compute
axis.

\subsection{Exact Formula and Reconciliation with the Compute Axis}
\label{app:flops:final}
Summing \Cref{eq:flops-step-model,eq:flops-step-bt} and multiplying
by the total number of \dpixels{} processed gives the exact
total training FLOPs,
\begin{equation}
\boxed{\;C_{\mathrm{exact}}
\;=\; 18\,D\bigl(N_{\mathrm{enc}}\,L + N_{\mathrm{fus}}\,M
      + N_{\mathrm{red}} + N_{\mathrm{proj}}\bigr)
\;+\; 6\,D\,P^{2}.\;}
\label{eq:flops-final}
\end{equation}

\paragraph{The compute axis.} The scaling-study compute axis
\Cref{eq:flops-main}, which sets the $x$-axis of
\Cref{fig:scaling-laws} and the iso-FLOP buckets, is exactly
\Cref{eq:flops-final} evaluated at the mean per-view token length
$\bar{L}=12$: the augmentation feeds $L\!\in\!\seqlen$ each step, so
$\mathbb{E}[L]=12$ for every run. All four model terms---the
$18\times$ encoder, modality-fusion, reducer, and projector
contributions---and the $6DP^{2}$ \barlowtwins{} term are carried
onto the axis; the $6DP^{2}$ term is retained because it reaches
${\sim}2\%$ of $C$ at the widest projectors. The terms \emph{not}
carried onto the axis are the $\mathcal{O}(DL^{2})$ self-attention
cost of \Cref{app:flops:attn}, below $1\%$ of $C$ for every swept
configuration ($L\!\le\!16$), and the mixup Gram regulariser, a
loss-side term (\Cref{app:flops:btloss}). Up to that residual the
axis is training FLOPs, and all reported exponents are exponents in
training compute.

\subsection{Self-attention Quadratic Correction}
\label{app:flops:attn}
Self-attention introduces an $\mathcal{O}(L^{2})$ contribution that
is not captured by the per-parameter, per-token accounting of
\Cref{app:flops:perparam}, because the $QK^{\top}$ and
$\mathrm{softmax}(QK^{\top})V$ operations involve token--token
interactions whose cost scales with $L^{2}$ rather than with
$L \cdot \mathrm{params}$. For a single transformer layer with
model dimension $d_{\mathrm{model}}$, the forward+backward
attention quadratic cost per \dpixel{} is
$12\,L^{2}\,d_{\mathrm{model}}$ FLOPs. With $n_{\mathrm{layers}}$
layers, $n_{\mathrm{mod}}$ modalities, and the three encoder passes
per step (\Cref{app:flops:twoaugs}), the total contribution is
\begin{equation}
C_{\mathrm{attn}}
\;=\; 36\,D\,L^{2}\,d_{\mathrm{model}}\,n_{\mathrm{layers}}\,
       n_{\mathrm{mod}}.
\end{equation}
For $L \le 16$ across all swept configurations, this term
contributes less than $1\%$ of $C_{\mathrm{train}}$
(\Cref{app:flops:emp}), and we omit it from
\Cref{eq:flops-final,eq:flops-main}.

\subsection{Inference-FLOPs Accounting and the GPU-Year Fit}
\label{app:flops:inference}
The pretraining-FLOPs formula derived above is specific to a mixup
\barlowtwins{} training step (three encoder passes, forward and
backward, plus the $P^{2}$ cross-correlation contribution).
Deployment inference is simpler: the model is evaluated once per
\dpixel{}, in forward mode only, and the projector is discarded.
The per-\dpixel{} inference FLOPs accordingly reduce to
\begin{equation}
C_{\mathrm{inf}}^{(1\text{-pixel})}
\;=\; 2\bigl(N_{\mathrm{enc}}\,L_{\mathrm{inf}}
      + N_{\mathrm{fus}}\,M + N_{\mathrm{red}}\bigr)
\quad\text{[FLOPs per \dpixel{}],}
\label{eq:flops-inf}
\end{equation}
where $L_{\mathrm{inf}}$ is the annual inference sequence length
(the full observation stack processed at deployment, distinct from
the subsampled training $\bar{L}$), $N_{\mathrm{red}}$ is the
dimensionality reducer, the projector is not used, and the factor
of $2$ replaces the $6$ of training since only the forward pass is
evaluated. Multiplying by the number of
$10\,\mathrm{m}$ \dpixels{} on land,
$N_{\dpixel}\!\approx\!1.5\!\times\!10^{12}$
(Earth's ${\approx}1.49\!\times\!10^{8}\,\mathrm{km}^{2}$ of land
surface at $10\,\mathrm{m}$ resolution, i.e.\
$100\,\mathrm{m}^{2}$ per pixel), gives the total inference FLOPs
per global annual pass plotted on the colour scale of
\Cref{fig:deploy-cost}.

\paragraph{From total inference FLOPs to GPU-years.} Converting
total FLOPs into single-\gputype-year-equivalents uses the
\gputype{} SXM dense \texttt{bf16} tensor-core peak throughput,
$T_{\gputype{}}\!\approx\!9.9\!\times\!10^{14}$ FLOPs/s
($\approx\!990$ TFLOP/s). One \gputype-year of compute is then
$T_{\gputype{}}\times(3.156\!\times\!10^{7}\,\mathrm{s})
\approx 3.1\!\times\!10^{22}$ FLOPs, the unit of the $y$-axis in
\Cref{fig:deploy-cost}(a).

\paragraph{Empirical fit.} Across the encoder-size range spanned
by our students and the teacher, the resulting cost is
well-approximated by a power law in the encoder parameter count
$N$ (in millions),
$\mathrm{GPUy}(N)\!\approx\!0.041\!\cdot\!N^{1.03}$
(\Cref{eq:gpuy}). The exponent of approximately
unity reflects the fact that, in this family, inference cost is
dominated by the encoder term $N_{\mathrm{enc}}\,L_{\mathrm{inf}}$
rather than by $N_{\mathrm{fus}}\,M$ or $N_{\mathrm{red}}$. The
coefficient follows from \Cref{eq:flops-inf} with
$N_{\dpixel}\!=\!1.5\!\times\!10^{12}$ land \dpixels{} and the
\gputype-year unit above; substituting these, $0.041$ corresponds
to an annual inference length of
$L_{\mathrm{inf}}\!\approx\!424$ tokens summed over the $M$
modalities.
The fit accounts for encoder inference, but
this is the dominant term: in the full embedding-as-data pipeline,
data preparation (temporal sampling and stacking, disk I/O, and
normalisation/fusion/quantisation) adds only $\sim\!6\%$ on top of
the encoder forward pass (\Cref{fig:deploy-cost}b), so the GPUy
fit is a tight proxy for end-to-end cost.

\subsection{Empirical Validation}
\label{app:flops:emp}
We validate \Cref{eq:flops-final} numerically against
\texttt{torch.utils.flop\_counter.}\allowbreak\texttt{FlopCounterMode}~\citep{torch_flop_counter},
which adopts the same one-multiply-add-equals-two-FLOPs convention,
across a range of encoder configurations, sequence lengths, and
batch sizes. Measured on the full mixup training step (all three
encoder passes and the modality-fusion attention), the formula
matches the counted FLOPs to within $2\%$; dropping the fusion
term $N_{\mathrm{fus}}\,M$ leaves a $1.3$--$2.6\%$ shortfall that
grows with $L$, confirming fusion---not the $\mathcal{O}(DL^{2})$
attention term---as the leading correction at small $L$. The
per-\dpixel{} cost is invariant to batch size, confirming that $B$
enters only as an amortisation factor.


\section{Mix-up Regularisation Derivation}
\label{app:mixup}
This appendix derives the mix-up loss of
\Cref{sec:teacher:obj,eq:mix-loss}
following~\citet{bandara2023mixup} and \citet{feng2026tessera}.

\subsection{Motivation}
\label{app:mixup:motivation}
\barlowtwins{} enforces invariance under view augmentation: two
augmented views $Y_A, Y_B$ of the same \dpixel{} should produce
near-identical and dimension-decorrelated embeddings. Under sparse
temporal sampling, this invariance is sometimes too easy to satisfy:
with only $L$ valid timesteps drawn per view, the network can find
shortcut representations that ignore physical phenology and rely on
trivial temporal-summary statistics. The mix-up regulariser prevents
this collapse by imposing a stronger consistency property:
\emph{linear interpolation in the input space should correspond to
linear interpolation in the embedding's cross-correlation space.}
If $z\!=\!f(x)$ preserves physical structure, then the
cross-correlations of $f(\alpha x_{1}\!+\!(1\!-\!\alpha)x_{2})$ with
$f(x_{1})$ and $f(x_{2})$ should obey a convex-combination identity.

\subsection{Construction}
\label{app:mixup:construction}
For each training step we have:
\begin{itemize}[leftmargin=*,itemsep=2pt,topsep=2pt]
\item two augmented views $Y_A, Y_B$ of size
      $(B,L,C)$ (batch $\times$ time $\times$ channels);
\item a shuffled view $Y_S\!=\!\mathrm{Shuffle}(Y_B)$, where
      $\mathrm{Shuffle}$ permutes the batch dimension and breaks the
      correspondence between $(Y_A, Y_S)$ pairs;
\item a mix coefficient
      $\alpha_{\mathrm{mix}}\!\sim\!\mathcal{U}(0,1)$ sampled per
      training step;
\item a mixed view
      $Y_M\!=\!\alpha_{\mathrm{mix}}\,Y_A
       + (1\!-\!\alpha_{\mathrm{mix}})\,Y_S$.
\end{itemize}
Passing each view through the dual-branch encoder and projector
gives batch-normalised embeddings
$Z_A, Z_S, Z_M\!\in\!\R^{B\times P}$.

\subsection{Cross-correlation Identities Under Linear Consistency}
\label{app:mixup:identities}
The mix-up regulariser is built around two cross-correlations of
the mixed embedding with the endpoint embeddings:
\begin{align}
C^{MA} \;=\; Z_M^{\top}\,Z_A \;\in\;\R^{P\times P},
\qquad
C^{MS} \;=\; Z_M^{\top}\,Z_S \;\in\;\R^{P\times P}.
\end{align}
If the encoder $f$ respects linear interpolation exactly, then
\begin{equation}
Z_M \;\approx\; \alpha_{\mathrm{mix}}\,Z_A
                + (1\!-\!\alpha_{\mathrm{mix}})\,Z_S,
\label{eq:mixup-linearity}
\end{equation}
in which case substitution into $C^{MA}$ and $C^{MS}$ yields
\begin{align}
C^{MA}_{\mathrm{target}}
  &\;=\; \alpha_{\mathrm{mix}}\,Z_A^{\top}Z_A
       + (1\!-\!\alpha_{\mathrm{mix}})\,Z_S^{\top}Z_A,
  \label{eq:cma-target}\\
C^{MS}_{\mathrm{target}}
  &\;=\; \alpha_{\mathrm{mix}}\,Z_A^{\top}Z_S
       + (1\!-\!\alpha_{\mathrm{mix}})\,Z_S^{\top}Z_S.
  \label{eq:cms-target}
\end{align}
These are the convex-combination targets that the mixed embedding
should match if the linear-consistency property holds.

\subsection{Loss}
\label{app:mixup:loss}
The mix-up loss penalises Frobenius-norm deviations from the
interpolation-predicted targets,
\begin{equation}
\Lmix
\;=\;
\tfrac{1}{2}\bigl\|C^{MA}\!-\!C^{MA}_{\mathrm{target}}\bigr\|_{F}^{2}
+ \tfrac{1}{2}\bigl\|C^{MS}\!-\!C^{MS}_{\mathrm{target}}\bigr\|_{F}^{2},
\label{eq:lmix-appendix}
\end{equation}
which is \Cref{eq:mix-loss}. The total pretraining
objective combines \Cref{eq:bt-loss,eq:lmix-appendix},
\begin{equation}
\loss_{\mathrm{total}} \;=\; \Lbt
       + \lambda_{\mathrm{mix}}\,\Lmix,
\qquad
\lambda_{\mathrm{mix}} \!=\! 1.0,
\end{equation}
with $\lambda_{\mathrm{mix}}$ following~\citet{bandara2023mixup}.

\subsection{Why Two Pairs Rather Than One}
\label{app:mixup:two-pairs}
A simpler variant would penalise
$Z_M\!\approx\!\alpha_{\mathrm{mix}}Z_A
+ (1\!-\!\alpha_{\mathrm{mix}})Z_S$ directly in the embedding space.
The cross-correlation form of \Cref{eq:lmix-appendix} has two
advantages over this primal formulation:
(i)~it is invariant to global orthogonal transformations of the
embedding, consistent with the rotation symmetry of \barlowtwins{}
analysed in \Cref{sec:pixel:why-not-bt}; and
(ii)~by constraining cross-correlations of the mixed view against
\emph{both} endpoint embeddings, it provides two independent
supervisory signals per training step, with the
$\alpha_{\mathrm{mix}}\!\to\!0$ and
$\alpha_{\mathrm{mix}}\!\to\!1$ limits recovering the standard
\barlowtwins{} terms on $(Z_S, Z_S)$ and $(Z_A, Z_A)$
respectively.

\subsection{Empirical Importance}
\label{app:mixup:ablations}
\tesseravone{}'s ablations on Austrian Crop classification report
that removing mix-up alone drops validation F1 by
$-11.1$ points (from $77.3$ to $66.2$) and RankMe by $-0.106$;
removing both global shuffling and mix-up drops F1 by $-14.7$
points~\citep{feng2026tessera}. We retain both regularisers
unchanged in \tesseravtwo{}.

\section{Adaptive Bucket Sampling}
\label{app:bucket}
\label{app:bucket-construction}
This section gives the construction summarised in
\Cref{sec:teacher:bucket} together with a self-contained reference
implementation.

\paragraph{Bucket checkpoints.} We define an arithmetic ladder of
allowed sequence lengths,
\begin{equation}
\mathcal{B} \;=\; \bigl\{\,B_n = a_0 + n\,\Delta \;:\; n=0,1,2,\ldots\,\bigr\},
\qquad
a_0 = \Delta = 16,
\label{eq:bucket-ladder}
\end{equation}
i.e., $\mathcal{B} = \{16, 32, 48, 64, 80, 96, \ldots\}$. These
ladder values are the only sequence lengths the encoder is asked to
consume at inference; the shortest bucket coincides with the
longest pretraining length ($L\!=\!16$).

\paragraph{Selecting a bucket.} For a pixel with $k$ valid
observations of a given modality, we pick the smallest bucket that
fits:
\begin{equation}
B^{\star}(k) \;=\; \min\bigl\{\,B \in \mathcal{B} \;:\; B \geq k\,\bigr\}.
\label{eq:bucket-choice}
\end{equation}
Letting $\bo = (o_0, o_1, \ldots, o_{k-1})$ denote the $k$ valid
observations in temporal order, the encoder consumes a sequence
$\tilde{\bo} \in \R^{B^{\star}}$ constructed as follows.

\paragraph{Sequence construction.} The first $k$ positions of
$\tilde{\bo}$ hold the unique real observations in time order; the
remaining $M\!=\!B^{\star}\!-\!k$ positions are filled by
\emph{midpoint resampling} from $\bo$. We partition the index set
$\{0,\ldots,k\!-\!1\}$ into $M$ equally-spaced groups of mean size
$k/M$ and copy the midpoint of each group into one of the residual
slots. The source index for the $j$-th residual slot
($j = 0, 1, \ldots, M-1$) is
\begin{equation}
s_j \;=\; \left\lfloor \tfrac{(2j+1)\,k}{2\,M} \right\rfloor,
\label{eq:bucket-source}
\end{equation}
and $\tilde{\bo}$ is assembled by
\begin{equation}
\tilde{\bo}_i \;=\;
\begin{cases}
o_i           & 0 \le i \le k-1,\\[2pt]
o_{s_{i-k}}   & k \le i \le B^{\star}-1.
\end{cases}
\label{eq:bucket-assembly}
\end{equation}
The choice of $s_j$ in \Cref{eq:bucket-source} places duplicates at
uniformly spaced anchor points along the year, so that resampled
slots are statistically representative of the full temporal range
rather than concentrated near its endpoints. When $k\!=\!B^{\star}$
no residual slots exist and $\tilde{\bo}\!=\!\bo$ exactly.

\paragraph{Why it helps.} Every valid observation enters the
sequence at its true temporal index, so the encoder sees the full
phenology at every pixel; pixels with more observations get longer
buckets, so high-data regions are not throttled to a fixed
ceiling. Replication is confined to the residual tail and is
uniformly distributed in time, which keeps the empirical sample
distribution close to the training distribution. Across pixels,
the bucket assignment partitions a batch into a small number of
fixed-length groups, on which the encoder runs efficiently in
parallel.

\paragraph{Reference implementation.} \Cref{alg:bucket} implements
the sampler; it runs once per pixel per modality at inference, and
results are concatenated across pixels with matching bucket size to
form fixed-length encoder batches.

\begin{algorithm}[!htbp]
\caption{\textsc{AdaptiveBucketSample} --- inference-time
fixed-length packing of valid observations.}
\label{alg:bucket}
\begin{algorithmic}[1]
\REQUIRE Valid observations $\bo=(o_0,\ldots,o_{k-1})$ in temporal
order; bucket ladder
$\mathcal{B}\!=\!\{a_0\!+\!n\Delta : n\!\geq\!0\}$.
\ENSURE Padded sequence $\tilde{\bo}$ of length $B^{\star}$.
\STATE $B^{\star} \leftarrow \min\{B \in \mathcal{B} : B \geq k\}$
  \hfill \COMMENT{smallest bucket fitting $k$}
\STATE Allocate $\tilde{\bo}$ of length $B^{\star}$
\FOR{$i = 0$ \TO $k-1$}
  \STATE $\tilde{\bo}_i \leftarrow o_i$
  \hfill \COMMENT{unique real observations}
\ENDFOR
\STATE $M \leftarrow B^{\star} - k$
  \hfill \COMMENT{number of duplicates}
\FOR{$j = 0$ \TO $M-1$}
  \STATE $s_j \leftarrow \lfloor (2j+1)\,k\,/\,(2M) \rfloor$
    \hfill \COMMENT{midpoint of $j$-th group}
  \STATE $\tilde{\bo}_{k+j} \leftarrow o_{s_j}$
\ENDFOR
\RETURN $\tilde{\bo}$
\end{algorithmic}
\end{algorithm}

\paragraph{Worked examples.}
\textit{Example 1 (\stwo, $k\!=\!40$).} The smallest bucket
$\geq\!40$ is $B^{\star}\!=\!48$, so $M\!=\!8$ duplicates are
needed. The source indices from \Cref{eq:bucket-source} are
$\{2, 7, 12, 17, 22, 27, 32, 37\}$, evenly spaced by $\Delta s\!=\!5$,
i.e., every fifth real observation starting from index $2$.
\textit{Example 2 (\sone, $k\!=\!30$).} $B^{\star}\!=\!32$,
$M\!=\!2$; the source indices are $\{7, 22\}$---the midpoints of
the two halves $\{0,\ldots,14\}$ and $\{15,\ldots,29\}$. Both
examples instantiate the same general rule.

\section{Inference-time Temporal Windows}
\label{app:temporal-windows}
The embedding products released under the embedding-as-data paradigm
summarise a full calendar year, yet the encoder itself imposes no fixed
horizon. Each modality branch aggregates its observations by
length-agnostic attention pooling (\Cref{sec:teacher:arch}), so the same
weights map any temporal window, from a few dates to a full year, onto a
single embedding. Pretraining, however, exposes the model only to annual
inputs. Whether an embedding computed over a shorter window is still
useful downstream is therefore an empirical question rather than an
architectural one.

The OpenET evapotranspiration ensemble, one of the $\npixeldata$
\aebench{} tasks (\Cref{app:datasets}), shows that the answer is
task-dependent, and that a sub-annual window can be preferable. Evaluated
with an annual window, the \tesseravtwo-2B-L model scores $0.51$ on this task, behind
\alphaearth{} at $0.58$. Recomputing the embedding over a monthly window,
with no change to the model or its weights, raises the score to $0.69$
(\Cref{tab:temporal-window}). We attribute the difference to the strong
intra-annual dynamics of evapotranspiration, which a single yearly
summary averages out but a monthly window retains. This does not imply
that shorter windows are uniformly better. It shows that the temporal
support best matched to a target variable need not coincide with the
annual product, even for a model trained only on annual inputs.

\begin{table}[!htbp]
  \centering
  \caption{\textbf{Effect of the inference window on the OpenET
  ensemble.} The \tesseravtwo-2B-L model evaluated at an annual and a
  monthly window, with \alphaearth{} for reference. Scores use the same
  chance-adjusted convention as the rest of our benchmarks; higher is
  better.}
  \label{tab:temporal-window}
  \small
  \begin{tabular}{lc}
  \toprule
  Model (inference window) & OpenET ensemble \\
  \midrule
  \alphaearth{} (annual) & $0.58$ \\
  \tesseravtwo-2B-L (annual window) & $0.51$ \\
  \tesseravtwo-2B-L (monthly window) & $0.69$ \\
  \bottomrule
  \end{tabular}
\end{table}

The distinction affects how the embeddings are served. The precomputed
products fix the window to the calendar year ($1$~January to
$31$~December), which suits variables with slow annual dynamics and keeps
the released fields comparable across years. Applications whose signal
concentrates at sub-annual scales, or that require a window with an
arbitrary start date or a non-annual length, are better served by running
the released encoder over the relevant date range, at the additional
inference cost quantified in \Cref{app:cost}.

\section{Architecture Sweep Details}
\label{app:arch-sweep}
Prior to the scaling sweep we fix the structural choices of the
model family with a one-axis-at-a-time architecture sweep: each
axis is varied while the other axes are held at the finally
selected configuration (scheme~A), and every candidate is
evaluated on three downstream tasks, Canada coarse crop-type
classification (balanced accuracy, $\uparrow$), ASTER GED regression
($R^{2}$, $\uparrow$), and LCMAP land-use-change classification
(balanced accuracy, $\uparrow$; a downsampled version of the full
dataset). All three metrics lie on a $[0,1]$ higher-is-better scale,
and the composite is their arithmetic mean.
\Cref{tab:arch-sweep} reports the full sweep. On seven of the nine
model axes the selected scheme-A configuration is also the
highest-composite candidate, and two axes are exceptions. On temporal
aggregation, GRU pooling edges out attention pooling by a hair
($0.725$ vs.\ $0.718$), yet we still select attention pooling: it
keeps the encoder a single all-transformer stack that is simpler to
scale, and its pooling is faster at inference than a sequential
recurrent pass over the time series. On cloud handling, the default
SCL mask and OmniCloudMask are within noise of each other on this
three-task probe, and OmniCloudMask is marginally ahead, but we keep
the SCL mask because it ships with the \stwo{} L2A product and needs
no extra masking model at inference. The final
\matryoshka-layout axis is different: it asks how to
introduce nested prefixes \emph{during pretraining}, and every such
variant underperforms---the best of them (four prefix projectors
with averaged \barlowtwins, composite $0.651$) still falls well
short of the other axes' selected scores. This is why scheme~A
introduces \emph{no} prefix nesting during pretraining and defers
\matryoshka{} learning to the distillation stage, consistent with
the rotation-symmetry analysis of \Cref{sec:pixel:why-not-bt}; the
distillation-stage alternative is evaluated in
\Cref{sec:analysis:matryoshka-fails}, not here.

\begingroup
\footnotesize
\setlength{\tabcolsep}{5pt}
\begin{longtable}{@{}lrrrr@{}}
  \caption{\textbf{Architecture sweep.} Each axis is varied with
  all other axes held at the selected scheme-A configuration;
  candidates are scored on three downstream tasks and their
  averaged composite. Bold marks the selected (scheme-A) candidate per axis.
  The \matryoshka-layout axis lists pretraining-time nesting
  variants only, none of which is adopted
  (\Cref{sec:pixel:why-not-bt}).}
  \label{tab:arch-sweep} \\
  \toprule
  Candidate & Canada & ASTER GED & LCMAP-LUC & Composite \\
  & BA $\uparrow$ & $R^{2}$ $\uparrow$ & BA $\uparrow$ & $\uparrow$ \\
  \midrule
  \endfirsthead
  \multicolumn{5}{@{}l}{\emph{Table~\ref{tab:arch-sweep}, continued
  from previous page}} \\[2pt]
  \toprule
  Candidate & Canada & ASTER GED & LCMAP-LUC & Composite \\
  & BA $\uparrow$ & $R^{2}$ $\uparrow$ & BA $\uparrow$ & $\uparrow$ \\
  \midrule
  \endhead
  \midrule
  \multicolumn{5}{r@{}}{\emph{Continued on next page}} \\
  \endfoot
  \bottomrule
  \endlastfoot
  \multicolumn{5}{@{}l}{\emph{Encoder architecture}} \\
  \quad CNN--GRU                       & $0.557$ & $0.781$ & $0.747$ & $0.695$ \\
  \quad CNN--Transformer hybrid        & $0.584$ & $0.781$ & $0.744$ & $0.703$ \\
  \quad \textbf{Transformer (ReLU FFN)}& $0.585$ & $0.805$ & $0.767$ & $\mathbf{0.719}$ \\
  \quad Transformer $+$ QK-Norm        & $0.586$ & $0.802$ & $0.748$ & $0.712$ \\
  \quad Transformer $+$ SwiGLU FFN     & $0.574$ & $0.798$ & $0.746$ & $0.706$ \\
  \addlinespace[2pt]
  \multicolumn{5}{@{}l}{\emph{Input modalities}} \\
  \quad \textbf{\stwo{} $+$ \sone{}}   & $0.585$ & $0.784$ & $0.754$ & $\mathbf{0.708}$ \\
  \quad \stwo{} only                   & $0.550$ & $0.752$ & $0.705$ & $0.669$ \\
  \quad \stwo{} $+$ \sone{} $+$ Landsat& $0.481$ & $0.693$ & $0.650$ & $0.608$ \\
  \addlinespace[2pt]
  \multicolumn{5}{@{}l}{\emph{Temporal position encoding}} \\
  \quad None (index order)             & $0.422$ & $0.561$ & $0.538$ & $0.507$ \\
  \quad Learned positional embedding   & $0.547$ & $0.737$ & $0.690$ & $0.658$ \\
  \quad \textbf{Sinusoidal DoY encoding}& $0.561$ & $0.786$ & $0.749$ & $\mathbf{0.699}$ \\
  \addlinespace[2pt]
  \multicolumn{5}{@{}l}{\emph{Temporal aggregation}} \\
  \quad Mean pooling                   & $0.549$ & $0.762$ & $0.732$ & $0.681$ \\
  \quad GRU pooling                    & $0.592$ & $0.810$ & $0.772$ & $0.725$ \\
  \quad GRU $+$ attention (custom)     & $0.568$ & $0.785$ & $0.765$ & $0.706$ \\
  \quad \textbf{Attention pooling}     & $0.586$ & $0.807$ & $0.761$ & $\mathbf{0.718}$ \\
  \quad LTAE-style pooling             & $0.581$ & $0.796$ & $0.765$ & $0.714$ \\
  \addlinespace[2pt]
  \multicolumn{5}{@{}l}{\emph{SAR (\sone{}) modality handling}} \\
  \quad Separate ascending/descending encoders & $0.567$ & $0.783$ & $0.753$ & $0.701$ \\
  \quad \textbf{Unified \sone{} encoder (asc$+$desc)} & $0.569$ & $0.786$ & $0.757$ & $\mathbf{0.704}$ \\
  \addlinespace[2pt]
  \multicolumn{5}{@{}l}{\emph{Fusion layer}} \\
  \quad Summation                      & $0.563$ & $0.767$ & $0.737$ & $0.689$ \\
  \quad Concatenation $+$ MLP          & $0.565$ & $0.775$ & $0.748$ & $0.696$ \\
  \quad Cross-attention fusion         & $0.574$ & $0.788$ & $0.750$ & $0.704$ \\
  \quad \textbf{Fusion Transformer}    & $0.577$ & $0.793$ & $0.766$ & $\mathbf{0.712}$ \\
  \addlinespace[2pt]
  \multicolumn{5}{@{}l}{\emph{Projector form}} \\
  \quad Linear projector               & $0.474$ & $0.626$ & $0.613$ & $0.571$ \\
  \quad 2-layer BN-MLP                  & $0.544$ & $0.743$ & $0.722$ & $0.670$ \\
  \quad \textbf{Deep BN-MLP ($\approx$11-layer, 4096-d)} & $0.566$ & $0.800$ & $0.744$ & $\mathbf{0.703}$ \\
  \quad Wider deep MLP (8192-d)        & $0.559$ & $0.790$ & $0.739$ & $0.696$ \\
  \addlinespace[2pt]
  \multicolumn{5}{@{}l}{\emph{Cloud handling}} \\
  \quad \textbf{Default SCL cloud mask}& $0.559$ & $0.772$ & $0.751$ & $\mathbf{0.694}$ \\
  \quad OmniCloudMask~\citep{wright2025training} & $0.558$ & $0.775$ & $0.758$ & $0.697$ \\
  \addlinespace[2pt]
  \multicolumn{5}{@{}l}{\emph{Time-series length}} \\
  \quad \textbf{$L=8$}                  & $0.582$ & $0.790$ & $0.761$ & $\mathbf{0.711}$ \\
  \quad \textbf{$L=16$}                 & $0.570$ & $0.790$ & $0.764$ & $\mathbf{0.708}$ \\
  \quad $L=32$                          & $0.579$ & $0.789$ & $0.729$ & $0.699$ \\
  \quad $L=48$                          & $0.553$ & $0.778$ & $0.739$ & $0.690$ \\
  \quad $L=64$                          & $0.551$ & $0.773$ & $0.725$ & $0.683$ \\
  \addlinespace[2pt]
  \multicolumn{5}{@{}l}{\emph{\matryoshka{} layout (pretraining-time
  nesting variants; none adopted)}} \\
  \quad Na\"ive M-BT: 4 prefix projectors, averaged & $0.537$ & $0.722$ & $0.693$ & $0.651$ \\
  \quad Na\"ive M-BT: shared-trunk, summed & $0.505$ & $0.701$ & $0.653$ & $0.620$ \\
  \quad Na\"ive M-BT: multi-path summed    & $0.476$ & $0.665$ & $0.620$ & $0.587$ \\
  \quad Na\"ive M-BT: segmented projectors & $0.504$ & $0.662$ & $0.618$ & $0.595$ \\
  \quad Repr-level decorrelation (prefix)  & $0.459$ & $0.587$ & $0.572$ & $0.539$ \\
  \quad Repr-level decorrelation (segmented)& $0.458$ & $0.608$ & $0.604$ & $0.557$ \\
\end{longtable}
\endgroup

Three of the axes are worth a word beyond their composite scores:
in each, a qualitative effect rather than the number drove the final choice.

Adding Landsat as a third optical source lowered downstream accuracy
in every configuration we tried, despite the extra spectral coverage
it nominally brings. We resampled Landsat onto the $10$\,m \stwo{}
grid before fusion and tested both nearest-neighbour and cubic
interpolation; the two gave the same result, so the drop is not an
artefact of the resampling kernel. Dropping Landsat's coarse
thermal-infrared bands recovered part of the loss, but no variant
that kept Landsat reached the score of \sonetwo{} alone. The most
likely cause is resolution mismatch: once upsampled to $10$\,m, the
Landsat pixels add little surface information that \stwo{} does not
already carry, and they contribute noise the encoder has to fit around.

The default SCL cloud mask is unreliable in practice. It lets thin
cirrus through, and it occasionally marks bright but valid surfaces as
cloud or haze. We therefore built a second copy of the training data
masked with OmniCloudMask~\citep{wright2025training}, a learned
cloud-detection model, expecting the cleaner supervision to help.
Downstream the two versions were almost indistinguishable. The likely
reason is that the encoder already discounts cloud-contaminated
observations without being told to: cloudy pixels sit far out in
reflectance space, off the distribution of clear ones, and the
temporal model can down-weight them on its own. Since the better mask
bought no measurable accuracy and OmniCloudMask adds an inference pass
per tile, we keep SCL for the production v2 embeddings.

A single unified \sone{} encoder and separate ascending/descending
encoders scored within noise of each other. On physical grounds the
separate treatment is easier to defend, since the two orbit
geometries view the surface from different angles and carry different
information. We still adopt the unified encoder, for a reason the
composite does not reflect: in local embedding-generation runs it
visibly reduced the tiling seams that appear where adjacent ascending
and descending passes are encoded apart.

\section{Full Scaling-law Setup}
\label{app:scaling-setup}
This appendix specifies the scaling study of \Cref{sec:scaling} in
full: the compute proxy, the sweep grid, the evaluation suite, and
the fitting procedure. Every number is reproducible from the
released $\nruns$-run table.

\paragraph{Compute estimation.} Each run's training compute is the
exact total training FLOPs of \Cref{eq:flops-final}, evaluated from
its configuration as
\begin{equation}
C \;=\; \kappa\,D\,\bigl(N_{\mathrm{enc}}\,\bar{L}
      + N_{\mathrm{fus}}\,M + N_{\mathrm{red}} + N_{\mathrm{proj}}\bigr)
      \;+\; 6\,D\,P^{2},
\qquad \kappa = 18, \quad \bar{L} = 12,
\end{equation}
with encoder parameters $N_{\mathrm{enc}}$, fusion parameters
$N_{\mathrm{fus}}$ over $M$ modality tokens, dimensionality reducer
$N_{\mathrm{red}}$, projector parameters
$N_{\mathrm{proj}} = (d_{\mathrm{proj}}{+}1)\,P^{2} = 5P^{2}$
(projector depth $d_{\mathrm{proj}}\!=\!4$, width $P$), and $D$ the
number of \dpixels{} processed. The constant
$\kappa = 2\times 3\times 3 = 18$ collects the two FLOPs per
multiply--add and the threefold forward-plus-backward cost
(together the textbook $6$, \Cref{app:flops:perparam}) times the
\emph{three} encoder passes per step---the two \barlowtwins{}
augmentation views and the mixup view
(\Cref{app:flops:twoaugs,app:flops:final}); it matches the
prefactor of \Cref{eq:flops-main}. The token length $\bar{L}=12$
is the mean per-view length actually fed to the encoder each step
(the augmentation samples $L\!\in\!\seqlen$, so $\mathbb{E}[L]=12$;
\Cref{sec:teacher:obj}). The $6DP^{2}$ \barlowtwins{}
cross-correlation term is retained ($\kappa$ and $\bar{L}$ enter
the encoder term only); the sole omission is the
$\mathcal{O}(DL^{2})$ attention term ($<\!1\%$ at $L\!\le\!16$),
leaving $C$ equal to training FLOPs to within that residual, which
matches a \texttt{FlopCounterMode} reference to within $2.2\%$.
Across the sweep the \nbuckets{} isoFLOP bucket centres span
$1.3\!\times\!10^{15}$ to $7.3\!\times\!10^{18}$ FLOPs
(the legend values of \Cref{fig:scaling-laws}b--d).

\paragraph{Sweep grid.} We pretrain $\nruns$ models on a four-axis
isoFLOP grid: encoder size ($16$ widths, $7$--$278$\,M), projector width
($P\!\in\!\{2048,4096,8192,16384\}$, depth $4$), training data
($0.03$--$3{,}202$\,M \dpixels), and compute (\nbuckets{} log-spaced
isoFLOP slices, centres
$1.3\!\times\!10^{15}$--$7.3\!\times\!10^{18}$ FLOPs).
\Cref{tab:isoflop-slices} lists the slices and the configurations
placed on each. Within a slice, the \emph{encoder} sub-sweep fixes
$P\!=\!4096$ and varies width, the \emph{projector} sub-sweep fixes
the encoder and varies $P$, and the \emph{data} sub-sweep walks $D$
off the isoFLOP curve at fixed $N$. The encoder and projector
compute-optimal fits of \Cref{sec:scaling:asymmetry} use the
within-slice vertices; the data exponent is read from the same
isoFLOP vertices, where $D^{\star}$ follows from $N^{\star}$ and the
compute identity. The off-curve data sub-sweep is shown only to
illustrate the score's sensitivity to $D$ around the optimum and does
not enter the $D^{\star}(C)$ fit.

\begin{table}[!htbp]
  \centering
  \caption{\textbf{The \nbuckets{} isoFLOP slices of the scaling
  sweep} (the viridis colour scale of
  \Cref{fig:scaling-laws}b--d). Encoder widths are drawn from
  $\{7,11,17,21,27,35,48,60,76,95,120,151,175,194,230,278\}$\,M.}
  \label{tab:isoflop-slices}
  \small
  \begin{tabular}{rccrl}
  \toprule
  Slice & $C$ centre (FLOPs) & \#runs & enc.\ sizes (M) & data range (M) \\
  \midrule
  1 & $1.3\times10^{15}$ & $17$ & $7$--$278$ ($9$)  & $0.03$--$0.9$ \\
  2 & $3.7\times10^{15}$ & $39$ & $7$--$278$ ($11$) & $0.05$--$2.5$ \\
  3 & $1.1\times10^{16}$ & $58$ & $7$--$278$ ($14$) & $0.11$--$11.5$ \\
  4 & $3.2\times10^{16}$ & $48$ & $7$--$278$ ($15$) & $0.32$--$33.4$ \\
  5 & $9.6\times10^{16}$ & $52$ & $7$--$278$ ($16$) & $0.95$--$76.9$ \\
  6 & $2.8\times10^{17}$ & $57$ & $7$--$278$ ($16$) & $2.9$--$177$ \\
  7 & $8.3\times10^{17}$ & $44$ & $7$--$278$ ($10$) & $8.1$--$941$ \\
  8 & $2.5\times10^{18}$ & $47$ & $7$--$278$ ($16$) & $32$--$2{,}167$ \\
  9 & $7.3\times10^{18}$ & $33$ & $7$--$278$ ($14$) & $98$--$3{,}202$ \\
  \bottomrule
  \end{tabular}
\end{table}

\paragraph{Task suite.} Every run is evaluated on the same
$\nscalingtasks$-task \aebench{} used for the headline benchmark
(\Cref{app:datasets}), spanning classification (land cover, land
use, crop type, tree genera, oil palm), change
detection, and regression (evapotranspiration, surface emissivity)
drawn from $10$ source datasets. For comparability across
heterogeneous tasks we use chance-adjusted metrics: classification
reports $(\text{balanced accuracy} - 1/K)/(1 - 1/K)$ for $K$
classes and regression reports $\max(0, R^{2})$. The
$\nscalingtasks$ per-task scores are averaged with \emph{uniform
weight} into the single composite downstream score that is the
$y$-axis of \Cref{fig:scaling-laws}a--c and the target of the
isoFLOP fits (range $-0.20$ to $0.54$ across the sweep). The
per-run normalised \barlowtwins{} validation loss is logged but
\emph{not} optimised against (range $0.004$ to $0.046$).

\paragraph{Fitting procedure.} Within each isoFLOP slice we fit a
quadratic to the downstream score as a function of $\log N$
(encoder or projector); the vertex is the compute-optimal size at
that compute level. A log--log least-squares fit through the
per-slice vertices gives the power laws of
\Cref{sec:scaling:asymmetry}. The encoder and data
laws are fit on all \nbuckets{} slices (compute-optimal $=$ argmax
composite within the slice). The projector optimum is only
well-resolved once the downstream score clears the noise floor, so
the projector law is fit on the mid-to-high compute slices rather
than all \nbuckets{}. Uncertainty is a
$2000$-resample bootstrap over the per-slice vertices; robustness
is assessed by leave-one-slice-out (LOBO) refits. The resulting
exponents, confidence intervals, and LOBO spreads are reported in
\Cref{tab:scaling-exponents}: the encoder exponent is $\encexp$ and
the projector exponent is statistically indistinguishable from zero.
The data exponent is not fit independently: on the isoFLOP vertices,
fixing $N^{\star}$ pins $D^{\star}$ through $C\!\propto\!N\!\cdot\!D$,
so $\encexp$ and $\dataexp$ necessarily sum to $\approx\!1$ (here
$0.99$). The substantive quantity is therefore the encoder exponent
itself; the sum matches the \citet{hoffmann2022chinchilla} balance by
construction. Pretraining loss and downstream score are weakly
anti-correlated over all $\nruns$ runs (Pearson $r\!=\!-0.18$,
Spearman $\rho\!=\!-0.16$).

\paragraph{A data-heavy allocation.} While the exponent \emph{sum} is
fixed by the isoFLOP constraint, the \emph{split} is not, and it is
informative. The compute-optimal allocation is markedly data-heavy:
at the fitted exponents, doubling the budget calls for about
$2^{\encexp}\!\approx\!1.28\times$ the encoder but
$2^{\dataexp}\!\approx\!1.55\times$ the data, so the compute-optimal
frontier grows data faster than parameters
($D^{\star}\!\propto\!N^{\star\,\dataexp/\encexp}$, exponent
$\approx\!1.75$). This departs from the roughly balanced
parameter/data split \citet{hoffmann2022chinchilla} report for
language. We have no first-principles account, but the likely cause
is informational: a single \dpixel{} carries far less independent
signal than a text token, since cloud cover and repeated orbital
passes make consecutive observations highly redundant, so useful
supervision accrues mainly through data volume rather than encoder
capacity. It is also the tilt behind the enlarged teacher's
correspondingly large training corpus (\Cref{sec:teacher}).

\paragraph{Large-model validation.} Guided by the fitted law we train
three teachers at $0.5$B/$1$B/$2$B encoders on the compute-optimal
data prescribed by $D^{\star}(C)$ (the $2$B teacher uses ${\sim}14$B
\dpixels). Evaluated on the same suite they score
$0.572/0.596/0.608$, landing on the extrapolated small-model curve
(\Cref{fig:scaling-laws}f) and surpassing \tesseravone{} and
\alphaearth{}; distillation to compact students
(\Cref{fig:scaling-laws}f, inset) costs ${\le}0.015$ downstream
score.

\paragraph{Within-bucket loss--score correlation.} The pooled Pearson
$r\!=\!-0.18$ of \Cref{sec:scaling:loss} mixes runs across four orders
of magnitude of compute and four projector widths, which dilutes any
within-budget relationship. The quantity that actually governs model
selection is the loss--score correlation \emph{inside} a fixed compute
budget. \Cref{tab:within-bucket-rho} reports the Spearman $\rho$
within each of the \nbuckets{} isoFLOP buckets: the values scatter
around zero (mean $-0.08$, median $-0.13$, range $[-0.25,+0.07]$), and
no bucket reaches a correlation strong enough to rank runs reliably by
loss. Controlling for compute therefore does not uncover a hidden
predictive signal; the loss is a weak proxy for downstream utility at
every budget, not only in aggregate. \textbf{F1} rests on this
within-budget statement, so removing the pooling confound strengthens
it.

\begin{table}[!htbp]
  \centering
  \caption{\textbf{Loss--score Spearman $\rho$ within each isoFLOP
  bucket}, over the runs in each bucket of
  \Cref{tab:isoflop-slices}. A value near zero means the
  \barlowtwins{} loss does not rank runs by downstream score at that
  compute budget; every bucket is close to zero.}
  \label{tab:within-bucket-rho}
  \small
  \begin{tabular}{@{}rccr@{}}
  \toprule
  Bucket & $C$ centre (FLOPs) & $n$ & Spearman $\rho$ \\
  \midrule
  1 & $1.3\times10^{15}$ & $17$ & $-0.245$ \\
  2 & $3.7\times10^{15}$ & $39$ & $-0.230$ \\
  3 & $1.1\times10^{16}$ & $58$ & $+0.041$ \\
  4 & $3.2\times10^{16}$ & $48$ & $-0.145$ \\
  5 & $9.6\times10^{16}$ & $52$ & $+0.016$ \\
  6 & $2.8\times10^{17}$ & $57$ & $+0.066$ \\
  7 & $8.3\times10^{17}$ & $44$ & $-0.130$ \\
  8 & $2.5\times10^{18}$ & $47$ & $-0.137$ \\
  9 & $7.3\times10^{18}$ & $33$ & $+0.035$ \\
  \midrule
  \multicolumn{3}{@{}l}{mean $-0.08$; median $-0.13$} & $[-0.25,+0.07]$ \\
  \bottomrule
  \end{tabular}
\end{table}

\paragraph{Compute cost of loss-based model selection.} The
$\approx\!\wastemult$ figure quoted in
\Cref{sec:scaling:loss} is obtained by converting a downstream
\emph{score} gap into a \emph{compute} multiple, in three steps.
\emph{(i)~Per-slice score gap.} In each of the \nbuckets{} isoFLOP
slices we compare two ways of picking a single run: the
\emph{score-optimal} run (argmax composite) and the
\emph{loss-optimal} run (argmin normalised \barlowtwins{} loss).
Because loss and downstream score are nearly uncorrelated, the
loss-optimal run scores lower; averaged over slices the deficit is
\begin{equation}
\overline{\Delta s} \;=\;
\tfrac{1}{\text{\#slices}}\sum_{\text{slices}}
\bigl(s_{\mathrm{score\text{-}opt}} - s_{\mathrm{loss\text{-}opt}}\bigr)
\;=\; 0.0486 .
\end{equation}
\emph{(ii)~Score--compute slope.} Fitting the score-optimal
per-slice peak scores against $\log_{10} C$ gives the marginal
return of compute, the \emph{score envelope} slope
\begin{equation}
b_s \;=\;
\frac{\mathrm{d}\,s}{\mathrm{d}\log_{10} C} \;=\; 0.0886
\quad\text{(score per compute decade).}
\end{equation}
\emph{(iii)~Translate score into compute.} Recovering a score
deficit $\overline{\Delta s}$ by adding compute requires
$\overline{\Delta s}/b_s$ decades, so matching downstream-driven
selection by loss-based selection costs a compute multiple
\begin{equation}
10^{\,\overline{\Delta s}/b_s}
\;=\; 10^{\,0.0486/0.0886}
\;=\; 10^{\,0.548}
\;\approx\; 3.5\times ,
\label{eq:waste}
\end{equation}
i.e.\ an excess (``wasted'') fraction of
$\bigl(10^{\,\overline{\Delta s}/b_s}-1\bigr)\times 100\%
\approx\wasteexcess$. Both inputs ($\overline{\Delta s}$ and $b_s$) are estimated from
\nbuckets{} slice peaks and are sensitive to loss noise, and two
effects inflate the point estimate. First, the argmax/argmin choice in
step~(i) is a winner's-curse estimator: on a subset of configurations
(three per bucket, three seeds each) the composite score has a seed
standard deviation of about $0.008$, and selecting the extreme run
over that noise systematically enlarges $\overline{\Delta s}$. Second,
$\overline{\Delta s}$ and $b_s$ are each estimated from only
\nbuckets{} points; bootstrapping them over the slices puts the excess
fraction in a $95\%$ interval of roughly $[156\%,\,451\%]$, that is, a
compute multiple of about $2.6$ to $5.5\times$.

\section{Architecture Details}
\label{app:arch}
\paragraph{Teacher.} The teacher is an all-Transformer dual-branch
pixel-wise encoder. Each modality branch (\stwo, $10$ optical
bands; \sone, VV/VH with ascending and descending merged) applies a
two-layer input MLP ($\mathrm{Linear}\!\to\!\mathrm{ReLU}\!\to\!\mathrm{Linear}$),
a sinusoidal day-of-year positional
encoding, and a four-layer Transformer encoder, followed by learned
attention pooling. The two per-modality representations are
augmented with learned modality embeddings and fused by a two-layer
cross-modal Transformer, then reduced to a
$\dimteacher$-dimensional embedding. A batch-normalised MLP
projector ($9$ hidden layers, width $4096$) is attached for the
\barlowtwins{}/mix-up objective and discarded at inference. The
deployed teacher used for all reported results is the
$\teacherparams$ member of a width-scaled family: its per-modality
encoders use model width $d_{\mathrm{model}}\!=\!4096$ ($=\!4\times$
the latent dimension $1024$), four heads, a $16384$-wide
feed-forward, and a representation dimension
$\dimteacher\!=\!1024$. Only the Transformer width scales across the
family (\Cref{tab:teacher-family}); depth, head count, fusion,
pooling, and projector form are fixed. The width is chosen at the
compute-optimal point identified in \Cref{sec:scaling}.

\begin{table}[!htbp]
  \centering
  \caption{\textbf{Teacher model family.} Only the Transformer
  width scales; the $\teacherparams$ teacher is used for all
  reported results. The \barlowtwins{} projector ($\sim\!170$\,M)
  is discarded at inference and is approximately width-invariant
  (\Cref{sec:scaling:asymmetry}), so it is a shrinking fraction of
  the total as the encoder grows ($25\%\!\to\!14\%\!\to\!8\%$).}
  \label{tab:teacher-family}
  \small
  \begin{tabular}{lccc}
  \toprule
  & $500$\,M & $1$\,B & $2$\,B (reported) \\
  \midrule
  Latent dim                       & $512$  & $768$  & $1024$ \\
  Model width $d_{\mathrm{model}}$ & $2048$ & $3072$ & $4096$ \\
  Feed-forward width               & $8192$ & $9216$ & $16384$ \\
  Representation dim $\dimteacher$  & $512$  & $768$  & $1024$ \\
  \midrule
  Encoder (deployed)               & $0.52$\,B & $1.01$\,B & $2.06$\,B \\
  Projector (SSL head, discarded)  & $170$\,M & $171$\,M & $172$\,M \\
  Total (training)                 & $0.69$\,B & $1.18$\,B & $2.24$\,B \\
  \bottomrule
  \end{tabular}
\end{table}

\paragraph{Pixel-wise students (N/S/M/L).} The students share a
single design: two per-modality Transformer backbones
(\stwo{} and merged \sone) with sinusoidal DoY encoding and
attention pooling, whose outputs are concatenated and passed
through an MLP dimensionality reducer
($\mathrm{Linear}\!\to\!\mathrm{LayerNorm}\!\to\!\mathrm{ReLU}\!\to\!\mathrm{Dropout}\!\to\!\mathrm{Linear}$)
to a $\dimfull$-dimensional \matryoshka{} embedding, followed by a
final affine-free LayerNorm as in the teacher
(\Cref{sec:teacher:arch}). The four
sizes differ only in backbone width and depth
(\Cref{tab:student-arch}); the large student is wider rather than
deeper. The training-only \matryoshka{} projection heads add a fixed
$\sim\!2.35$\,M parameters (\Cref{tab:student-arch}), independent of
student size and discarded at inference, so only the deployed encoder
counts at serving time.

\begin{table}[!htbp]
  \centering
  \caption{\textbf{Pixel-wise student family.} All four emit a
  $\dimfull$-dimensional \matryoshka{} embedding; only backbone
  width and depth differ. Encoder parameters are the deployed
  count (prefix heads discarded at inference).}
  \label{tab:student-arch}
  \footnotesize
  \setlength{\tabcolsep}{3pt}
  \resizebox{\linewidth}{!}{%
  \begin{tabular}{lcccc}
  \toprule
  & \tesseravtwo-2B-N & \tesseravtwo-2B-S & \tesseravtwo-2B-M & \tesseravtwo-2B-L \\
  \midrule
  Latent dim                       & $36$  & $64$   & $110$  & $160$ \\
  Backbone width $d_{\mathrm{model}}$ & $144$ & $256$ & $440$ & $640$ \\
  Layers per backbone              & $2$   & $4$    & $4$    & $4$ \\
  Attention heads                  & $4$   & $4$    & $4$    & $4$ \\
  Feed-forward width               & $384$ & $1024$ & $1792$ & $2560$ \\
  Output dim (\matryoshka)          & $128$ & $128$  & $128$  & $128$ \\
  \midrule
  Encoder params (deployed)        & $1.1$\,M & $7.1$\,M & $21.0$\,M & $43.8$\,M \\
  Total params (training)          & $3.4$\,M & $9.5$\,M & $23.4$\,M & $46.2$\,M \\
  \bottomrule
  \end{tabular}}
\end{table}

\paragraph{\matryoshka{} heads.} For each prefix
$k\!\in\!\dimset$, a small MLP head
$h_k\colon\R^{k}\!\to\!\R^{\dimteacher}$ maps the student prefix
into the teacher embedding space (\Cref{eq:matryoshka-distill}).
The heads are used only during distillation and are discarded at
inference; the deployed student outputs the raw
$\dimfull$-dimensional \matryoshka{} embedding.

\section{Teacher Pretraining Details}
\label{app:training:teacher}
The teacher is pretrained for a single epoch over approximately
$14$~billion pixel-wise multi-temporal \dpixels{} with the
\barlowtwins~$+$~mix-up objective of \Cref{sec:teacher:obj}.
Temporal augmentation draws two independent variable-length
sub-sequences per \dpixel{} (\Cref{sec:teacher:obj}); each modality
is additionally dropped out as a whole with a small probability.
The training corpus is organised into buckets by valid-observation
count, and each step samples across buckets so that a single epoch
exposes the model to the full range of temporal coverage rather
than to a fixed sequence length. \Cref{tab:teacher-hparams} lists
the optimisation and systems configuration; the learning-rate
schedule has no plateau phase (linear warmup directly into cosine
decay).

\begin{figure}[!htbp]
  \centering
  \includegraphics[width=\linewidth]{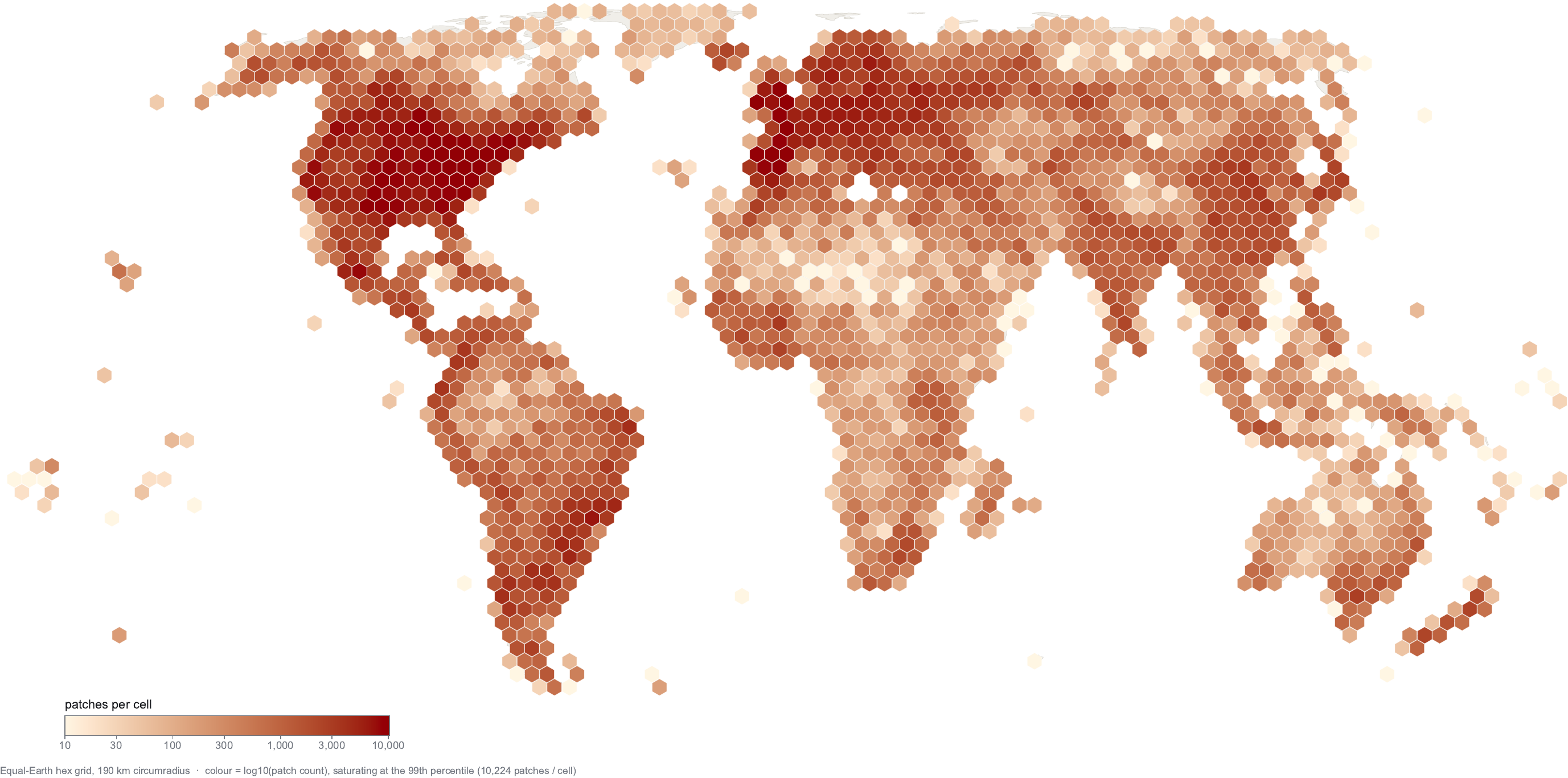}
  \caption{\textbf{Geographic distribution of the pretraining corpus.}
  Each cell of an Equal-Earth hexagonal grid ($190$\,km circumradius)
  is coloured by the number of $64\times64$ \sonetwo{} patches it
  contains ($\log_{10}$ scale, saturating at the $99$th percentile of
  $10{,}224$ patches per cell). The corpus comprises $3{,}420{,}048$
  patches; because \tesseravtwo{} is trained pixel-wise, the
  ${\sim}14$ billion training \dpixels{} are obtained by randomly
  sampling pixels from within these patches.}
  \label{fig:patch-density}
\end{figure}

\Cref{fig:patch-density} maps the geographic distribution of the
pretraining corpus.

\begin{table}[!htbp]
  \centering
  \caption{\textbf{Teacher pretraining configuration.}}
  \label{tab:teacher-hparams}
  \small
  \begin{tabular}{ll}
  \toprule
  Setting & Value \\
  \midrule
  Parameters & $\teacherparams$ \\
  Modalities & \stwo{} ($10$ bands); \sone{} (VV/VH, ASC$+$DESC merged) \\
  Self-supervised objective & \barlowtwins{}
    ($\lambda_{\mathrm{BT}}\!=\!5\!\times\!10^{-3}$) $+$ mix-up
    ($\alpha\!\sim\!\mathrm{Beta}(1,1)$, $\lambda_{\mathrm{mix}}\!=\!1.0$) \\
  Teacher embedding dim $\dimteacher$ & $1024$ \\
  Optimiser & AdamW ($\beta_1\!=\!0.9$, $\beta_2\!=\!0.999$,
    $\epsilon\!=\!10^{-8}$) \\
  Weight decay & $10^{-6}$ \\
  Peak learning rate & $2\!\times\!10^{-4}$ \\
  LR schedule & linear warmup ($10\%$ of steps) $\to$ cosine decay to $0$ \\
  Per-GPU / global batch & $256$ / $131{,}072$ \\
  Epochs & $1$ (single pass over $\sim\!14$\,B \dpixels{}) \\
  Mixed precision & AMP, \texttt{bf16} \\
  Parallelism & FSDP (hybrid shard: intra-node \texttt{FULL\_SHARD},
    inter-node DDP) \\
  Hardware & $512$ GPUs ($128\!\times\!4$-GPU GB200 NVL72-class nodes) \\
  \bottomrule
  \end{tabular}
\end{table}

\paragraph{Mixed precision.} Pretraining runs entirely in
automatic mixed precision with \texttt{bfloat16}: the FSDP
\texttt{MixedPrecision} policy sets both the parameter dtype and
the communication-reduction dtype to \texttt{bf16}, and the
training loop wraps the forward and loss computation in
\texttt{torch.autocast(bf16)}. On GB200-class tensor cores
\texttt{bf16} delivers roughly $1.9\times$ the throughput of
\texttt{fp32} while leaving the learned representation essentially
unchanged (cosine similarity to an \texttt{fp32} reference
$\approx\!1.0$), so the speed-up comes at negligible numerical
cost.

\paragraph{Systems stack.} Pretraining is implemented on a
PhysicsNeMo-based training stack (\texttt{torch.compile} under
FSDP, a streaming data pipeline, topology-matched sharding, and
fused-kernel/NCCL/NUMA tuning). These are systems-level
optimisations only: the model, loss, optimiser, regularisation,
batch, and data are unchanged, so they affect wall-clock
throughput but not the learned representation. They make a single
epoch over $14$\,B \dpixels{} at $512$ GPUs practical.

\section{Distillation Details}
\label{app:training:distill}
The pixel-wise students are distilled against cached teacher
embeddings (\Cref{sec:pixel:objective}) over $1.2$ billion
pixel-wise \dpixels{} covering diverse global land cover and
phenology; the data-scaling behaviour behind this budget is in
\Cref{app:distill-scaling}. \stwo{} optical bands are standardised with
per-band statistics; \sone{} ascending and descending passes are
standardised with their own statistics before being merged.

\paragraph{Per-pixel binning.} Valid-observation counts vary widely
across pixels. For batch efficiency we bin each pixel's observation
count to a fixed ladder of bucket boundaries
$\{8, 16, 24, \ldots, 256\}$, rounding up to the nearest bucket and
filling the residual slots with a split-and-repeat pattern (pixels
exceeding the largest bucket are down-sampled); binning is applied
independently to the optical and SAR branches, and pixels sharing
the same $(\stwo, \sone)$ bucket pair are processed in one forward
call. This is the distillation-time analogue of the inference
bucket sampler of \Cref{sec:teacher:bucket}.

\paragraph{Cross-patch shuffle.} To break the spatial correlation
between pixels of the same image patch and maximise per-batch
statistical diversity, we draw each batch by uniformly sampling
pixels from a buffer of many patches, with each pixel sampled
exactly once per epoch---the pixel-level analogue of \tesseravone's
global \dpixel{} shuffling.

\paragraph{Distributed implementation.} Because each batch yields
many variable-length $(\stwo, \sone)$ bucket groups, each requiring
its own forward pass, the per-forward gradient-bucket bookkeeping
of standard \texttt{DistributedDataParallel} becomes a bottleneck.
We instead broadcast rank-$0$ weights at start-up and, after each
local backward pass, flatten all parameter gradients into one
contiguous buffer and perform a single all-reduce averaged by world
size---equivalent to DDP's averaged gradient but with one
collective per step. Training runs as a single flat
\texttt{while}-loop over global steps with no epoch barriers; all
ranks issue an identical collective sequence each step, and on a
non-finite loss the step's gradients are zeroed but still
all-reduced to keep the collective sequence aligned across ranks.
\Cref{tab:distill-hparams} lists the configuration.

\begin{table}[!htbp]
  \centering
  \caption{\textbf{Distillation configuration.}}
  \label{tab:distill-hparams}
  \small
  \begin{tabular}{ll}
  \toprule
  Setting & Value \\
  \midrule
  Hardware & $64\times$ \gputype{} \\
  Distillation data & $1.2$\,B pixel-wise \dpixels{} \\
  Teacher targets & cached $\dimteacher\!=\!1024$-d embeddings (fp16) \\
  Student output & $\dimfull$-d \matryoshka{} ($\dimset$) \\
  Loss & $\sum_{k\in\dimset}\bigl(1-\cos(h_k(\bs_{1:k}),\bt)\bigr)$ \\
  Per-GPU batch & $8192$ pixels \\
  Optimiser & AdamW ($\beta_1\!=\!0.9$, $\beta_2\!=\!0.95$) \\
  Peak learning rate & $8\!\times\!10^{-4}$ \\
  LR schedule & linear warmup ($5\%$ of steps) $\to$ cosine decay to $0$ \\
  Weight decay & $10^{-4}$ \\
  Gradient clipping & $1.0$ (global norm) \\
  Dropout & $0.1$ \\
  Mixed precision & \texttt{bf16} autocast \\
  \bottomrule
  \end{tabular}
\end{table}

\section{Distillation Data Scaling}
\label{app:distill-scaling}
A distilled student inherits its quality from two sources: the teacher
it copies, and how efficiently distillation transfers that teacher
into the compact student. The first axis, the teacher's own scaling,
is characterised in \Cref{sec:scaling} and \Cref{fig:scaling-laws}f.
Here we isolate the second. Holding the teacher fixed at
\tesseravtwo{}-2B, we vary the data seen during distillation from
$0.1$ to $1.2$ billion \dpixels{} and track two quantities: how
closely the large student reproduces the teacher, measured by the cosine
similarity of each \matryoshka{} prefix to the teacher's
$\dimteacher{=}1024$-d embedding, and downstream quality on the
$\npixeldata$-task \aebench{}.

\begin{figure}[!htbp]
  \centering
  \includegraphics[width=\linewidth]{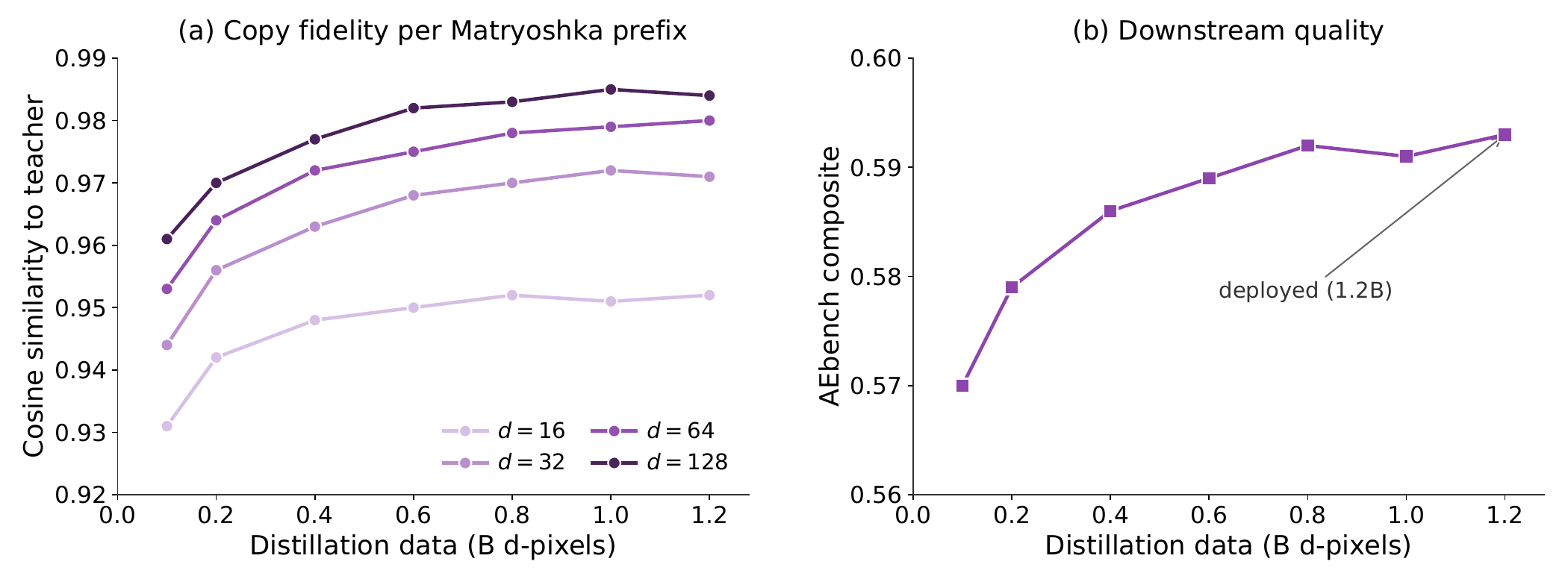}
  \caption{\textbf{Distillation data scaling}, teacher fixed at
  \tesseravtwo{}-2B-L. \textbf{(a)}~Cosine similarity between each
  \matryoshka{} prefix ($d\!\in\!\dimset$) and the $1024$-d teacher
  embedding. \textbf{(b)}~\aebench{} composite over the $\npixeldata$
  tasks. Both saturate near $0.6$\,B \dpixels{}; the deployed student
  uses $1.2$\,B.}
  \label{fig:distill-scaling}
\end{figure}

Both curves rise steeply and then flatten (\Cref{fig:distill-scaling},
\Cref{tab:distill-scaling}). Most of the transfer is done within the
first $0.6$\,B \dpixels{}: past that point every prefix sits within
about $0.005$ of its final cosine similarity, and the \aebench{}
composite has already reached $0.589$ of its eventual $0.593$. The
wider prefixes copy the teacher more faithfully at every budget, from
$0.952$ at $d{=}16$ to $0.984$ at $d{=}128$, as their greater capacity
to match the $1024$-d target would suggest.

We still distil the deployed student on the full $1.2$\,B \dpixels{}.
Distillation reads cached teacher embeddings and updates only the student, so it is far cheaper than teacher pretraining
(\Cref{app:deploy-cost}); the gain past $0.6$\,B is marginal but
almost free in wall-clock, so we take it. This is the reverse of the
pretraining-data trade-off of \Cref{sec:scaling}, where every extra
\dpixel{} is costly and must be weighed against encoder size.

\begin{table}[!htbp]
  \centering
  \caption{\textbf{Distillation data scaling} (teacher fixed at
  \tesseravtwo{}-2B-L). Cosine similarity of each \matryoshka{} prefix
  to the $1024$-d teacher and the $\npixeldata$-task \aebench{}
  composite, versus distillation \dpixels{}. The deployed run (bold)
  uses $1.2$\,B.}
  \label{tab:distill-scaling}
  \small
  \begin{tabular}{rccccc}
  \toprule
  Distill.\ data (B) & Cos-$16$ & Cos-$32$ & Cos-$64$ & Cos-$128$ & \aebench{} \\
  \midrule
  $0.1$ & $0.931$ & $0.944$ & $0.953$ & $0.961$ & $0.570$ \\
  $0.2$ & $0.942$ & $0.956$ & $0.964$ & $0.970$ & $0.579$ \\
  $0.4$ & $0.948$ & $0.963$ & $0.972$ & $0.977$ & $0.586$ \\
  $0.6$ & $0.950$ & $0.968$ & $0.975$ & $0.982$ & $0.589$ \\
  $0.8$ & $0.952$ & $0.970$ & $0.978$ & $0.983$ & $0.592$ \\
  $1.0$ & $0.951$ & $0.972$ & $0.979$ & $0.985$ & $0.591$ \\
  $\mathbf{1.2}$ & $\mathbf{0.952}$ & $\mathbf{0.971}$ & $\mathbf{0.980}$ & $\mathbf{0.984}$ & $\mathbf{0.593}$ \\
  \bottomrule
  \end{tabular}
\end{table}

\section{The \texorpdfstring{$\npixeldata$}{15} \aebench{}
Evaluation Tasks}
\label{app:datasets}
The $\npixeldata$ downstream tasks used throughout this paper are
drawn from $10$ source datasets and cover classification,
regression, and change detection across multiple
regions and time horizons (\Cref{tab:datasets}). The same task
list underwrites both the downstream-driven scaling-law evaluation
(\Cref{sec:scaling}) and the pixel-wise benchmark
(\Cref{sec:pixel:bench}); we do not introduce a separate evaluation
suite for the benchmark in order to keep the scaling-law fits and
the headline results commensurate.

\begin{table}[!htbp]
  \centering
  \caption{\textbf{The $\npixeldata$ \aebench{} evaluation tasks,
  grouped by source dataset.}
  Tasks span land cover, land use, change detection,
  crop typing, biodiversity mapping, and physical-attribute
  regression. Metrics follow the chance-adjusted convention of
  \Cref{sec:scaling:design} (classification: chance-adjusted
  balanced accuracy; regression: $\max(0,R^{2})$). Parenthesised
  values are the maximum number of labelled training samples
  (shots) per task.}
  \label{tab:datasets}
  \footnotesize
  \setlength{\tabcolsep}{4pt}
  \begin{tabular}{@{}p{2.0cm}p{3.3cm}p{2.0cm}p{5.7cm}@{}}
    \toprule
    Source dataset & Evaluation task(s) & Region~/~time &
    Brief description \\
    \midrule
    LCMAP &
    land cover (300); land use (300); land-cover change (300);
    land-use change (150) &
    CONUS, annual &
    USGS reference data for the contiguous United States; used to
    evaluate national-scale land-cover and land-use
    classification, and annual change detection. \\
    \addlinespace[2pt]
    LUCAS &
    land cover (300); land use (300) &
    Europe, single-date / annual &
    EU ground-survey points with fine taxonomy and large sample
    size; tests fine-grained ground-concept discrimination. \\
    \addlinespace[2pt]
    GLaNCE &
    global land cover (300) &
    Global, annual &
    Global land-cover training data; tests planetary-scale
    general-purpose land-cover mapping. \\
    \addlinespace[2pt]
    Africa crop mask &
    cropland / non-cropland (200) &
    Sub-Saharan Africa, annual &
    Photo-interpreted cropland reference across multiple
    Sub-Saharan African countries; tests agricultural-extent
    classification in regions where existing maps disagree
    strongly. \\
    \addlinespace[2pt]
    Canada / AAFC &
    coarse crop type (68); fine crop type (75) &
    Canada, single-date &
    AAFC in-field crop-type survey; tests hierarchical
    (coarse-/fine-grained) crop classification at national scale. \\
    \addlinespace[2pt]
    Ethiopia / EthCT2020 &
    crop type (49) &
    Ethiopia, annual &
    Smallholder field-level crop labels (wheat, barley, maize,
    teff, \ldots); tests generalisation to complex smallholder
    systems with sparse high-quality labels. \\
    \addlinespace[2pt]
    US trees (GBIF / iNaturalist) &
    tree genera, $39$ classes (300) &
    United States, single-date &
    GBIF / iNaturalist genus-level observations of $39$ tree
    genera; tests biodiversity / species-composition mapping. \\
    \addlinespace[2pt]
    Descals oil palm &
    other / industrial / smallholder oil palm, 3 classes (200) &
    Global, annual &
    Global oil-palm plantation reference, three-class; tests
    fine industrial-vs-smallholder discrimination of a tropical
    commodity crop. \\
    \addlinespace[2pt]
    OpenET ensemble &
    evapotranspiration, regression (300) &
    Western US, monthly &
    Monthly OpenET ET ensemble product; used as a proxy
    regression task that probes whether the embeddings can
    approximate a complex multi-model ensemble output. \\
    \addlinespace[2pt]
    ASTER GED &
    surface emissivity, regression (200) &
    Global, annual &
    $100\,\mathrm{m}$ ASTER Global Emissivity Database; tests the
    representation of a continuous physical surface attribute. \\
    \bottomrule
  \end{tabular}
\end{table}

\section{Held-out Generalisation Suite}
\label{app:heldout}
The $\npixeldata$ \aebench{} tasks were available throughout model
development, so good numbers on them leave open how the embeddings
behave on data the design never saw. We therefore assembled a second
suite of $\nheldout$ datasets that we did not consult while building
\tesseravtwo{}, and ran only the four strongest systems on it. The
suite splits into a \emph{vegetation} group (tree species, crop and
parcel mapping, forest biomass) and an \emph{urban} group (land use
and a range of socioeconomic and environmental layers), and it
reaches countries and label types that the \aebench{} tasks
under-represent (\Cref{tab:heldout}).

\paragraph{Why we treat the held-out suite as primary.} The two suites
are not equally neutral for comparing systems. We used the $\npixeldata$
\aebench{} tasks during development to choose v2 architectures and
checkpoints, so our own numbers on them are selected, in part, to look
good. Those tasks also come from widely used public label sets, and we
cannot assume the baselines avoided them: \alphaearth{} does not
disclose its training corpus, so its
\aebench{} labels may have been seen during its own training. A system
that has already seen a task's labels, directly or through a close
proxy, holds an advantage that says little about generalisation. None of
the $\nheldout$ held-out datasets were consulted while building
\tesseravtwo{}, and they fall outside the label sources the \aebench{}
draws on, so we report the held-out composite as the headline number and
use the \aebench{} to describe in-distribution behaviour.

\begin{table}[!htbp]
  \centering
  \caption{\textbf{The $\nheldout$ held-out generalisation datasets},
  none used during model development. Metrics are mapped to a common
  $[0,1]$ higher-is-better scale before averaging (text below);
  $K$ is the class count for chance-anchored metrics. All
  \textsc{CityRep} tasks come from the same multi-city
  benchmark~\citep{liu2026cityrep}.}
  \label{tab:heldout}
  \footnotesize
  \setlength{\tabcolsep}{4pt}
  \begin{tabular}{@{}p{2.5cm}p{3.0cm}p{1.9cm}p{5.5cm}@{}}
    \toprule
    Dataset & Task (metric) & Region / group & Brief description \\
    \midrule
    TreeSatAI-TS~\citep{astruc2024omnisat} &
    tree species, multi-label (mAP) & Germany / veg. &
    Species labels over Sentinel time series; tests fine
    species discrimination. \\
    \addlinespace[2pt]
    PASTIS-R~\citep{garnot2021mmfusion} &
    parcel segmentation (mIoU, $K{=}18$) & France / veg. &
    Agricultural-parcel labels with radar; tests dense crop-type
    segmentation. \\
    \addlinespace[2pt]
    Austrian Crop~\citep{feng2026tessera} &
    crop segmentation (mIoU, $K{=}17$) & Austria / veg. &
    National crop semantic segmentation. \\
    \addlinespace[2pt]
    BioMassters~\citep{nascetti2023biomassters} &
    above-ground biomass ($R^2$) & Finland / veg. &
    Forest biomass regression from multi-modal time series. \\
    \addlinespace[2pt]
    CropHarvest~\citep{tseng2021cropharvest} &
    crop / non-crop (accuracy, $K{=}2$) & Togo; China / veg. &
    Two regional splits of a global crop benchmark; tests transfer
    to under-mapped smallholder systems. \\
    \addlinespace[2pt]
    \textsc{CityRep}-LUC &
    land use (macro-F1, $K{=}12$) & cities / urban &
    Zoning / land-use classes (residential, commercial, industrial,
    green space, \ldots); tests urban semantic function. \\
    \addlinespace[2pt]
    \textsc{CityRep}-RDE &
    road density ($R^2$) & cities / urban &
    OpenStreetMap road network; tests physical street structure. \\
    \addlinespace[2pt]
    \textsc{CityRep}-POP &
    population ($R^2$) & cities / urban &
    WorldPop gridded population; tests spatial population intensity. \\
    \addlinespace[2pt]
    \textsc{CityRep}-GDP &
    GDP ($R^2$) & cities / urban &
    Gridded GDP product; tests spatial economic output. \\
    \addlinespace[2pt]
    \textsc{CityRep}-PM2.5 &
    PM\textsubscript{2.5} ($R^2$) & cities / urban &
    Annual surface PM\textsubscript{2.5}; tests pollution exposure. \\
    \addlinespace[2pt]
    \textsc{CityRep}-LST &
    surface temperature ($R^2$) & cities / urban &
    MODIS daytime land-surface temperature; tests surface thermal
    conditions. \\
    \addlinespace[2pt]
    \textsc{CityRep}-NTL &
    nighttime lights ($R^2$) & cities / urban &
    VIIRS annual nighttime-lights product; tests human activity,
    electrification, and infrastructure intensity. \\
    \addlinespace[2pt]
    \textsc{CityRep}-AGE &
    age distribution ($e^{-\mathrm{KL}}$) & cities / urban &
    WorldPop age structure; tests demographic composition. \\
    \bottomrule
  \end{tabular}
\end{table}

\paragraph{Scoring and the full-suite composite.} Each held-out
dataset is evaluated at its native metric and mapped to $[0,1]$ under
a higher-is-better convention before any averaging, mirroring the
chance-adjusted treatment of the held-in tasks
(\Cref{sec:scaling:design}). Classification and segmentation use the
chance-anchored map $(s-1/K)/(1-1/K)$ for $K$ classes (CropHarvest
$K{=}2$, \textsc{CityRep}-LUC $K{=}12$, PASTIS-R $K{=}18$, Austrian
Crop $K{=}17$); metrics already on a natural scale use
$\max(0,R^2)$ for the regressions and BioMassters and
$\max(0,\mathrm{mAP})$ for TreeSatAI-TS; and the age-distribution
match uses $e^{-\mathrm{KL}}$. Averaging the $\nheldout$ normalised
scores with uniform weight gives a held-out composite, which folds
into the held-in composite as
$c_{\nfullsuite}=(\npixeldata\,c_{\npixeldata}+\nheldout\,c_{\nheldout})/\nfullsuite$.
Every one of the $\nfullsuite$ datasets then carries weight
$1/\nfullsuite$, so the full-suite number extends the
$\npixeldata$-task composite rather than re-weighting it. A system
with no prediction for a dataset scores zero there, as in the
held-in composite.

\paragraph{Label efficiency.} On the held-out datasets we also vary
how much supervision the task head receives. For each dataset we
train at $1\%$, $30\%$, and $100\%$ of the available labels with two
heads of very different capacity, a linear probe and a small
${\sim}2$M-parameter CNN, and read off, per dataset, the gap to the
best non-v2 baseline trained the same way (\Cref{fig:pixel-bench}f).
The \tesseravtwo{} students lead on most datasets at every budget,
and the margin is widest at $1\%$, where the quality of the frozen
features matters most and a strong pretrained representation stands
in for labels that are not there.

\paragraph{Per-task variability.} \Cref{tab:heldout-veg-std} gives
the primary metric on each of the six non-\textsc{CityRep} held-out
datasets, with its standard deviation over five seeds. The spread is
small on these single-region tasks, at most a few points, so the
composite differences of \Cref{app:heldout-robust} are not seed
noise. \tesseravtwo{} is best on four of the six, with its widest
margins on the dense segmentation tasks (PASTIS-R, Austrian Crop). On
BioMassters and CropHarvest-Togo a baseline edges ahead, but in both
cases the gap is smaller than a seed standard deviation. The eight
\textsc{CityRep} tasks and their cross-city deviation are tabulated
separately in \Cref{tab:cityrep-linear}.

\begin{table}[!htbp]
  \centering
  \caption{\textbf{Per-task scores and seed variability on the six
  non-\textsc{CityRep} held-out datasets.} Primary metric for the four
  strongest systems, with the standard deviation over five seeds as a
  subscript; the best per task is in bold. Metrics are mAP
  (TreeSatAI-TS), mIoU (PASTIS-R, Austrian Crop), $R^2$
  (BioMassters), and accuracy (CropHarvest); \tesseravtwo{} uses the
  $d\!=\!128$ student. These are the raw scores behind the held-out
  composite (\Cref{app:heldout-robust}).}
  \label{tab:heldout-veg-std}
  \small
  \setlength{\tabcolsep}{4pt}
  \begin{tabular}{@{}lcccccc@{}}
  \toprule
  Model & TreeSatAI & PASTIS-R & Austrian & BioMass. & CH-Togo & CH-China \\
        & mAP$\uparrow$ & mIoU$\uparrow$ & mIoU$\uparrow$ & $R^2\!\uparrow$ & acc$\uparrow$ & acc$\uparrow$ \\
  \midrule
  \tesseravtwo{}-2B-L & $\mathbf{0.832}_{\,0.018}$ & $\mathbf{0.666}_{\,0.021}$ & $\mathbf{0.461}_{\,0.031}$ & $0.813_{\,0.019}$ & $0.767_{\,0.020}$ & $\mathbf{0.805}_{\,0.022}$ \\
  \tesseravone{}      & $0.816_{\,0.011}$ & $0.536_{\,0.023}$ & $0.455_{\,0.027}$ & $0.801_{\,0.009}$ & $\mathbf{0.780}_{\,0.014}$ & $0.722_{\,0.008}$ \\
  \alphaearth{}       & $0.783_{\,0.028}$ & $0.542_{\,0.025}$ & $0.236_{\,0.007}$ & $\mathbf{0.816}_{\,0.033}$ & $0.774_{\,0.032}$ & $0.785_{\,0.024}$ \\
  OlmoEarth-L         & $0.798_{\,0.010}$ & $0.512_{\,0.006}$ & $0.385_{\,0.013}$ & $0.804_{\,0.029}$ & $0.778_{\,0.015}$ & $0.725_{\,0.012}$ \\
  \bottomrule
  \end{tabular}
\end{table}

\section{Robustness of the Held-out Ranking}
\label{app:heldout-robust}
\tesseravone{} embeddings have already been validated by the community
across a broad range of vegetation tasks, including tree-species
mapping, crop-type classification, and biomass and canopy-height
regression~\citep{feng6142416applications,ball2026geospatial,lisaius2026embedding}.
This body of work establishes that the \barlowtwins{}-trained
\tessera{} family carries strong predictive signal for vegetation, so a
held-out suite drawn primarily from that domain would largely
re-confirm what is already known. Urban and socioeconomic settings, by
contrast, remain comparatively under-evaluated. We therefore
oversampled them when assembling the held-out suite, which is why
\textsc{CityRep}~\citep{liu2026cityrep} supplies eight of the
$\nheldout$ tasks.

This deliberate imbalance concentrates weight on a single benchmark.
Several of the \textsc{CityRep} targets, among them nighttime lights,
GDP, population, and land-surface temperature, are spatially
correlated, so a uniform mean over the fourteen tasks assigns one
benchmark family more than half of the effective weight. It is
therefore fair to ask whether the held-out headline depends on that
family. We test the ranking in three ways: by removing \textsc{CityRep}
entirely, by replacing the scoring convention with two alternatives,
and by rescoring the \textsc{CityRep} tasks with the original
benchmark's point linear probe in place of the CNN head.
\tesseravtwo{}-2B-L finishes first in every case
(\Cref{tab:heldout-robust}).

\begin{table}[!htbp]
  \centering
  \caption{\textbf{The held-out ranking survives every stress test.}
  Top block: the chance-adjusted composite (the headline scheme) over
  different task subsets, and over all $\nheldout$ tasks with the
  \textsc{CityRep} scores taken from the original point linear probe
  rather than the CNN head. Bottom block: all $\nheldout$ tasks under
  two alternative aggregations. \tesseravtwo{}-2B-L is first in every
  row (bold).}
  \label{tab:heldout-robust}
  \small
  \setlength{\tabcolsep}{6pt}
  \begin{tabular}{@{}lcccc@{}}
    \toprule
    & \tesseravtwo-L & OlmoEarth-L & \tesseravone{} & \alphaearth{} \\
    \midrule
    \multicolumn{5}{@{}l}{\emph{Chance-adjusted composite (headline scheme)}} \\
    \quad All $\nheldout$ tasks            & $\mathbf{0.647}$ & $0.614$ & $0.614$ & $0.590$ \\
    \quad \textsc{CityRep} only ($8$)      & $\mathbf{0.649}$ & $0.644$ & $0.630$ & $0.604$ \\
    \quad Excluding \textsc{CityRep} ($6$) & $\mathbf{0.644}$ & $0.573$ & $0.592$ & $0.570$ \\
    \quad All $\nheldout$, \textsc{CityRep} linear probe & $\mathbf{0.597}$ & $0.573$ & $0.560$ & $0.557$ \\
    \addlinespace[2pt]
    \multicolumn{5}{@{}l}{\emph{All $\nheldout$ tasks, alternative aggregations}} \\
    \quad Plain mean of raw metrics        & $\mathbf{0.685}$ & $0.658$ & $0.658$ & $0.631$ \\
    \quad Per-task min-max, then mean      & $\mathbf{0.790}$ & $0.596$ & $0.433$ & $0.318$ \\
    \bottomrule
  \end{tabular}
\end{table}

Removing \textsc{CityRep} does not change who leads. On the six
vegetation tasks alone, \tesseravtwo{}-2B-L scores $0.644$, and its
margin over the strongest baseline grows rather than shrinks, from
$+0.033$ across all fourteen tasks to $+0.052$ once the urban family is
gone. On the eight \textsc{CityRep} tasks by themselves it is first
again, now only narrowly ahead of OlmoEarth-L. The one thing that moves
between the two slices is the runner-up: \tesseravone{} trails closest
on vegetation, OlmoEarth-L on the urban tasks.

Scoring \textsc{CityRep} by the original linear probe is the most
adversarial of the three tests for \tesseravtwo{}, whose urban edge
comes largely from a small CNN over the $64\times64$ patch around each
point (\Cref{app:cityrep-linear}). Replacing that CNN with the
benchmark's point probe on all eight \textsc{CityRep} tasks, and
folding the result back into the fourteen-task composite, OlmoEarth-L
does move ahead on the \textsc{CityRep} subset ($0.574$ versus $0.562$),
in line with its stronger point-feature readout. Over the full
fourteen tasks, however, \tesseravtwo{} still leads ($0.597$ versus
$0.573$), because its margin on the six vegetation tasks ($0.644$
versus $0.573$) outweighs the urban gap. Even under the evaluation head
that most favours the baselines, the held-out headline does not change
hands.

The averaging rule matters just as little. The headline maps each
metric to $[0,1]$ with a chance-adjustment and averages with uniform
weight, but that is one reasonable recipe among several. Averaging the
raw native metrics with no chance-adjustment keeps \tesseravtwo{}-2B-L
first ($0.685$), and so does a per-task min-max normalisation across
the four systems ($0.790$). Min-max rewards being best on each task
rather than best on average, and it widens the gap here because
\tesseravtwo{}-2B-L is top or near-top on most individual tasks.
Neither the \textsc{CityRep} family nor the specific averaging rule
carries the held-out result.

\section{Evaluation Protocol and Data Preparation}
\label{app:eval-protocol}
\paragraph{Baselines.} \Cref{sec:pixel:bench} names only the systems
that carry the headline comparison. The full \aebench{} comparison set
comprises the two directly comparable embedding-as-data systems,
\alphaearth~\citep{brown2025alphaearth} and
\tesseravone~\citep{feng2026tessera}, together with
\presto~\citep{tseng2024presto},
\textsc{OlmoEarth}~\citep{herzog2026olmoearth},
\textsc{MOSAIKS}~\citep{rolf2021mosaiks}, and a suite of remote-sensing
foundation models and generic
backbones~\citep{fuller2023croma,guo2024skysense,reed2023scalemae,%
tseng2025galileo,astruc2025anysat,dosovitskiy2021vit,ronneberger2015unet,%
klemmer2023satclip,bastani2023satlas,liu2024remoteclip,xiong2024dofa,%
jakubik2023foundation,hong2024spectralgpt,wu2025skysenseplusplus,%
clay2024,forgaard2026thor,jakubik2025terramind}. The four strongest on
this suite (\tesseravtwo{}, \tesseravone{}, \alphaearth{}, and
\textsc{OlmoEarth}) carry forward to the held-out suite.

Every system is scored on the shared \aebench{} (\Cref{app:datasets})
and on the held-out suite (\Cref{app:heldout}) through a single
procedure. Where \alphaearth~\citep{brown2025alphaearth} already reports
a baseline on the shared suite we adopt its published number, since
re-running the model under an identical setup would only reproduce it;
the remaining systems we run ourselves, staying as close to the
\alphaearth{} evaluation as their released weights allow so that every
figure is obtained the same way.

The shared-suite annotations are sparse geographic points rather than
dense masks, which leaves open how a spatially structured model should
turn a point into a feature vector. For our own students and the other
embedding-as-data systems the question does not arise: they emit a
per-pixel $10$\,m embedding, so we read the embedding at the labelled pixel directly. It
does arise for backbones such as ViT~\citep{dosovitskiy2021vit}, Prithvi, and Clay~\citep{clay2024}, which return only a coarse grid of tokens, and following
\alphaearth{} we resolve it by reading a single feature at each label
instead of propagating the label across the chip or training a dense
decoder on the backbone. Writing the encoder output as a feature grid
$F\in\R^{H_t\times W_t\times D}$ coarser than the $H\times W$ input, we
regard each token as a georeferenced cell and place the label, at its
exact coordinate, at a fractional grid position $(u,v)$. The point
feature is the bilinear interpolation of the four enclosing tokens
$\mathcal{N}(u,v)$,
\begin{equation}
  e(p) = \sum_{(a,b)\in\mathcal{N}(u,v)} w_{ab}\,F_{a,b},
  \qquad
  w_{ab} = \bigl(1-|u-a|\bigr)\bigl(1-|v-b|\bigr),
  \label{eq:bilinear-point}
\end{equation}
with $\sum_{(a,b)} w_{ab}=1$. This coincides with resampling every
channel of $F$ to $10$\,m and sampling at the label, but reads the
fractional coordinate directly, so it needs neither an upsampling decoder
nor any rounding of the label to the nearest token. When the encoder
returns one grid per acquisition we concatenate the per-date features
into $e(p)\in\R^{TD}$, and each labelled point $i$ enters the evaluation
as a single pair $(e_i,\ell_i)$, scored, following \alphaearth{}, by $1$- and $3$-nearest-neighbour classifiers and a linear probe, of which we report the best.

Reducing a patch to one point in this way is deliberate rather than a
convenience. Because each chip is cropped so that its target lies at the
centre, a decoder trained to label the full patch would receive gradient
at that single position and converge to the same head the centre probe
fits in closed form, so extracting the interpolated centre feature is the
identical experiment without the decoder. What it isolates is the quality
of a model's pointwise representation, which is what the shared suite is
meant to probe, rather than the capacity of a segmentation head or the
artefact of copying one label across an entire patch.

How these scores should be read depends on how coarse the underlying grid
is, and Prithvi~1.0~\citep{jakubik2023foundation} makes the dependence
concrete. It ingests $224\times224$ HLS~L30 chips at $30$\,m, a window of
about $6.72$\,km, over up to three acquisitions; with a patch size of
$16$ it emits a $14\times14$ grid of $768$-dimensional tokens, so one
token summarises roughly $480$\,m of ground and the concatenated point
feature has dimension $768\,T\in\{768,1536,2304\}$. Interpolating this
field to $10$\,m yields a visually smooth raster but no additional
spatial information, and the effective support of the extracted feature
stays at the ${\approx}480$\,m token. We keep this in mind when comparing
a low-resolution backbone against natively high-resolution embeddings.

All imagery is retrieved from the Microsoft Planetary Computer. We take
\stwo{} from the L2A
collection\footnote{\url{https://planetarycomputer.microsoft.com/dataset/sentinel-2-l2a}}
and \sone{} from either the
RTC\footnote{\url{https://planetarycomputer.microsoft.com/dataset/sentinel-1-rtc}}
or the
GRD\footnote{\url{https://planetarycomputer.microsoft.com/dataset/sentinel-1-grd}}
product, according to which one the model in question was trained on. Each
modality is normalised with the statistics published for that model, and
its \stwo{} cloud and haze handling is reproduced wherever it is
documented, but we apply no further radiometric processing: corrections
that some pipelines perform, such as the $\sigma^{0}$ conversion and
thermal-noise removal that THOR~\citep{forgaard2026thor} applies to \sone{} GRD, are omitted so
that a score reflects the released model on standard inputs rather than
our own preparation.

The held-out suite follows the same retrieval, normalisation, and
readout. The only labels that need separate handling are those of
\textsc{CityRep}~\citep{liu2026cityrep}, which attach to individual city
locations rather than to image tiles; for each we crop a $64\times64$
patch centred on the labelled point. The linear and nearest-neighbour
probes read its centre feature as above, while the small CNN head
takes the whole patch and can use the surrounding spatial context;
\Cref{app:cityrep-linear} contrasts the two heads on \textsc{CityRep}.

\section{The Evaluation Head Matters: \texorpdfstring{\textsc{CityRep}}{CityRep} under a Linear Probe}
\label{app:cityrep-linear}
The held-out \textsc{CityRep} numbers in the main text use a small CNN
over each label's $64\times64$ point-centred patch, which can exploit
local spatial structure. The original \textsc{CityRep}
benchmark~\citep{liu2026cityrep} instead reads each representation with
a linear probe on the point feature alone. To compare like for like
against those published numbers, we rerun the same linear probe on all
four systems: \tesseravone{} and \alphaearth{} are the values reported
by \citet{liu2026cityrep}, while \tesseravtwo{}-2B-L and OlmoEarth-L
are evaluated here (\Cref{tab:cityrep-linear}).

Under the linear probe, OlmoEarth-L and \tesseravtwo{}-2B-L are the
two strongest systems and sit close together, with OlmoEarth-L just
ahead (mean rank $1.50$ versus $1.75$) and both well clear of
\alphaearth{} and \tesseravone{} ($3.25$ and $3.50$). Their strengths
split by task type: OlmoEarth-L leads on the socioeconomic and
land-use layers (population, GDP, nighttime lights,
PM\textsubscript{2.5}, land use, and demographics), whereas
\tesseravtwo{} is best on the more physically structured targets, road
density and surface temperature, and never places below second;
OlmoEarth-L is the weakest of the four on road density. Switching to a
small CNN head tips this near-tie the other way: on the same
\textsc{CityRep} tasks \tesseravtwo{} takes the best composite
(\Cref{app:heldout-robust}), edging past OlmoEarth-L. The
lever is the head. A linear probe reads only the centre feature and
discards the surrounding $64\times64$ patch, whereas a small CNN over
that patch can use the local spatial structure that
\tesseravtwo{}'s pixel-wise embeddings carry. A dataset's headline
number therefore reflects the evaluation head as much as the
embedding, and pixel-wise representations like \tesseravtwo{} are best
read with a head that can use their spatial neighbourhood.

\begin{table}[!htbp]
  \centering
  \caption{\textbf{Linear-probe results on \textsc{CityRep}.}
  Following the original benchmark~\citep{liu2026cityrep}, each
  representation is read by a linear probe on the point feature, with
  no spatial head. Per task we report the mean primary metric across
  the $8$ cities and $5$ seeds, with the cross-city standard deviation
  as a subscript; the best per task is in bold. \tesseravone{} and
  \alphaearth{} are the published numbers~\citep{liu2026cityrep};
  \tesseravtwo{}-2B-L and OlmoEarth-L are evaluated here. Metrics are
  macro-F1 for LUC, symmetric KL for AGE (lower is better), and $R^2$
  otherwise; the final column is the mean task-level rank over the four
  systems (lower is better). Contrast the CNN-head ordering of
  \Cref{app:heldout-robust}.}
  \label{tab:cityrep-linear}
  \footnotesize
  \setlength{\tabcolsep}{3pt}
  \resizebox{\textwidth}{!}{%
  \begin{tabular}{@{}lccccccccc@{}}
  \toprule
  Model & LUC & RDE & POP & AGE & GDP & NTL & PM$_{2.5}$ & LST & Rank \\
        & F1$\uparrow$ & $R^2\!\uparrow$ & $R^2\!\uparrow$ & KL$\downarrow$ & $R^2\!\uparrow$ & $R^2\!\uparrow$ & $R^2\!\uparrow$ & $R^2\!\uparrow$ & $\downarrow$ \\
  \midrule
  \tesseravtwo{}-2B-L & $0.264_{\,0.059}$ & $\mathbf{0.551}_{\,0.074}$ & $0.624_{\,0.126}$ & $0.040_{\,0.017}$ & $0.611_{\,0.121}$ & $0.641_{\,0.129}$ & $0.457_{\,0.145}$ & $\mathbf{0.456}_{\,0.148}$ & $1.75$ \\
  \tesseravone{}      & $0.258_{\,0.067}$ & $0.538_{\,0.086}$ & $0.592_{\,0.140}$ & $0.043_{\,0.020}$ & $0.596_{\,0.134}$ & $0.595_{\,0.143}$ & $0.445_{\,0.159}$ & $0.373_{\,0.161}$ & $3.50$ \\
  \alphaearth{}       & $0.254_{\,0.076}$ & $0.514_{\,0.087}$ & $0.615_{\,0.158}$ & $0.042_{\,0.019}$ & $0.598_{\,0.127}$ & $0.632_{\,0.105}$ & $0.436_{\,0.149}$ & $0.442_{\,0.139}$ & $3.25$ \\
  OlmoEarth-L         & $\mathbf{0.268}_{\,0.091}$ & $0.503_{\,0.104}$ & $\mathbf{0.658}_{\,0.174}$ & $\mathbf{0.038}_{\,0.026}$ & $\mathbf{0.662}_{\,0.152}$ & $\mathbf{0.667}_{\,0.148}$ & $\mathbf{0.488}_{\,0.183}$ & $0.447_{\,0.172}$ & $1.50$ \\
  \bottomrule
  \end{tabular}}
\end{table}

\section{Full Pixel-wise Results}
\label{app:full-pixel-results}
\Cref{tab:full-pixel-matryoshka} gives the per-dataset
\textsc{first}-$d$ scores of the distilled pixel-wise student
behind \Cref{fig:matryoshka-empirical}, at the maximum shot per
dataset, for $d\!\in\!\dimset$, alongside the naive \matryoshka-\barlowtwins{} baseline. This student is a preliminary distillation checkpoint used only for the mechanism study of \Cref{sec:pixel:why-not-bt}, not one of the deployed N/S/M/L models; its absolute composite ($0.517$ at $d{=}128$) therefore sits well below the headline \tesseravtwo-2B-L ($0.593$), and only the distilled-versus-naive gap is meaningful here. Distilled \matryoshka{}
dominates the naive baseline at every reported prefix dimension;
the gap is largest at $d{=}16$ and narrows as $d$ grows, consistent
with the interpretation (\Cref{sec:pixel:why-not-bt}) that the
naive method receives stronger gradient pressure on early
coordinates but does not learn a robust semantic ordering.

\begin{table}[!htbp]
  \centering
  \caption{\textbf{Per-dataset \matryoshka{} prefix scores}
  for the pixel-wise distillation student of
  \Cref{fig:matryoshka-empirical} (\textsc{first}-$d$, maximum
  shot). Top: distilled \matryoshka; bottom: naive
  \matryoshka-\barlowtwins. Per-dataset entries are raw balanced
  accuracy (classification / change detection) and $R^{2}$
  (regression); the \textbf{Composite} row is the chance-adjusted
  downstream score of \Cref{sec:scaling:design}.}
  \label{tab:full-pixel-matryoshka}
  \scriptsize
  \setlength{\tabcolsep}{4pt}
  \begin{tabular}{lcccc@{\hskip 1.2em}cccc}
  \toprule
  & \multicolumn{4}{c}{Distilled \matryoshka{}}
  & \multicolumn{4}{c}{Naive \matryoshka-\barlowtwins{}} \\
  \cmidrule(lr){2-5}\cmidrule(lr){6-9}
  Dataset & $16$ & $32$ & $64$ & $128$ & $16$ & $32$ & $64$ & $128$ \\
  \midrule
  Africa crop mask & $0.840$ & $0.867$ & $0.852$ & $0.854$ & $0.800$ & $0.835$ & $0.838$ & $0.849$ \\
  ASTER GED        & $0.770$ & $0.788$ & $0.805$ & $0.819$ & $0.735$ & $0.770$ & $0.793$ & $0.810$ \\
  CA crops (coarse)& $0.460$ & $0.521$ & $0.547$ & $0.555$ & $0.447$ & $0.504$ & $0.530$ & $0.545$ \\
  CA crops (fine)  & $0.407$ & $0.465$ & $0.489$ & $0.509$ & $0.386$ & $0.443$ & $0.476$ & $0.499$ \\
  Descals oil palm & $0.626$ & $0.703$ & $0.690$ & $0.690$ & $0.604$ & $0.680$ & $0.685$ & $0.681$ \\
  Ethiopia crops   & $0.349$ & $0.354$ & $0.376$ & $0.378$ & $0.348$ & $0.348$ & $0.376$ & $0.370$ \\
  GLanCE           & $0.616$ & $0.625$ & $0.645$ & $0.664$ & $0.562$ & $0.603$ & $0.632$ & $0.642$ \\
  LCMAP-LC         & $0.749$ & $0.773$ & $0.793$ & $0.800$ & $0.714$ & $0.745$ & $0.778$ & $0.791$ \\
  LCMAP-$\Delta$LC & $0.740$ & $0.762$ & $0.756$ & $0.747$ & $0.716$ & $0.738$ & $0.736$ & $0.745$ \\
  LCMAP-LU         & $0.746$ & $0.771$ & $0.798$ & $0.804$ & $0.699$ & $0.747$ & $0.781$ & $0.791$ \\
  LCMAP-$\Delta$LU & $0.757$ & $0.748$ & $0.742$ & $0.731$ & $0.708$ & $0.722$ & $0.731$ & $0.722$ \\
  LUCAS-LC         & $0.306$ & $0.345$ & $0.384$ & $0.407$ & $0.285$ & $0.319$ & $0.367$ & $0.383$ \\
  LUCAS-LU         & $0.318$ & $0.340$ & $0.361$ & $0.371$ & $0.293$ & $0.318$ & $0.345$ & $0.350$ \\
  OpenET ET        & $0.477$ & $0.564$ & $0.617$ & $0.636$ & $0.440$ & $0.532$ & $0.601$ & $0.619$ \\
  US trees         & $0.143$ & $0.157$ & $0.177$ & $0.184$ & $0.123$ & $0.140$ & $0.166$ & $0.171$ \\
  \midrule
  \textbf{Composite} & $\mathbf{0.454}$ & $\mathbf{0.493}$ &
  $\mathbf{0.510}$ & $\mathbf{0.517}$ & $\mathbf{0.413}$ &
  $\mathbf{0.463}$ & $\mathbf{0.492}$ & $\mathbf{0.503}$ \\
  \bottomrule
  \end{tabular}
\end{table}

\section{Matryoshka Failure Analysis}
\label{app:matryoshka-failure}

\paragraph{Coordinate-exchangeability seed variance.} The
\textsc{random}-16 scores of \Cref{tab:matryoshka-analysis} are
averaged over five seeds. Across the classification subset the
seed standard deviation is small---$0.001$--$0.013$ for distilled
\matryoshka{} and $0.004$--$0.016$ for naive
\matryoshka-\barlowtwins{} (\Cref{tab:seed-std})---and well below
the $0.044$ \textsc{first}-vs-\textsc{random} gap of the naive
model, so the observed ordering is not a seed-noise artefact.

\begin{table}[!htbp]
  \centering
  \caption{\textbf{\textsc{random}-16 seed standard deviation}
  across five seeds, per classification dataset.}
  \label{tab:seed-std}
  \small
  \begin{tabular}{lcc}
  \toprule
  Dataset & Distilled $\sigma$ & Naive $\sigma$ \\
  \midrule
  Africa crop mask & $0.008$ & $0.016$ \\
  ASTER GED        & $0.011$ & $0.009$ \\
  GLaNCE           & $0.008$ & $0.010$ \\
  LCMAP-LC         & $0.007$ & $0.004$ \\
  LCMAP-LU         & $0.013$ & $0.013$ \\
  Descals oil palm & $0.006$ & $0.004$ \\
  LUCAS-LC         & $0.009$ & $0.010$ \\
  US trees         & $0.001$ & $0.005$ \\
  \bottomrule
  \end{tabular}
\end{table}

\paragraph{Per-coordinate scale staircase.} The naive model
exhibits a staircase in per-coordinate standard deviation
$\sigma_d$ aligned with the prefix-loss multiplicities $m_j$ of
\Cref{eq:multiplicity}, while distilled \matryoshka{} remains
flat (\Cref{tab:sigma-staircase}). This is the quantitative form
of \Cref{fig:matryoshka-empirical}(c): naive early coordinates
($m_j\!=\!4$) are over-constrained and heavy-tailed, and the
distortion weakens monotonically as $m_j$ decreases toward the
single-prefix tail.

\begin{table}[!htbp]
  \centering
  \caption{\textbf{Per-coordinate scale by prefix-loss
  multiplicity.} Mean per-dimension standard deviation $\sigma_d$
  within each coordinate group.}
  \label{tab:sigma-staircase}
  \small
  \begin{tabular}{lcccc}
  \toprule
  Group & $m_j$ & Distilled $\sigma_d$ & Naive $\sigma_d$ &
  Naive rel.\ grad.\ norm \\
  \midrule
  $1$--$16$   & $4$ & ${\sim}1.00$ & ${\sim}2.0$ & $3.92$ \\
  $17$--$32$  & $3$ & ${\sim}1.00$ & ${\sim}1.5$ & $2.93$ \\
  $33$--$64$  & $2$ & ${\sim}1.00$ & ${\sim}1.2$ & $1.97$ \\
  $65$--$128$ & $1$ & ${\sim}1.00$ & ${\sim}1.0$ & $1.00$ \\
  \bottomrule
  \end{tabular}
\end{table}

\paragraph{Gradient norm per coordinate group.} The last column of
\Cref{tab:sigma-staircase} reports the naive model's mean gradient
norm per coordinate group, normalised to the $m_j\!=\!1$ tail. It
tracks the prefix-loss multiplicity almost exactly
($\approx\!m_j$-fold), whereas the distilled model stays close to
flat ($0.98$--$1.09$ across groups). This confirms that the naive
\textsc{first}-$d$ gains arise from unequal optimisation pressure
rather than from a learned semantic coordinate order.

\paragraph{Effective rank, CKA, and Procrustes alignment.} Student
prefixes retain higher normalised effective rank and stronger
alignment to the full teacher embedding than teacher
coordinate-prefixes at the same $k$ (\Cref{tab:spectral-diag}); the
gap is largest at $k\!=\!16$ and vanishes at $k\!=\!\dimfull$ where
both use the full coordinate set. We report \emph{normalised
effective rank}---the effective rank (the entropy of the normalised
singular-value spectrum) divided by the prefix dimension $k$---so
the metric lies in $[0,1]$ and measures the fraction of the $k$
available axes that the prefix actually exercises. The student
prefix uses $\sim\!0.9$ of its axes at every $k$, whereas the
teacher coordinate-prefix uses barely half at $k\!=\!16$.

\begin{table}[!htbp]
  \centering
  \caption{\textbf{Spectral and alignment diagnostics} for student
  prefix $\bs_{1:k}$ vs.\ teacher coordinate-prefix $\bt_{1:k}$,
  against the full teacher embedding. Normalised effective rank
  $\in[0,1]$ is the effective rank divided by $k$.}
  \label{tab:spectral-diag}
  \scriptsize
  \setlength{\tabcolsep}{4pt}
  \begin{tabular}{ccccccc}
  \toprule
  & \multicolumn{2}{c}{Norm.\ eff.\ rank} & \multicolumn{2}{c}{CKA to $\bt$}
  & \multicolumn{2}{c}{Procrustes resid.} \\
  \cmidrule(lr){2-3}\cmidrule(lr){4-5}\cmidrule(lr){6-7}
  $k$ & student & teacher & student & teacher & student & teacher \\
  \midrule
  $16$  & $0.92$ & $0.54$ & $0.742$ & $0.493$ & $0.412$ & $0.611$ \\
  $32$  & $0.91$ & $0.58$ & $0.831$ & $0.621$ & $0.323$ & $0.512$ \\
  $64$  & $0.92$ & $0.65$ & $0.912$ & $0.764$ & $0.218$ & $0.392$ \\
  $128$ & $0.87$ & $0.87$ & $0.958$ & $0.958$ & $0.124$ & $0.124$ \\
  \bottomrule
  \end{tabular}
\end{table}

\paragraph{Teacher-target dimensionality.} The gradient-multiplicity
argument concerns the within-$\dimfull$ coordinate gauge that
\barlowtwins{} does not fix, and is independent of the teacher
target dimension $\dimteacher$: changing $\dimteacher$ changes the
distillation compression ratio
($\dimteacher/d$, from $8\times$ at $d{=}128$ to $64\times$ at
$d{=}16$) but does not remove the unequal self-supervised gradient
pressure in naive prefix-BT. Empirically, a $\dimteacher\!=\!512$
target gives slightly lower but monotonically identical prefix
retention to the production $\dimteacher\!=\!1024$ target: the
$d\!=\!16/32/64/128$ composites are $0.441/0.485/0.498/0.506$ at
$\dimteacher\!=\!512$ versus $0.454/0.493/0.510/0.517$ at
$\dimteacher\!=\!1024$.


\section{Scaling study: additional material}
\label{app:scaling-extra}

\subsection{Scope of the findings}
\label{sec:scaling:scope}
\textbf{F1} and \textbf{F2} are properties of the architectural family we sweep:
pixel-wise \sonetwo{} encoders pretrained with
\barlowtwins~\citep{zbontar2021barlow} at short sequence lengths
$L\!\in\!\seqlen$. They are not advanced as universal EO scaling
laws. We do not claim, and our experiments do not establish, that:
(a)~patch-level architectures with masked-image-modelling
objectives obey the same exponents;
(b)~the same exponents extend to other satellite modalities
(e.g., hyperspectral or sub-meter optical imagery);
(c)~the projector-saturation pattern would replicate under
contrastive or generative pretraining objectives whose projector
plays a different role;
(d)~the compute-waste quantification associated with loss-based
model selection is invariant to evaluation suite---it is computed
against the $\nscalingtasks$-task composite of this study, and a
different task mixture would give a different number.

\subsection{Why the loss correlates so weakly with downstream score}
\label{app:scaling-extra:why-weak}
We attribute the weak loss--score correlation of \textbf{F1} to four
properties of EO that are not jointly present in the language and
natural-image regimes where loss-based scaling laws have been
validated:
(i)~heterogeneous downstream tasks with different sensitivities
to the representation;
(ii)~cloud and orbital sampling whose variance dominates the loss
without being downstream-relevant;
(iii)~multi-sensor signals with different physical meaning and
noise characteristics; and
(iv)~the susceptibility of redundancy-reduction objectives to
shortcut invariances orthogonal to the physical processes
downstream tasks depend on.

\subsection{Loss-optimal versus score-optimal selection}
\label{app:scaling-extra:selection}
Within each of the \nbuckets{} compute buckets of
\Cref{fig:scaling-laws}b--d we contrast two selection criteria:
the run with the highest downstream score (\emph{score-optimal})
and the run with the lowest normalised \barlowtwins{} loss
(\emph{loss-optimal}). Fitting power laws through the resulting
bucket peaks separately gives
$N_{\mathrm{enc}}^{\star}\!\propto\!C^{\encexp}$ for score-optimal
selection but a nearly flat
$N_{\mathrm{enc}}^{(\ell)}\!\propto\!C^{\encexploss}$ for
loss-optimal selection; the two laws share roughly the same
low-compute behaviour but diverge as $C$ grows. Translating the
per-bucket score gap (mean gap $0.049$ along a score-envelope
slope of $0.089$ per compute decade) into compute units yields the
$\wastemult$ figure quoted in the main text
(derivation in \Cref{app:scaling-setup}, \Cref{eq:waste}). For every run we
record the converged \barlowtwins{} loss on a held-out \dpixel{}
validation set, normalised by the projector-determined ceiling
$P + \lambda_{\mathrm{BT}}\,P(P-1)$ so that runs with different
projector widths are comparable.

\subsection{Fitted exponents}
\label{app:scaling-extra:exponents}
\Cref{tab:scaling-exponents} reports the fitted exponents with
bootstrap confidence intervals and leave-one-bucket-out (LOBO)
stability.

\begin{table}[!htbp]
  \centering
  \caption{Fitted scaling exponents ($\nruns$ runs; bootstrap
  $95\%$ CI; LOBO mean$\pm\sigma$ over \nbuckets{} leave-one-out
  fits, seven for the projector).}
  \label{tab:scaling-exponents}
  \small
  \begin{tabular}{lccc}
  \toprule
  Component & Exponent & 95\% CI & LOBO mean$\pm\sigma$ \\
  \midrule
  Encoder $N_{\mathrm{enc}}^{\star}(C)$
    & $\encexp$ & $[+0.31,+0.39]$ & $0.36\pm0.01$ \\
  Projector $N_{\mathrm{proj}}^{\star}(C)$
    & $\projexp$ & $[-0.02,+0.04]$ & $0.01\pm0.01$ \\
  Data $D^{\star}(C)$
    & $\dataexp$ & $[+0.61,+0.69]$ & $0.63\pm0.01$ \\
  \midrule
  \multicolumn{2}{l}{Loss-optimal encoder $N_{\mathrm{enc}}^{(\ell)}(C)$}
    & \multicolumn{2}{l}{$\propto C^{\encexploss}$} \\
  \bottomrule
  \end{tabular}
\end{table}

\section{Teacher objective details}
\label{app:teacher-objective}
The two views $Y_A, Y_B$ of each \dpixel{} are encoded by the
dual-branch encoder and projected to $\R^{N_{\mathrm{proj}}}$,
giving batch-normalised embeddings $Z_A, Z_B \in \R^{B\times P}$.
The \barlowtwins{} loss enforces invariance to the random temporal
sampling while decorrelating embedding dimensions:
\begin{equation}
\Lbt \;=\; \sum_{i}\bigl(1-C_{ii}\bigr)^{2}
        + \lambda_{\mathrm{BT}}\sum_{i}\sum_{j\neq i} C_{ij}^{2},
\qquad
C \;=\; \tfrac{1}{B}\,Z_{A}^{\top}Z_{B},
\label{eq:bt-loss}
\end{equation}
with $\lambda_{\mathrm{BT}}\!=\!5\!\times\!10^{-3}$ following
\citet{zbontar2021barlow}.

The mix-up regulariser enforces a linear-consistency property:
linear interpolation in the input space should correspond to linear
interpolation in the embedding's cross-correlation space. At each
step we sample $\alpha_{\mathrm{mix}}\!\sim\!\mathcal{U}(0,1)$,
construct a batch-shuffled view $Y_S\!=\!\mathrm{Shuffle}(Y_B)$,
mix $Y_M\!=\!\alpha_{\mathrm{mix}} Y_A + (1\!-\!\alpha_{\mathrm{mix}}) Y_S$,
and compare the mixed embedding's cross-correlations against the
interpolation-predicted targets:
\begin{equation}
\Lmix
\;=\;
\tfrac{1}{2}\bigl\|C^{MA}\!-\!C^{MA}_{\mathrm{target}}\bigr\|_{F}^{2}
+ \tfrac{1}{2}\bigl\|C^{MS}\!-\!C^{MS}_{\mathrm{target}}\bigr\|_{F}^{2},
\label{eq:mix-loss}
\end{equation}
where $C^{MA}\!=\!Z_M^{\top}Z_A$, $C^{MS}\!=\!Z_M^{\top}Z_S$, and
the targets are the convex combinations
$\alpha_{\mathrm{mix}} Z_A^{\top}Z_{\cdot} + (1\!-\!\alpha_{\mathrm{mix}}) Z_S^{\top}Z_{\cdot}$
predicted by linear interpolation; the full derivation is in
\Cref{app:mixup}. The total objective is
\begin{equation}
\loss_{\mathrm{total}} \;=\; \Lbt + \lambda_{\mathrm{mix}}\,\Lmix,
\qquad
\lambda_{\mathrm{mix}}\!=\!1.0,
\label{eq:total-loss}
\end{equation}
with $\lambda_{\mathrm{mix}}$ recommended by
\citet{bandara2023mixup}. We retain \tesseravone's global shuffling
of \dpixels{} across MGRS tiles to prevent spatial-autocorrelation
bias during batching; \tesseravone{} ablations show that removing
either mix-up or global shuffling degrades validation accuracy
substantially under temporal sparsity.

\section{Deployment Cost and Choosing an Artefact}
\label{app:deploy-cost}
\label{app:cost}
\begin{figure}[!htbp]
  \centering
  \includegraphics[width=\linewidth]{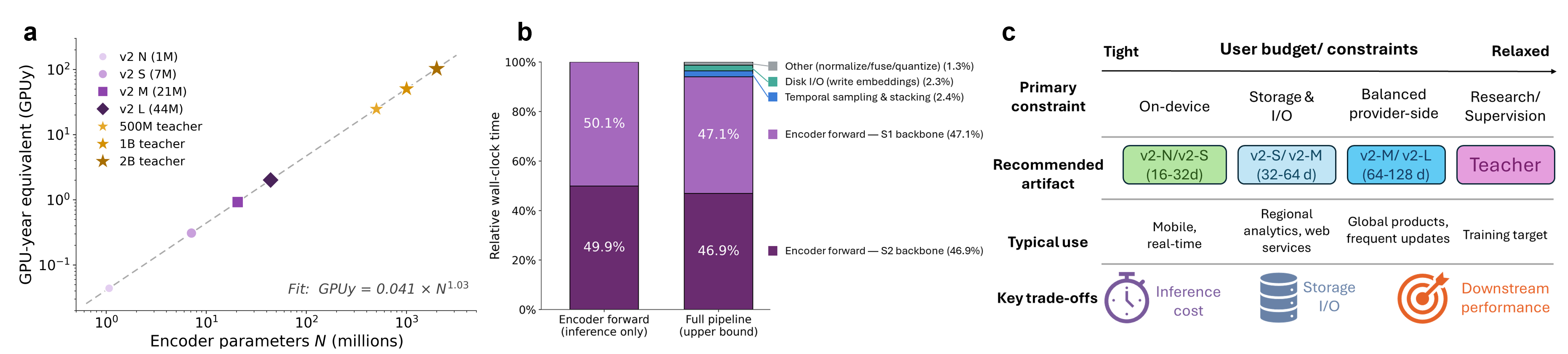}
  \caption{Deployment cost of the student family.
  \textbf{(a)}~GPUy cost of one annual global pass vs.\ encoder
  size, with the fit of \Cref{eq:gpuy}; markers are the
  \tesseravtwo{} students and the $500$M/$1$B/$2$B teachers.
  \textbf{(b)}~Wall-clock breakdown of the embedding-as-data
  pipeline: the encoder forward pass accounts for $\sim\!94\%$.
  \textbf{(c)}~Recommended artefact and prefix dimension by
  user-budget regime.}
  \label{fig:deploy-cost}
\end{figure}

\Cref{fig:deploy-cost} summarises the inference cost of one global,
annual, $10\,\mathrm{m}$ \sonetwo{} pass, expressed in
single-\gputype-year equivalents ($\mathrm{GPUy}$). Across the
encoder-size range spanned by our students and the teacher, the
cost follows a near-linear power law in the encoder parameter
count $N$ (in millions),
\begin{equation}
\mathrm{GPUy}(N)\;\approx\;0.041\cdot N^{1.03},
\label{eq:gpuy}
\end{equation}
fit to the inference-FLOPs accounting of
\Cref{app:flops:inference} and the per-FLOP throughput of the
\gputype{} accelerator. Over our students and teacher this gives
$\approx\!100~\mathrm{GPUy}$ for the $\teacherparams$ teacher and
$\approx\!0.3$, $0.9$, and $2~\mathrm{GPUy}$ for
\tesseravtwo-2B-S ($\studentS$), -M ($\studentM$), and -L
($\studentL$) respectively.

\paragraph{Encoder inference dominates wall-clock cost.}
\Cref{eq:gpuy} counts encoder forward-pass FLOPs, which is the
quantity that matters: in the full embedding-as-data pipeline
(\Cref{fig:deploy-cost}b), the encoder forward pass, split almost
evenly between the \stwo{} and \sone{} backbones ($\sim\!47\%$
each), accounts for $\sim\!94\%$ of wall-clock time, while data
preparation adds only $\sim\!6\%$ (temporal sampling and stacking
$2.4\%$, disk I/O for writing embeddings $2.3\%$, and
normalisation/fusion/quantisation $1.3\%$). The full pipeline is
therefore only $\sim\!1.06\times$ the encoder-only cost, so the
GPUy estimates of \Cref{fig:deploy-cost}(a) are a tight proxy for
end-to-end cost.

\paragraph{Choosing an artefact in practice.} Different users have
different binding constraints; \Cref{fig:deploy-cost}(c)
summarises which artefact and prefix dimension we recommend for
each regime. On-device and edge deployments (field apps, mobile,
real-time) are best served by \tesseravtwo-2B-S at
$d\!\in\!\{16,32\}$; storage-constrained regional pipelines by
\tesseravtwo-2B-S/M at $d\!\in\!\{32,64\}$; balanced provider-side
products by \tesseravtwo-2B-M/L at $d\!\in\!\{64,128\}$; and
research or further-supervision use cases by the $\teacherparams$
teacher at $d\!=\!128$.

\section{Why naive \matryoshka{} self-supervision fails}
\label{sec:pixel:why-not-bt}
A natural baseline is to add \matryoshka{} prefixes during
self-supervised pretraining: have a single encoder produce
$\bz\in\R^{\dimfull}$, attach a separate projector $g_k$ to each
prefix $\bz_{1:k}$, and minimise
$\sum_{k\in\mathcal{K}}\Lbt\!\bigl(g_k(\bz^{A}_{1:k}),
g_k(\bz^{B}_{1:k})\bigr)$. We tried this and it does not work
(\Cref{sec:analysis:matryoshka-fails}). The failure has a clean
mathematical explanation that, we argue, generalises beyond
\barlowtwins{} and beyond EO.

\paragraph{Rotation symmetry.} \barlowtwins{} identifies
representation \emph{subspaces} only up to orthogonal
transformations: if $Z\!\in\!\R^{B\times \dimfull}$ is a feasible
(decorrelated, invariant) solution then so is $ZQ$ for any
orthogonal $Q\!\in\!\R^{\dimfull\times \dimfull}$. Coordinate
ordering within the embedding is therefore a gauge degree of
freedom; nothing in the loss specifies which axes should carry
which information.

\paragraph{Prefix-BT breaks the symmetry through gradient
multiplicity, not through a semantic signal.} Let
$\mathcal{K}\!=\!\dimset$ be the prefix set, and define the
\emph{gradient multiplicity} of coordinate $j$ as the number of
prefix losses that act on it,
\begin{equation}
m_{j} \;=\; \bigl|\bigl\{k \in \mathcal{K} : j \le k\bigr\}\bigr|
        \;=\; \begin{cases}
            4 & j \in [1,16],\\
            3 & j \in [17,32],\\
            2 & j \in [33,64],\\
            1 & j \in [65,128].
        \end{cases}
\label{eq:multiplicity}
\end{equation}
The total gradient feeding coordinate $j$ is the sum of $m_{j}$
contributions from prefix losses $\{\Lbt^{(k)}\}_{k\ge j}$.
Adding these losses does break the rotation symmetry of single-prefix
\barlowtwins, but it breaks it through gradient pressure: nothing in
the construction specifies that the most informative axes should be
the early ones, only that early ones receive more gradient.

\begin{figure}[!htbp]
  \centering
  \includegraphics[width=\linewidth]{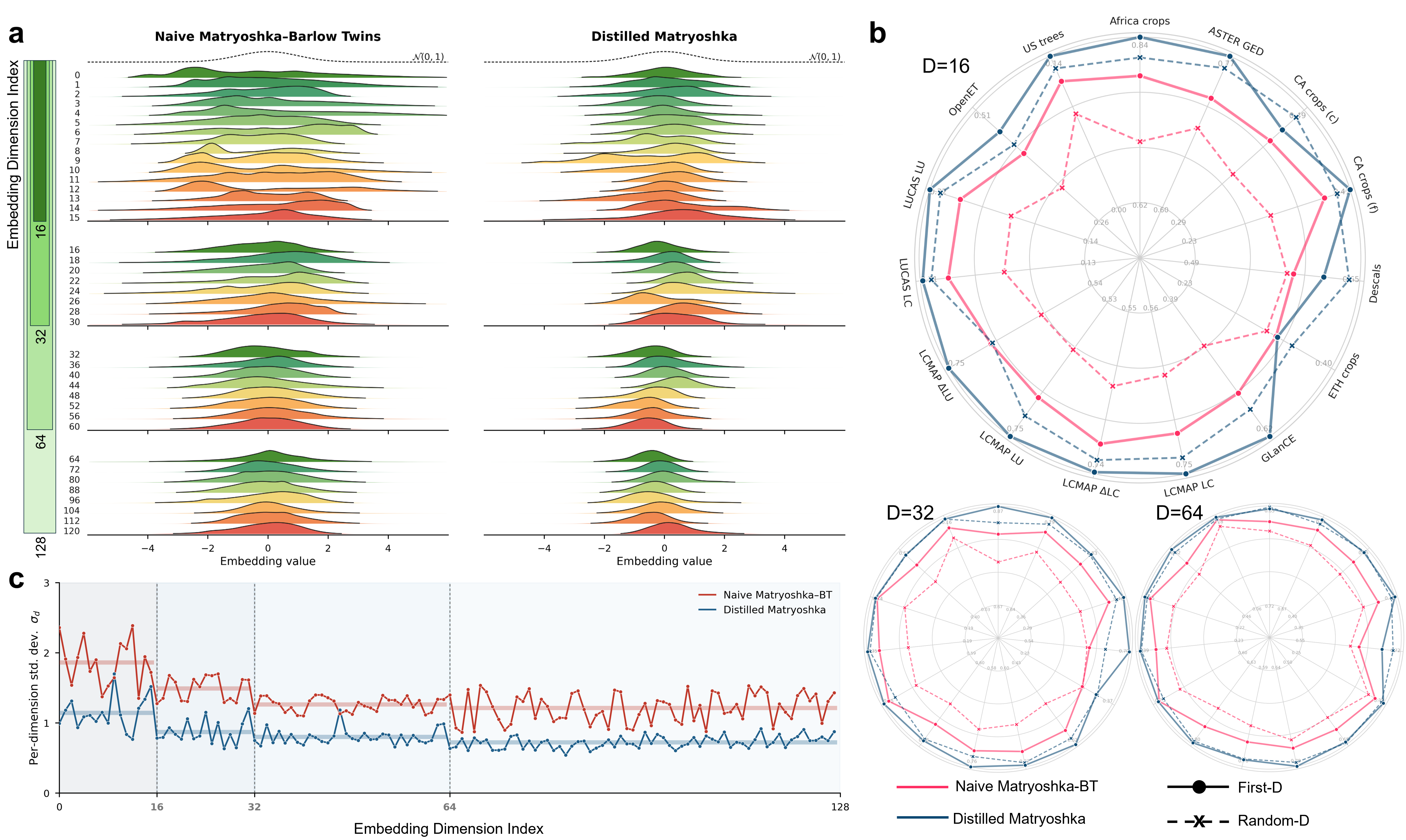}
  \caption{Why naive \matryoshka-\barlowtwins{} fails.
  \textbf{(a)}~Per-dimension marginals for the naive method (left)
  and distillation (right), grouped by prefix-loss multiplicity
  $m_j$; reference $\mathcal{N}(0,1)$.
  \textbf{(c)}~Per-dimension standard deviation $\sigma_d$; dashed
  lines mark the prefix boundaries $d\!=\!16, 32, 64$.
  \textbf{(b)}~Downstream radar at prefix sizes
  $D\!\in\!\{16,32,64\}$: \textcolor{red!60}{naive} vs.\
  \textcolor{blue!70}{distilled}, \textsc{first}-$D$ vs.\
  \textsc{random}-$D$ (five seeds).}
  \label{fig:matryoshka-empirical}
\end{figure}

\paragraph{An implicit Gaussian baseline that prefix-BT disturbs.}
The visible footprint of this gradient imbalance is best understood
against a baseline that the loss itself does not impose.
\barlowtwins{} constrains only the cross-correlation matrix of the
projector outputs to be diagonal, and imposes no marginal-distribution
constraint. Empirically, however, single-prefix \barlowtwins{}
reliably produces approximately Gaussian per-coordinate marginals.
Three mechanisms cooperate to produce this implicit bias: (i)~batch
normalisation on the projector output fixes the first two moments;
(ii)~each projector coordinate is a high-dimensional linear
combination of upstream activations, which yields CLT-style mixing;
and (iii)~SGD with weight decay drifts toward maximum-entropy
solutions on the resulting flat manifold, and a Gaussian is the
maximum-entropy distribution at fixed mean and variance. Approximate
Gaussianity of standard \barlowtwins{} marginals is therefore a
\emph{default state} of the optimisation, not a target of the loss.
\Cref{fig:matryoshka-empirical}(a), right column, shows distilled
\matryoshka{} sitting on this default state across all four
prefix-multiplicity groups; the left column shows the naive method
departing from it on the heavily-loaded early dimensions.

\paragraph{Two footprints from one mechanism.} The
$m_{j}$-loaded coordinates suffer in two coupled ways.
\textbf{(i)~Variance inflation.} Whether the $m_{j}$ prefix
gradients are independent (so that stationary variance scales
linearly in $m_{j}$), aligned (quadratically), or partially
correlated (intermediate) depends on inter-loss gradient
correlation, which is loss-design-dependent and not predictable a
priori. We therefore make only the safe ordering claim:
\emph{stationary variance $\sigma_{d}^{2}$ at coordinate $j$ is
monotone non-decreasing in $m_{j}$.} This predicts the strict
ordering $\sigma_{[1,16]}\!>\!\sigma_{[17,32]}\!>\!\sigma_{[33,64]}\!>\!\sigma_{[65,128]}$
without committing to a specific exponent.
\textbf{(ii)~Marginal-shape distortion.} The $m_{j}$ prefix losses
are not jointly satisfiable: whitening of $Z_{1:16}^{\top}Z_{1:16}/B$
is a different constraint from whitening of
$Z_{1:32}^{\top}Z_{1:32}/B$, projected onto the same shared early
coordinates. Their stationary points generically conflict, and SGD
oscillates on a noisy plateau where no single constraint is exactly
satisfied. On this plateau, the implicit-Gaussian default no longer
applies, and marginal distributions are free to take ad-hoc shapes
(bimodal, heavy-tailed, skewed). The $\sigma_{d}$ step pattern in
\Cref{fig:matryoshka-empirical}(c) and the non-Gaussian marginals on
dims $0$--$15$ in panel~(a) are therefore not two phenomena: they
are the same gradient-multiplicity pathology observed through two
different statistics.

\begin{proposition}[Naive prefix-BT is unsupervised gauge-fixing
by gradient counting; distilled \matryoshka{} is supervised nested
rate--distortion.]
\label{prop:matryoshka}
Under prefix-\barlowtwins{} with prefix set $\mathcal{K}$ and
multiplicity $m_{j}$ defined by \cref{eq:multiplicity}, the
optimisation is invariant under any orthogonal transformation that
acts within each multiplicity-equivalence class
$\{j : m_{j}\!=\!m\}$ but not across them. The remaining gauge
freedom is therefore between equivalence classes, and is fixed at
convergence not by any external semantic signal but by the
optimisation dynamics: stationary variance $\sigma_{d}^{2}$ and
marginal-shape distortion at coordinate $j$ are both non-decreasing
in $m_{j}$, and the implicit Gaussian regularity of single-prefix
\barlowtwins{} is preserved on the $m_{j}\!=\!1$ tail and
progressively lost as $m_{j}$ grows. In contrast, the
\matryoshka{} distillation objective
$\sum_{k}\alpha_{k}\,d(h_{k}(s_{1:k}), t)$ supervises every prefix
against a single fixed target $\bt$, fixing the gauge through
$\bt$, eliminating cross-prefix gradient conflict, and inheriting
the implicit Gaussian regularity from the teacher.
\end{proposition}

\paragraph{Both predictions hold empirically.} \Cref{fig:matryoshka-empirical}
makes the test concrete. The $\sigma_{d}$ ordering of
\Cref{prop:matryoshka} is observed in panel~c: naive
$\sigma_{d}\!\in\![1.5,2.5]$ on dims $0$--$15$ ($m_{j}\!=\!4$),
steps down at $d\!=\!16$ on entering the $m_{j}\!=\!3$ group,
again at $d\!=\!32$ ($m_{j}\!=\!2$), and recovers to
$\sigma_{d}\!\approx\!1$ by $d\!=\!64$ ($m_{j}\!=\!1$). The step
locations match the prefix boundaries to within a single
coordinate. The marginal-shape distortion is observed in panel~a:
naive dims $0$--$15$ are heavy-tailed and bimodal, and the
distortion weakens monotonically with $m_{j}$. Distilled
\matryoshka{} sits at $\sigma_{d}\!\approx\!1$ and on Gaussian-like
marginals throughout all $128$ coordinates.

\paragraph{Downstream consequences.}
Both footprints translate directly into the radar plots in
\Cref{fig:matryoshka-empirical}(b). On $14$ tasks from the
\aebench, the four configurations sit in the predicted order at
every prefix size:
\textsc{distilled-first}\,$>$\,\textsc{distilled-random}\,$>$\,%
\textsc{naive-first}\,$>$\,\textsc{naive-random}. The
\textsc{first}-vs-\textsc{random} gap diagnoses whether a method
has acquired a \emph{real} coordinate ordering: it is small for
distilled \matryoshka{} (because random subsets of a faithfully
compressed code remain informative) and large for naive
\matryoshka-\barlowtwins{} (because the apparent advantage of
\textsc{first}-$D$ over \textsc{random}-$D$ comes from the
$m_{j}\!=\!4$ over-loading of dims $0$--$15$ rather than from
genuinely ordered semantic content). The gap is most pronounced
at $D\!=\!16$, where the gradient-multiplicity asymmetry is
extreme, and narrows at $D\!=\!32, 64$ as more tail dimensions
enter the prefix and the $m_{j}$ gap closes.

\paragraph{Composite scores behind \Cref{fig:matryoshka-empirical}(b).}
\Cref{tab:matryoshka-cls-composite} reports, averaged over the seven
balanced-accuracy classification tasks plotted in
\Cref{fig:matryoshka-empirical}(b) (Africa crop mask, GLanCE, LCMAP-LC,
LCMAP-LU, Descals oil palm, LUCAS-LC, US trees), the mean prefix scores,
chance-adjusted as in \Cref{sec:scaling:design}.

\begin{table}[!htbp]
  \centering
  \caption{Composite classification score per \matryoshka{} prefix
  behind \Cref{fig:matryoshka-empirical}(b), for distilled
  \matryoshka{} and the naive \matryoshka-\barlowtwins{} baseline, under
  \textsc{first}-$d$ and \textsc{random}-$d$ prefixes.}
  \label{tab:matryoshka-cls-composite}
  \small
  \begin{tabular}{lcccccc}
\toprule
& \multicolumn{2}{c}{$d{=}16$} & \multicolumn{2}{c}{$d{=}32$}
& \multicolumn{2}{c}{$d{=}64$} \\
\cmidrule(lr){2-3}\cmidrule(lr){4-5}\cmidrule(lr){6-7}
Method & \textsc{first} & \textsc{rand} & \textsc{first} &
\textsc{rand} & \textsc{first} & \textsc{rand} \\
\midrule
Distilled \matryoshka{}        & $0.486$ & $0.472$ & $0.529$ &
$0.511$ & $0.543$ & $0.545$ \\
Naive \matryoshka-\barlowtwins{} & $0.441$ & $0.382$ & $0.492$ &
$0.453$ & $0.519$ & $0.501$ \\
\bottomrule
  \end{tabular}
\end{table}
Distilled \textsc{first} and \textsc{random} differ by at most
$0.018$ at every prefix dimension and become indistinguishable by
$d\!=\!64$. Naive \textsc{first} exceeds naive \textsc{random} by
$0.059$ at $d\!=\!16$, $0.039$ at $d\!=\!32$, and $0.018$ at
$d\!=\!64$; the gap narrows as the $m_{j}\!=\!4$ early-coordinate
cluster contributes a smaller fraction of the prefix, exactly as
\Cref{prop:matryoshka} predicts. The distilled-vs-naive gap at
\textsc{first}-$d$ also contracts with dimension ($0.045$, $0.037$,
$0.024$ for $d\!=\!16,32,64$), but the \textsc{first}-vs-%
\textsc{random} gap within the naive model is the more diagnostic
quantity, since it isolates whether the apparent \textsc{first}-$d$
advantage reflects ordered semantic content or merely gradient
multiplicity.

\paragraph{Distillation supplies the missing signal on two
fronts.} Distillation against a fixed teacher target repairs both
footprints separated by \Cref{prop:matryoshka}.
\textbf{(i)~An external semantic ordering.} Each prefix $\bs_{1:k}$
is supervised to reconstruct the full teacher embedding $\bt$, not
the teacher's own first $k$ coordinates; the prefix becomes a
$k$-dimensional rate--distortion bottleneck on a fixed target
distribution, and later dimensions encode the residual. The gauge
that prefix-BT fixed by gradient counting is now fixed by the
teacher.
\textbf{(ii)~Inherited marginal regularity.} The
$\teacherparams$ teacher was trained with single-prefix
\barlowtwins{} and therefore inherits its implicit Gaussian
marginals. Under MSE-style reconstruction, the optimal compression
of an approximately Gaussian target produces approximately Gaussian
projections. Distilled prefix marginals look Gaussian because they
inherit the teacher's distribution, not because we asked them to.
The same pattern applies whenever a self-supervised objective
leaves a coordinate gauge and a nested-representation requirement
is imposed afterwards: the gauge must be fixed by an external
target distribution. \emph{Self-supervision learns a representation
subspace; distillation learns an ordered code.}

\subsection{Distillation turns prefix learning into ordered
compression}
\label{sec:pixel:ordered-compression}
A direct consequence is that the student's first $16$ dimensions
need not match, and typically should not match, the teacher's
first $16$ coordinates. The teacher's coordinates are arbitrary;
the student's first $16$ are a \emph{learned compressed code} for
the teacher's full $\dimteacher$-dimensional embedding. We confirm
this empirically: distilled \matryoshka{} prefixes
$\bs_{1:16},\bs_{1:32},\bs_{1:64}$ outperform the corresponding
teacher prefixes $\bt_{1:16},\bt_{1:32},\bt_{1:64}$, despite the
student being far smaller
(\Cref{sec:analysis:teacher-vs-student}).

\section{Ablations}
\label{sec:analysis}

\subsection{Distillation versus same-size from-scratch training}
\label{sec:analysis:from-scratch}
The production recipe trains a large teacher and distils it, which
raises the obvious control: is the teacher needed at all, or could a
$\studentL$ encoder, the size of the deployed \tesseravtwo{}-2B-L
student, be trained from scratch to the same quality? We test this
directly. Taking the identical $\studentL$ architecture and the
\barlowtwins{} objective of \Cref{sec:teacher}, and giving it the
teacher's own data budget, we train it from scratch to convergence. It
reaches an \aebench{} composite of $0.527$, below even \tesseravone{}
($0.541$) and well below the $0.593$ of the distilled student of the
same size (\Cref{tab:from-scratch}). This is the comparison the
per-encoder-best curve of \Cref{fig:scaling-laws}f already traces out:
that curve is the from-scratch score of pixel-wise encoders across the
sweep, and a $\studentL$ encoder sits on it well below both the
distilled student and the teacher. The $0.066$ gap is what
distillation buys. The $2$B teacher learns a representation that a
same-size encoder cannot recover on its own, and distillation
transfers most of it into $\studentL$ parameters, whose own
from-scratch ceiling is $0.527$ (the teacher itself scores $0.608$). Train-large-then-distil is therefore not a
convenience but a requirement: $\studentL$ parameters are enough to
\emph{host} the teacher's representation but not to \emph{discover} it.

\begin{table}[!htbp]
  \centering
  \caption{\textbf{A same-size encoder cannot be trained from scratch
  to the distilled student's quality.} \aebench{} composite for a
  $\studentL$ encoder trained from scratch with \barlowtwins{} on the
  teacher's data budget, versus distilled from the $2$B teacher; the
  $2$B teacher is an upper reference.}
  \label{tab:from-scratch}
  \small
  \begin{tabular}{@{}lc@{}}
  \toprule
  $\studentL$ encoder & \aebench{} composite \\
  \midrule
  From scratch (\barlowtwins{}, teacher data budget) & $0.527$ \\
  Distilled from the $2$B teacher (\tesseravtwo{}-2B-L) & $\mathbf{0.593}$ \\
  \addlinespace[2pt]
  \quad\emph{$2$B teacher, upper reference} & $0.608$ \\
  \bottomrule
  \end{tabular}
\end{table}

\subsection{Coordinate-exchangeability test}
\label{sec:analysis:matryoshka-fails}
\Cref{sec:pixel:why-not-bt} predicted that naive
\matryoshka-\barlowtwins{} should exhibit a gradient-imbalance
fingerprint at exactly the prefix boundaries $d\!=\!16, 32, 64$,
and \Cref{fig:matryoshka-empirical} confirmed that prediction both
at the level of representation geometry and at the level of
downstream task performance. The comparison against
\tesseravone{} below isolates \emph{which symmetry is broken, and
how}.

For each of three methods we evaluate $k\!=\!16$ prefixes
constructed three ways: \textsc{first-16}, \textsc{random-16},
\textsc{last-16}.
\begin{itemize}[leftmargin=*,itemsep=2pt,topsep=2pt]
  \item \emph{\tesseravone}~\citep{feng2026tessera} produces nearly
        identical performance from all three prefix variants,
        consistent with the rotation symmetry of \barlowtwins:
        there is no preferred coordinate ordering, and any
        $16$-dimensional subspace is essentially equivalent.
  \item \emph{Naive \matryoshka-\barlowtwins{}} produces the
        ordering \textsc{first-16}~$>$~\textsc{random-16}~$>$%
        ~\textsc{last-16}, i.e., the prefix losses do break the
        rotation symmetry, but the absolute quality of
        \textsc{first-16} is not better than \tesseravone's, and
        is sometimes worse. This is the gradient-imbalance
        signature visualised in \Cref{fig:matryoshka-empirical}:
        dimensions $0$--$15$ are overconstrained rather than
        informationally privileged.
  \item \emph{Distilled \matryoshka} produces a clean ordering with
        strong absolute quality at small $k$: its \textsc{first}-16
        prefix is the best of any method, ahead of the corresponding
        \tesseravone{} prefix and markedly ahead of the teacher's own
        coordinate prefix, and at $k\!=\!128$ it converges to
        \tesseravone{} and the teacher coordinate prefix.
\end{itemize}
Naive \matryoshka-\barlowtwins{} does break the rotation symmetry
of \barlowtwins, but it breaks it in the wrong way; only
distillation supplies the external target distribution that lets
prefix learning function as ordered compression rather than as
gradient-imbalance overtraining.

\paragraph{Protocol.} We evaluate three prefix-construction
protocols on each model's $128$-dimensional output:
\textsc{first}-16 (the first $16$ coordinates), \textsc{random}-16
(a uniformly sampled $16$-coordinate subset, averaged over five
seeds), and \textsc{last}-16 (the final $16$ coordinates).
Performance is the composite chance-adjusted downstream score
(\Cref{sec:scaling:design}) over the seven classification datasets
of \Cref{fig:matryoshka-empirical}(b); the $k\!=\!\dimfull$ column
reports the same composite at full dimensionality
(\Cref{tab:matryoshka-analysis}).

\begin{table}[!htbp]
  \centering
  \caption{Coordinate-exchangeability test (composite
  chance-adjusted downstream score over the $7$ classification
  datasets of \Cref{fig:matryoshka-empirical}(b), $\uparrow$). Best
  \textsc{first}-16 and best $k\!=\!\dimfull$ in bold.}
  \label{tab:matryoshka-analysis}
  \small
  \begin{tabular}{lcccc}
  \toprule
  Method & \textsc{first}-16 & \textsc{rand}-16 & \textsc{last}-16 &
  $k\!=\!\dimfull$ \\
  \midrule
  \tesseravone{}                   & $0.470$ & $0.473$ & $0.465$ &
  $0.559$ \\
  Naive \matryoshka-\barlowtwins{} & $0.441$ & $0.382$ & $0.360$ &
  $0.537$ \\
  Teacher coordinate prefixes      & $0.430$ & $0.427$ & $0.420$ &
  $\mathbf{0.565}$ \\
  Distilled \matryoshka{} (Ours)   & $\mathbf{0.486}$ & $0.472$ &
  $0.458$ & $0.553$ \\
  \bottomrule
  \end{tabular}
\end{table}

In \Cref{tab:matryoshka-analysis}, naive \matryoshka-\barlowtwins{}
separates \textsc{first}-16 from \textsc{random}-16 by $0.059$,
confirming that prefix losses break coordinate exchangeability, but
its \textsc{first}-16 score ($0.441$) remains \emph{below} the
distilled \textsc{random}-16 score ($0.472$), showing that the
naive \textsc{first} advantage is not evidence of useful ordered
compression. \tesseravone{} is approximately
coordinate-exchangeable, with \textsc{first}/\textsc{random}/%
\textsc{last} all near $0.47$. Teacher coordinate prefixes are also
weak at $k\!=\!16$ ($0.430$) because the teacher's coordinate basis
is not itself ordered by downstream utility. Distilled
\matryoshka{} obtains the strongest \textsc{first}-16 score
($0.486$) and a near-best full-dimensional score ($0.553$),
because its ordering signal comes from reconstructing a fixed
teacher target rather than from repeated self-supervised losses on
early coordinates.

\subsection{Teacher prefix versus student prefix}
\label{sec:analysis:teacher-vs-student}
A direct prediction of \Cref{sec:pixel:ordered-compression} is that
the \emph{student}'s first $k$ dimensions should be a learned
compression of the teacher's full embedding, not an imitation of
the teacher's first $k$ coordinates. The learned student prefix
$\bs_{1:k}$ is supervised to reconstruct the full teacher embedding
$\bt$, whereas a teacher coordinate-prefix $\bt_{1:k}$ is merely the
first $k$ coordinates of $\bt$ and has no reason to be an optimal
$k$-dimensional code: the teacher's coordinate basis is not ordered
by downstream utility.

We fit reconstructors from $\bs_{1:k}\!\to\!\bt$ and from
$\bt_{1:k}\!\to\!\bt$ and report cosine similarity, MSE, and linear
$R^{2}$ to the full teacher target
(\Cref{tab:prefix-reconstruction}). At every $k$ the student
prefix is a markedly better code for $\bt$ than the teacher
coordinate-prefix; the gap is largest at $k\!=\!16$ (cosine
$0.842$ vs.\ $0.684$, $R^{2}$ $0.621$ vs.\ $0.358$) and contracts
with $k$, vanishing at $k\!=\!\dimfull$ where both use the full
coordinate set. Spectral diagnostics (normalised effective rank,
CKA, and Procrustes alignment to the full teacher) tell the same
story and are reported in \Cref{app:matryoshka-failure}.

\begin{table}[!htbp]
  \centering
  \caption{Prefix-to-teacher reconstruction: fit quality of the
  student prefix $\bs_{1:k}$ vs.\ the teacher coordinate-prefix
  $\bt_{1:k}$ as codes for the full teacher embedding $\bt$.}
  \label{tab:prefix-reconstruction}
  \small
  \begin{tabular}{clccc}
  \toprule
  $k$ & Source & Cosine to $\bt$ $\uparrow$ & MSE $\downarrow$ &
  Linear $R^{2}$ $\uparrow$ \\
  \midrule
  \multirow{2}{*}{$16$}
    & student $\bs_{1:16}$  & $0.842$ & $0.178$ & $0.621$ \\
    & teacher $\bt_{1:16}$  & $0.684$ & $0.304$ & $0.358$ \\
  \multirow{2}{*}{$32$}
    & student $\bs_{1:32}$  & $0.905$ & $0.118$ & $0.759$ \\
    & teacher $\bt_{1:32}$  & $0.764$ & $0.238$ & $0.512$ \\
  \multirow{2}{*}{$64$}
    & student $\bs_{1:64}$  & $0.956$ & $0.062$ & $0.884$ \\
    & teacher $\bt_{1:64}$  & $0.863$ & $0.146$ & $0.692$ \\
  \bottomrule
  \end{tabular}
\end{table}

\subsection{Teacher size}
\label{sec:analysis:teacher-size}
We ablate teacher size by repeating the distillation pipeline from
teachers of the same architectural family at different widths and
measuring student performance. As predicted by the scaling laws,
larger teachers yield better students at fixed student size, and
they also yield better prefix retention: a larger teacher provides
a richer target distribution against which early-coordinate
compression has more to exploit (\Cref{tab:teacher-size}). Both
the full-dimensional composite and the $d{=}16$ retention ratio
rise with teacher size, from $91.1\%$ retention for the $500$\,M
teacher to $92.1\%$ for the $\teacherparams$ teacher used in all
reported results. This is the load-bearing reason ``train large''
precedes ``distill flexibly'': the teacher absorbs the pretraining
compute, while the student learns a deployable ordered code over
the teacher's representation distribution, and a weak teacher
cannot be papered over by clever distillation.

\begin{table}[!htbp]
  \centering
  \caption{Teacher size and prefix supervision:
  \tesseravtwo-L student composite at $d{=}16$ and $d{=}128$, and
  the $d{=}16$ retention ratio, as a function of teacher size.}
  \label{tab:teacher-size}
  \small
  \begin{tabular}{lccc}
  \toprule
  Teacher size & $d{=}16$ & $d{=}128$ & $d{=}16$ retained \\
  \midrule
  $500$\,M          & $0.513$ & $0.563$ & $91.1\%$ \\
  $1$\,B            & $0.534$ & $0.584$ & $91.4\%$ \\
  $2$\,B (reported) & $0.546$ & $0.593$ & $92.1\%$ \\
  \bottomrule
  \end{tabular}
\end{table}

\subsection{Storage and deployment trade-offs}
\label{sec:analysis:storage}
For embedding-as-data, the storage scaling is multiplicative:
$d=16$ uses $\nicefrac{1}{8}$ of the storage of $d=128$, and
combined with int8 quantisation~\citep{feng2026tessera} the saving
relative to fp32 \dimfull{}-D embeddings is $32\times$. At
planetary scale, this changes whether global annual embeddings fit
on a workstation or require a cloud-hosted store.
\Cref{tab:family-prefix} reports the per-prefix composite for the full student family, \Cref{tab:matryoshka-retention} the retention ratios for \tesseravtwo-2B-L, and \Cref{tab:storage-tradeoff} the corresponding storage footprints: each prefix corresponds to a global annual storage
budget at $10\,\mathrm{m}$ land scale, and the \matryoshka{} layout
turns $d$ into a deployment knob that users set against their I/O
and storage budget without retraining, preserving the same model
family, coordinate convention, and downstream interface. $d{=}16$
retains $\sim\!92\%$ of $d{=}128$ performance on average with
negligible degradation on classification-like tasks, while
$d{=}64$ recovers nearly the full composite at half the storage
and is the recommended default when regression or fine-grained
class distinctions are central.

\begin{table}[!htbp]
  \centering
  \caption{Composite downstream score over the $\npixeldata$ \aebench{}
  tasks for each distilled student and \matryoshka{} prefix.}
  \label{tab:family-prefix}
  \small
  \begin{tabular}{lcccc}
  \toprule
  Student & $d{=}16$ & $d{=}32$ & $d{=}64$ & $d{=}128$ \\
  \midrule
  \tesseravtwo-2B-N & $0.516$ & $0.542$ & $0.555$ & $0.560$ \\
  \tesseravtwo-2B-S & $0.531$ & $0.559$ & $0.572$ & $0.577$ \\
  \tesseravtwo-2B-M & $0.540$ & $0.567$ & $0.581$ & $0.586$ \\
  \tesseravtwo-2B-L & $0.546$ & $0.574$ & $0.588$ & $0.593$ \\
  \bottomrule
  \end{tabular}
\end{table}

\begin{table}[!htbp]
  \centering
  \caption{Composite downstream score per \matryoshka{} prefix for
  \tesseravtwo-2B-L, mean over the $\npixeldata$ \aebench{} tasks.}
  \label{tab:matryoshka-retention}
  \small
  \begin{tabular}{lcccc}
  \toprule
  Prefix $d$ & $16$ & $32$ & $64$ & $128$ \\
  \midrule
  Composite             & $0.546$ & $0.574$ & $0.588$ & $0.593$ \\
  Retained vs $d{=}128$ & $92.1\%$ & $96.8\%$ & $99.2\%$ & $100\%$ \\
  \bottomrule
  \end{tabular}
\end{table}

\begin{table}[!htbp]
  \centering
  \caption{Storage--accuracy trade-off across \matryoshka{}
  prefixes for \tesseravtwo-2B-L (int8). Global annual storage
  assumes $\approx\!1.5\!\times\!10^{12}$ land \dpixels{} at
  $10\,\mathrm{m}$ (\Cref{app:flops:inference}).}
  \label{tab:storage-tradeoff}
  \footnotesize
  \setlength{\tabcolsep}{3pt}
  \resizebox{\linewidth}{!}{%
  \begin{tabular}{ccccl}
  \toprule
  Prefix $d$ & Storage/\dpixel{} & Global annual storage &
  Composite & Recommended use \\
  \midrule
  $16$  & $16$\,B  & ${\sim}22$\,TiB  & $0.546$ &
    Edge / on-device; cheapest broad-coverage product \\
  $32$  & $32$\,B  & ${\sim}44$\,TiB  & $0.574$ &
    Regional, storage-constrained pipelines \\
  $64$  & $64$\,B  & ${\sim}89$\,TiB  & $0.588$ &
    Balanced product; near-full at half storage \\
  $128$ & $128$\,B & ${\sim}178$\,TiB & $0.593$ &
    Research / maximum-accuracy workflows \\
  \bottomrule
  \end{tabular}}
\end{table}

\section{Embedding artefacts and temporal stability}
\label{app:artefacts}
This appendix expands the embedding-quality summary of
\Cref{sec:pixel:artefacts}.

\begin{figure}[!htbp]
  \centering
  \includegraphics[width=\linewidth]{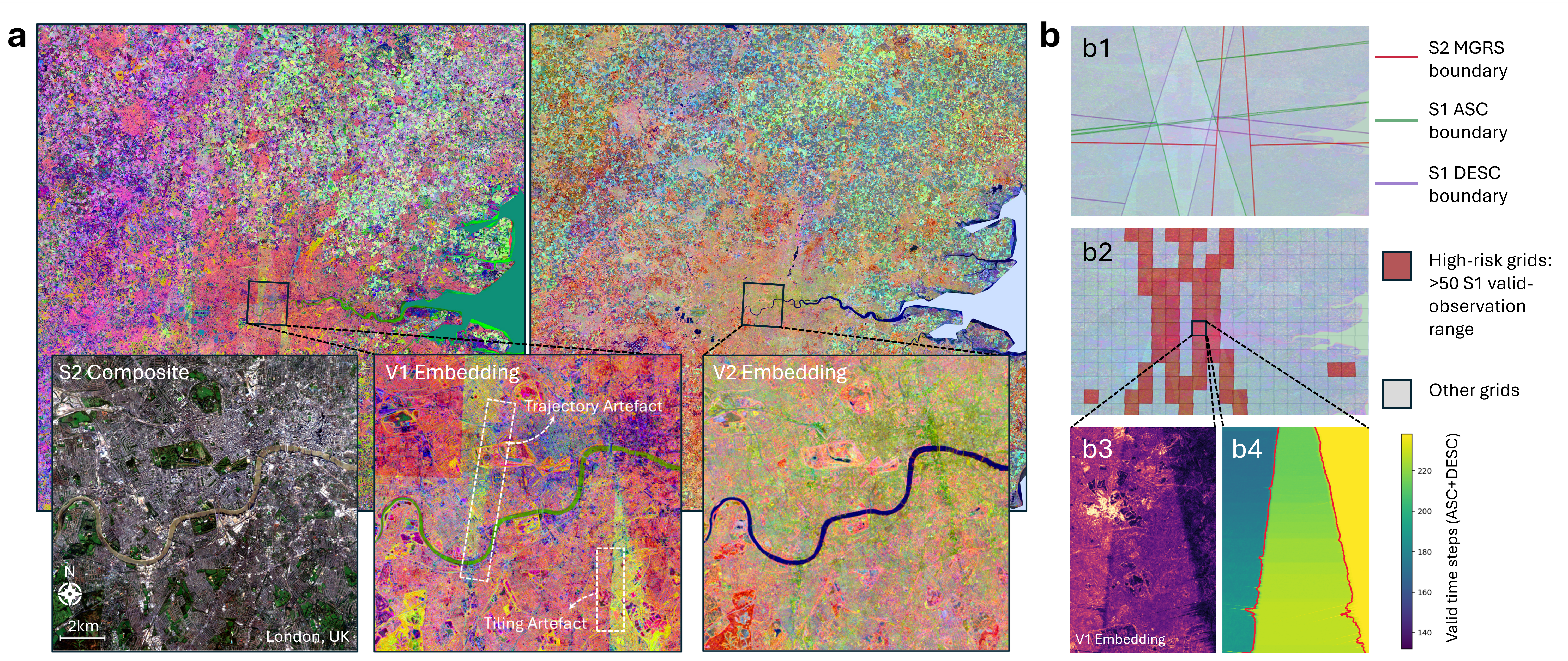}
  \caption{Acquisition-driven artefacts over London.
  \textbf{(a)}~RGB renderings of v1 and v2 embeddings; v1 shows
  along-track striping and seam discontinuities, v2 does not.
  \textbf{(b)}~Diagnosis: (b1)~\stwo{} MGRS and \sone{}
  ascending/descending boundaries overlaid on the embedding field;
  (b2)~high-risk cells on the $0.1^{\circ}$ inference grid;
  (b3)~a sharp v1 discontinuity in one such cell, coincident with
  (b4)~a strong gradient in \sone{} valid-observation count.}
  \label{fig:artefacts}
\end{figure}

\paragraph{Removal of acquisition-driven artefacts.}
\tesseravone{}'s embeddings, rendered as RGB false-colour maps,
exhibit visible \emph{trajectory} artefacts (along-track striping)
and \emph{tiling} artefacts (seam discontinuities at processing
boundaries): the embedding inherits, rather than absorbs,
sensor-acquisition nuisance (\Cref{fig:artefacts}a, insets). The
v2 student embeddings do not show these artefacts on the same
regions, while preserving genuine geographic structure: the River
Thames, the urban core, and the surrounding countryside all remain
crisp. We attribute the improvement to two changes: the larger
teacher learns a more sampling-invariant temporal representation
under \barlowtwins{} (aided by multi-scale training and
full-observation inference), and distillation transfers this
invariance to the deployed students rather than forcing them to
relearn it.

\paragraph{The artefacts are acquisition-driven and predictable.}
The artefacts in v1 are not random noise; they are systematic and
tied to sensor-acquisition geometry, which we establish through a
four-step diagnosis (\Cref{fig:artefacts}b). \emph{First},
overlaying \stwo{} MGRS boundaries and \sone{}
ascending/descending acquisition boundaries on the embedding field
shows that many visible discontinuities align with this geometry
rather than with land-cover boundaries (\Cref{fig:artefacts}(b1)).
\emph{Second}, the $0.1^{\circ}$ grid we use for global
precomputation lets us flag high-risk cells in advance: those that
intersect a \sone{} tangent boundary \emph{and} whose within-cell
range of valid \sone{} observation counts exceeds $50$
(\Cref{fig:artefacts}(b2)). \emph{Third}, a representative
high-risk cell indeed contains a sharp local discontinuity in the
v1 embedding (\Cref{fig:artefacts}(b3)). \emph{Fourth}, that same
cell exhibits a strong spatial gradient in \sone{}
valid-observation count, and the embedding discontinuity is
spatially coincident with it (\Cref{fig:artefacts}(b4)). The
mechanism is that regions receiving different numbers of valid
\sone{} observations induce different embedding statistics in v1,
which surface as acquisition-aligned seams. \tesseravtwo's
full-observation bucket inference (\Cref{sec:teacher:bucket})
removes the dependence on a fixed observation budget, which is why
the same regions are artefact-free under v2.

\paragraph{Inter-annual stability across $2017$--$2025$.} Annual
embeddings are intended to support time-series analyses
(disturbance detection, land-cover change). A good annual
representation should be stable on stable land cover and respond
cleanly to actual change, returning to a clean baseline once a
transient event has passed. On stable land-cover regions (intact
forest, urban cores) the consecutive-year cosine distance between
v2 embeddings is markedly lower than v1's; on disturbance pixels
(logging area) both models register the event, but only v2 returns
to a clean post-disturbance baseline, while v1 remains noisy
(\Cref{fig:teaser}(d2)).

\section{Broader impact}
\label{sec:disc:impact}
Lower-cost EO representations expand who can build planetary-scale
analyses (ecologists, smallholder agriculture programmes, climate
adaptation efforts, disaster-response teams) without proprietary
infrastructure. The same representations can in principle support
surveillance and sensitive land-use monitoring. Our release policy
is open under a CC0 license: we release the distilled pixel-wise
students and the \geotessera{} data product, as well as the $2B$ teacher checkpoints for full reproducibility and extensibility.